\documentclass{article} % For LaTeX2e
\usepackage{iclr2027_conference,times}

\usepackage{amsmath,amssymb,mathtools}
\usepackage{booktabs,tabularx,longtable,array,multirow,makecell}
\usepackage{xcolor}
\usepackage{colortbl}
\usepackage{bbding}
\usepackage{threeparttable}
\usepackage{graphicx}
\usepackage{fvextra}
\usepackage[skins,breakable]{tcolorbox}
\usepackage{enumitem}
\usepackage{xspace}
\usepackage{microtype}
\usepackage[bottom]{footmisc}
\usepackage{hyperref}
\usepackage{url}

\DeclareTextSymbolDefault{\textasciicircum}{T1}
\hypersetup{
  colorlinks=true,
  linkcolor=blue!55!black,
  citecolor=blue!55!black,
  urlcolor=blue!55!black
}

\newcommand{\method}{\texttt{EHR-RobustGym}\xspace}
\newcommand{\ehrrobust}{\method}
\newcommand{\answermatch}{\operatorname{Answer\,Match}}

\newcommand{\abstentionvalidity}{\operatorname{Abstention\,Validity}}
\newcommand{\tasksuccess}{\operatorname{Success}}

\definecolor{EHRTablePurple}{HTML}{66538F}
\definecolor{EHRTablePurpleSoft}{HTML}{E7E0F1}
\definecolor{EHRTablePurplePale}{HTML}{F4F1F8}
\definecolor{RoyalPurple}{RGB}{120,81,169}
\definecolor{EHRTableCheck}{HTML}{009B55}
\definecolor{EHRTableCross}{HTML}{FF0000}
\newcommand{\cmark}{\textcolor{EHRTableCheck}{\CheckmarkBold}}
\newcommand{\xmark}{\textcolor{EHRTableCross}{\XSolidBrush}}

\newcommand{\ehrtablenote}[1]{\par\vspace{0.35em}{\footnotesize\raggedright\textit{Note.} #1\par}}
\definecolor{EHRLowSuccess}{RGB}{254,237,237}
\makeatletter
\newcommand{\ehrlow}[1]{%
  \begingroup\setlength{\fboxsep}{0pt}%
  \kern-\ehrlowleft
  \colorbox{EHRLowSuccess}{\makebox[\dimexpr\ehrvaluewidth+\ehrlowleft+\ehrlowright\relax][r]{%
    \rule[-\dp\@arstrutbox]{0pt}{\dimexpr\ht\@arstrutbox+\dp\@arstrutbox\relax}%
    #1\hspace*{\ehrlowright}}}%
  \kern-\ehrlowright\endgroup}
\makeatother
\DeclareRobustCommand{\ehrclean}{\texorpdfstring{\scalebox{0.86}{Clean}}{Clean}}
\newcommand{\ehrappendixcontents}{%
  \begingroup
  \renewcommand{\contentsname}{Appendix Contents}%
  \hypersetup{linkcolor=black}%
  \tableofcontents
  \endgroup}

\title{EHR-RobustGym: Benchmarking and Training Agents for Robust Clinical Reasoning}
\author{{\fontsize{9.2}{11.5}\selectfont\bfseries Yitong Qiao$^{1,2,}$\thanks{Work done during Yitong's internship at Ant Group.}\hspace{0.2em},
Yancheng Jin$^{1,2}$,
Lei Liu$^{1,2,}$\thanks{Corresponding authors: Lei Liu and Zhixuan Chu.}\hspace{0.2em},
Yue Shen$^2$, Jian Wang$^2$, Jinjie Gu$^2$,
Zhixuan Chu$^{1,\dagger}$}\\
{\normalfont $^1$Zhejiang University \quad $^2$Ant Healthcare, Ant Group}\\
{\normalfont
\href{mailto:qiaoyt@zju.edu.cn}{\texttt{qiaoyt@zju.edu.cn}};
\href{mailto:liulei1497@gmail.com}{\texttt{liulei1497@gmail.com}};
\href{mailto:zhixuanchu@zju.edu.cn}{\texttt{zhixuanchu@zju.edu.cn}}}}
\date{}

\hypersetup{
  pdftitle={EHR-RobustGym: Benchmarking and Training Agents for Robust Clinical Reasoning},
  pdfauthor={Yitong Qiao, Yancheng Jin, Lei Liu, Yue Shen, Jian Wang, Jinjie Gu, Zhixuan Chu}
}
\iclrfinalcopy
\begin{document}
\AddToShipoutPictureFG*{%
  \AtPageUpperLeft{%
    \put(\LenToUnit{108bp},\LenToUnit{-78bp}){%
      \includegraphics[height=17bp]{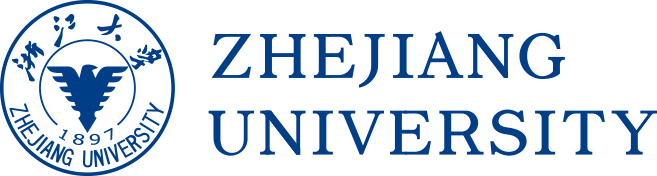}\hspace{18bp}%
      \includegraphics[height=17bp]{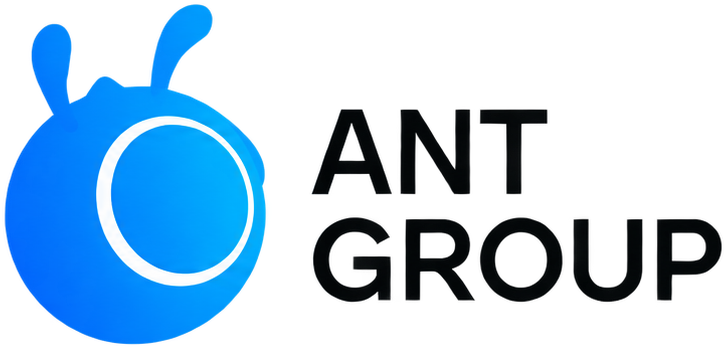}%
    }%
  }%
}
\addtocontents{toc}{\protect\setcounter{tocdepth}{-1}}
\maketitle
\lhead{\parbox{\textwidth}{\footnotesize EHR-RobustGym: Benchmarking and Training Agents for Robust Clinical Reasoning}}
\setlength{\headheight}{22pt}
\suppressfloats[t]

\begin{abstract}
In hospital workflows, electronic health records (EHRs) are often noisy, and may not contain the evidence needed to confirm events or measurements referenced in a clinical query.
Even when database retrieval succeeds, clinical agents can overlook such discrepancies and return plausible but unsupported answers.
We introduce \ehrrobust{}, a scalable and interactive environment for evaluating and training robust clinical agents grounded in noisy EHRs. Built on MIMIC-IV hospital records (365K patients, 31 tables, and over 500M records), \ehrrobust{} comprises 5,486 Clean-Noise pairs spanning six clinical intents and both patient-level and population-level queries. The pairs test robustness to Record-level, Value-level, and Query-level noise, while interactive SQL/Python execution and outcome verification support trajectory collection and training. Evaluating multiple LLMs reveals substantial robustness gaps: average task success across proprietary and large-scale open-weight models drops from 62.2\% on Clean questions to 37.9\% on Noise questions.
At $k=4$, pass\textasciicircum{}k consistency falls below 50\% for most evaluated models, exposing instability in clinical task completion. Supervised fine-tuning and reinforcement learning in \ehrrobust{} improve performance, with gains generalizing to five external EHR benchmarks. Together, these results position \ehrrobust{} as a testbed for evaluating and improving the evidence-grounded robustness of clinical agents.

\end{abstract}

\begin{figure}[!bp]
  \centering
  \includegraphics[height=5.76cm,keepaspectratio]{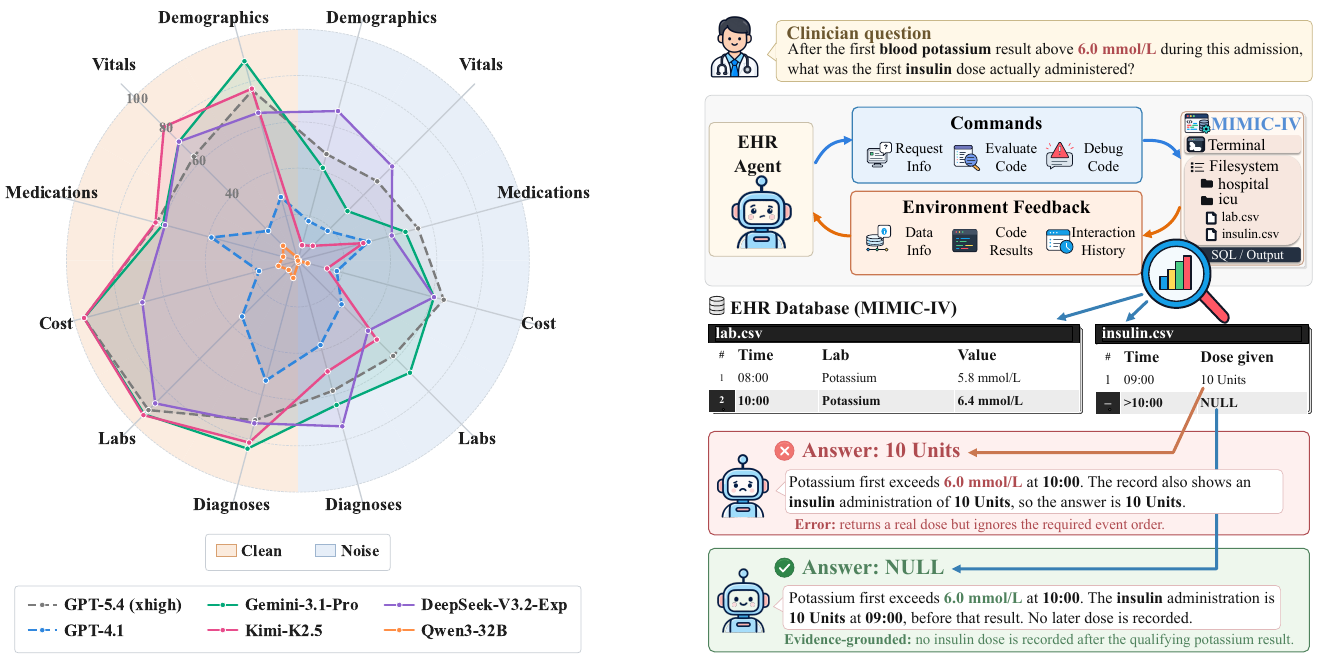}
  \caption{Performance Comparisons under \ehrclean{} and Noise Conditions.
  \textbf{(Left):} Base success rates (\%) for selected models across six clinical intents.
  \textbf{(Right):} A Query-level Noise request asks for the first insulin dose administered after blood potassium first exceeds 6.0~mmol/L.
  The incorrect response reports 10~Units given at 09:00, overlooking that the qualifying potassium result occurs at 10:00.
  In contrast, the evidence-grounded response returns \texttt{NULL} because the EHR contains no insulin administration after that result.}
  \label{fig:conflict-example}
\end{figure}

\section{Introduction}
\label{sec:introduction}

Electronic health records (EHRs) support the reconstruction of patient histories and the analysis of patient populations~\citep{ehrsql,ehrcomplex}. Yet they may omit relevant events, contain outdated or inconsistent copied information, or fail to support assumptions embedded in clinical questions~\citep{lin2018outsystem,weiskopf2013completeness,tsou2017copypaste}. Although large language model (LLM) agents provide a natural-language interface to EHR databases~\citep{ehragent,medagentgym}, these limitations complicate reliable answering. Therefore, robust clinical agents must do more than retrieve records or execute queries: they must verify that the evidence satisfies the question's constraints. For example, an agent asked for the first insulin dose administered after blood potassium first exceeds 6.0~mmol/L might return a dose given before the qualifying result (Figure~\ref{fig:conflict-example}). The answer may appear plausible, but it violates the required temporal order. The agent should identify this mismatch and abstain if no supported answer is available.

Recent work has advanced clinical database agents through interactive execution and trajectory-based training~\citep{ehragent,medagentgym}. EHR-ChatQA studies question clarification and terminology mismatches through user and tool interactions, while EHR-Complex evaluates longitudinal, multi-table reasoning over hospital-scale records~\citep{ehrchatqa,ehrcomplex}. We focus on a complementary requirement: recognizing when a question's necessary conditions are unsupported, without changing the request or rejecting answerable questions. Evaluation requires examining supported and unsupported requests, since an agent that refuses more often may appear cautious without becoming more reliable. Improving it requires interactive training environments with verifiable learning signals for accurate answering and evidence-grounded abstention.

To this end, we introduce \ehrrobust{}, a scalable and interactive environment for benchmarking and training robust clinical agents. Built on MIMIC-IV records containing 365K patients, 31 tables, and over 500M records~\citep{mimiciv31}, \ehrrobust{} comprises 5,486 Clean-Noise pairs spanning six clinical intents and both patient- and population-level queries. Each pair shares a clinical goal but differs in one necessary condition, contrasting a supported Clean question with an unsupported Noise counterpart without modifying the underlying database. Record-level, Value-level, and Query-level Noise capture unavailable records, missing or unsupported attributes of existing records, and relations that available records do not jointly satisfy, respectively.

\ehrrobust{} couples these paired tasks with interactive SQL/Python execution, allowing agents to inspect results and refine operations using database feedback. Completed trajectories are evaluated with a shared outcome verifier that checks answer correctness on Clean questions and requires evidence-grounded abstention on Noise questions. A query failure or a refusal alone does not constitute successful handling of an unsupported request: abstention must be justified by evidence obtained during interaction. Trajectory collection and outcome-based rewards enable supervised fine-tuning and reinforcement learning for both accurate answering and justified abstention.

Extensive benchmarking of 12 LLMs reveals substantial robustness gaps, with average task success across proprietary and large-scale open-weight models dropping from 62.2\% on Clean questions to 37.9\% on Noise questions. Repeated-run evaluation further shows that pass\textasciicircum{}k consistency, which requires all $k$ attempts to succeed, falls below 50\% for most evaluated models at $k=4$. These findings highlight that occasional success does not imply dependable task completion across repeated executions. Supervised fine-tuning and reinforcement learning with \ehrrobust{} yield substantial performance gains. The improvements extend to five external EHR benchmarks, demonstrating the environment's value as a training ground beyond its own clinical evaluation tasks.

\begingroup
\begin{table*}[htbp]
    \centering
    \begin{minipage}{0.94\linewidth}
    \caption{Comparison with existing EHR benchmarks.}
    \label{tab:ehrcomplex-comparison}
    \begingroup\fontsize{7.5}{9}\selectfont
    \setlength{\tabcolsep}{2pt}
    \renewcommand\arraystretch{1.12}
    \resizebox{\linewidth}{!}{%
    \begin{tabular}{@{\hspace{8pt}}lccccccrrr@{\hspace{8pt}}}
    \toprule
    \multirow{2}{*}{\textbf{Dataset}} & \multirow{2}{*}{\makecell{\textbf{Agentic}\\\textbf{Mode}}} & \multicolumn{2}{c}{\textbf{Robustness Evaluation}} & \multicolumn{3}{c}{\textbf{Database Scale}} & \multicolumn{3}{c}{\textbf{Task Instances}} \\
    \cmidrule(lr){3-4} \cmidrule(lr){5-7} \cmidrule(lr){8-10}
    & & \textbf{Noise Conditions} & \makecell{\textbf{Trajectory-level}\\\textbf{Evaluation}} & \textbf{Patients} & \textbf{Tables} & \textbf{Elements} & \textbf{Types} & \textbf{Test} & \textbf{Train} \\
    \midrule
    MedAgentBench~\citep{medagentbench} & \cmark & \xmark & \cmark & 100 & - & 700K & 10 & 300 & - \\
    EHR-Complex~\citep{ehrcomplex} & \cmark & \xmark & \cmark & 365K & 31 & \textgreater{}500M & 12 & 3,915 & 48,092 \\
    TREQS~\citep{wang2020text} & \xmark & \xmark & \xmark & 100 & 5 & 2.5M & 4 & 996 & 8,988 \\
    EHRSQL (eICU)~\citep{ehrsql} & \xmark & \xmark\rlap{\textsuperscript{\dag}} & \xmark & \textless{}1K & 10 & 1.5M & 9 & 611 & 6,213 \\
    EHRSQL (MIMIC-III)~\citep{ehrsql} & \xmark & \xmark\rlap{\textsuperscript{\dag}} & \xmark & \textless{}1K & 17 & 1.4M & 9 & 1,122 & 9,318 \\
    EHR-SeqSQL~\citep{ryu2024ehr} & \xmark & \xmark & \xmark & \textless{}1K & 17 & 1.4M & 4 & 7,913 & 18,950 \\
    \midrule
    \rowcolor{RoyalPurple!6} \textbf{\method{} (Ours)} & \textbf{\cmark} & \textbf{\cmark} & \textbf{\cmark} & \textbf{365K} & \textbf{31} & \textbf{\textgreater{}500M} & \textbf{12} & \textbf{2,016} & \textbf{8,956} \\
    \bottomrule
    \end{tabular}}
    \par\vspace{2pt}
    {\raggedright
    \noindent\textsuperscript{\dag} EHRSQL uses databases simplified through patient subsampling and measurement filtering. Its unanswerable questions concern schema incompatibility or external knowledge, without explicit noise evaluation within schema-valid requests.
    \par}
    \endgroup
    \end{minipage}
\end{table*}
\endgroup

Our contributions are as follows:
\begin{itemize}[leftmargin=*]

\item \textbf{A comprehensive benchmark for clinical agent robustness.}
We introduce a hospital-grounded benchmark of 5,486 Clean-Noise pairs covering Record-level, Value-level, and Query-level Noise, jointly evaluating accurate answering and recognition of unsupported clinical requests.

\item \textbf{An interactive environment for evidence-grounded agent training.}
We integrate SQL/Python execution, outcome verification, and trajectory collection to support supervised fine-tuning and reinforcement learning for accurate answering and evidence-grounded abstention.

\item \textbf{Extensive benchmarking and effective training of clinical agents.}
We evaluate 12 LLMs, characterize their robustness and repeated-run consistency, and demonstrate gains from trajectory-based training on \ehrrobust{} and five external EHR benchmarks.

\end{itemize}

\section{Related Work}
\label{sec:related-work}

\textbf{EHR Database Agents.}
Recent work has developed LLM agents that retrieve and reason over EHR databases.
EHRSQL~\citep{ehrsql} studies text-to-SQL over hospital records, while EHRAgent~\citep{ehragent} uses iterative code generation and execution for multi-table reasoning.
MedAgentBench~\citep{medagentbench}, MedAgentGym~\citep{medagentgym}, EHR-ChatQA~\citep{ehrchatqa}, and EHR-Complex~\citep{ehrcomplex} extend this setting to interactive EHR tasks, agent training, clarification, and complex longitudinal reasoning.
EHR-Complex provides fine-grained analysis of failed trajectories, while MedAgentGym includes an outcome-supervised verifier that reads interleaved agent trajectories and an analysis of agent errors~\citep{ehrcomplex,medagentgym}.
Our evaluation distinguishes supported answering from evidence-grounded abstention, while trajectory analysis examines how agents acquire and use evidence from EHR databases.

\textbf{Robustness under Noisy Conditions.}
EHRSQL~\citep{ehrsql} includes questions that are incompatible with the database schema or require external knowledge.
SCARE~\citep{scare} studies answerability and SQL correction, while EHR-ChatQA~\citep{ehrchatqa} examines ambiguous requests and value mismatches.
TrustEHRAgent~\citep{trust_ehr_agent} studies confidence-based abstention, and CuraView~\citep{curaview} verifies generated clinical text against patient records.
We study schema-valid requests with unavailable records, missing or conflicting field values, and unsatisfied relations among records.
\method{} pairs these Noise requests with supported Clean requests and requires retrieved evidence to justify abstention.
It supports training on interaction trajectories and task outcomes to improve accurate answering and evidence-grounded abstention.

\section{EHR-RobustGym}
\label{sec:formulation}

\subsection{Problem Formulation}
\label{sec:evidence-conflict}

\textbf{Task Formulation.}
Given a natural-language clinical question $q$ over an EHR database $D$ with schema $\Sigma$, an agent produces a multi-turn trajectory $\tau$ by executing SQL queries or Python code and using environment feedback to refine its query plan and debug its code.
The objective is to derive the requested answer from database evidence or return \texttt{NULL} with evidence identifying the unmet task requirement.
We distinguish three levels of missing or conflicting evidence.
For population-level questions, these definitions apply within each candidate unit before aggregation.
Population-level answers must also satisfy the minimum population size requirement specified in Appendix~\ref{app:outcome-evaluation}.

\textbf{Record-Level Noise.}
A patient, encounter, or clinical-event record required by $q$ is absent within the specified scope.
$D[u]=\varnothing$, where $D[u]$ denotes records for a required reference $u$ in $q$.

\textbf{Value-Level Noise.}
The required record exists, but a field value needed to answer $q$ is missing or contradicts an explicit assumption about that record.
For a required field with stored value $v$, this corresponds to $v=\texttt{NULL}$ or, when $q$ asserts a value or state $v^q$, to $v\neq v^q$.
When multiple records are candidates, this condition must hold for every candidate within the scope of $q$.

\textbf{Query-Level Noise.}
The required records and field values are available, but no combination satisfies the event-selection or relational conditions in $q$, including thresholds, temporal order, and encounter links.
Let $\mathcal{C}_q(D)$ denote candidate record combinations satisfying the record- and value-level requirements, and let $\phi_q(C)\in\{0,1\}$ indicate whether $C$ satisfies these query conditions.
Query-level Noise occurs when $\mathcal{C}_q(D)\neq\varnothing$ but $\phi_q(C)=0$ for every candidate $C\in\mathcal{C}_q(D)$.

\subsection{Data Construction}
\label{sec:data-construction}

\begin{figure}[t]
  \centering
  \includegraphics[width=0.9\textwidth]{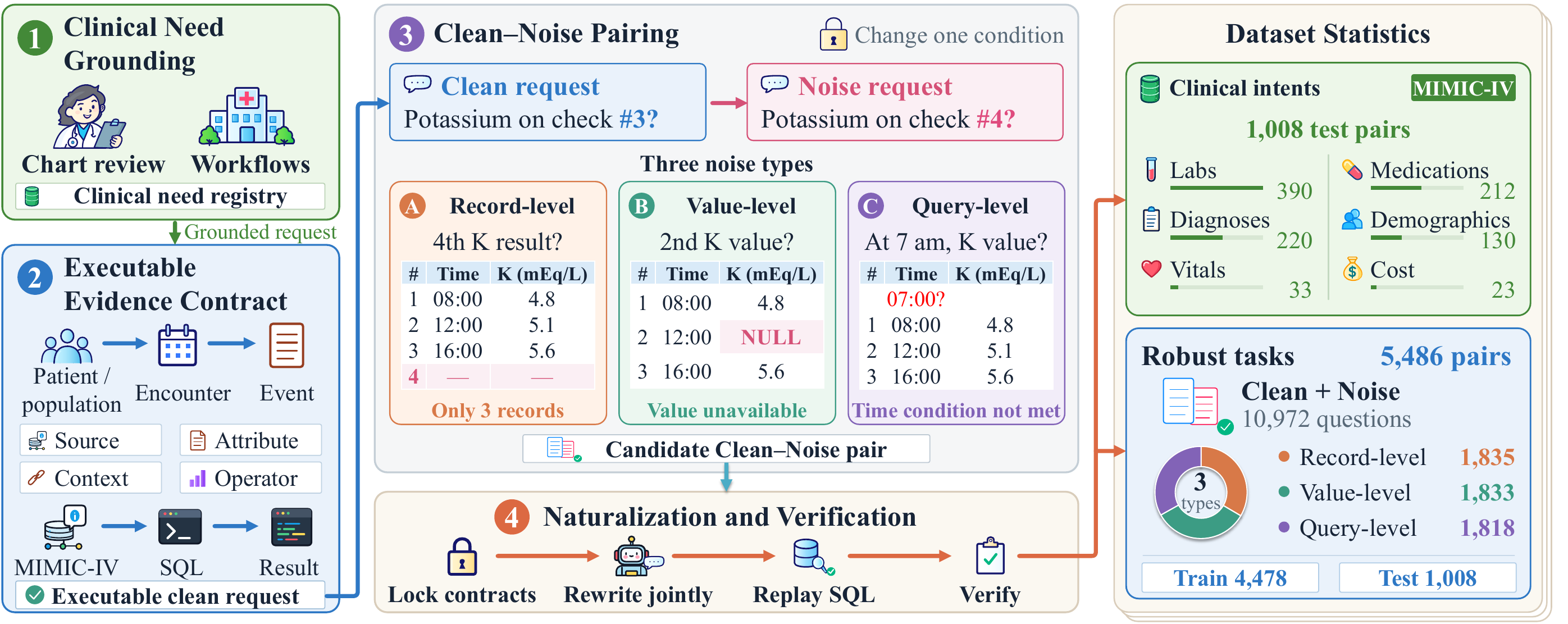}
  \caption{Construction pipeline and task composition of \method{}.
  Step-1. Based on chart-review needs and hospital workflows, clinical requests are derived.
  Step-2. These requests are grounded in executable evidence contracts with reference SQL and verified answers.
  Step-3. \ehrclean{}-Noise pairs are created by changing one necessary condition without modifying the database.
  Step-4. The paired questions are jointly rewritten with fixed contracts, then verified by SQL execution and semantic checks.
  The right panel shows clinical intents, noise types, and data splits.}
  \label{fig:pipeline}
\end{figure}

\textbf{Clinical Task Construction.}
Figure~\ref{fig:pipeline} illustrates the construction of Clean-Noise pairs from clinical evidence and executable SQL queries.
\method{} focuses on verifiable structured EHR tasks grounded in clinical information needs reflected in DiSCQ chart-review questions and EHRSQL workflows~\citep{discq,ehrsql}.
Following the clinical evidence-path construction in EHR-Complex~\citep{ehrcomplex}, we programmatically instantiate patient- and population-level queries over MIMIC-IV~3.1~\citep{mimiciv31}.
Each query encodes the intended temporal constraints and order of operations.
Population-level queries compute within each eligible unit before aggregation.
We retain instances whose reference queries execute successfully and yield answers satisfying the specified task requirements as Clean questions.

\textbf{Clean-Noise Pairing.}
We construct a Noise counterpart $\bar q_t$ for each Clean question $q$ by deterministically changing exactly one necessary condition while preserving the clinical goal and all remaining task requirements.
Here, $t\in\{\mathrm R,\mathrm V,\mathrm Q\}$ denotes the Record-, Value-, or Query-level Noise type, determined by the modified condition.
We retain a pair only when both reference SQL queries execute successfully and confirm that $q$ has a supported answer and $\bar q_t$ requires abstention.

\textbf{Question Rewriting and Verification.}
To improve naturalness and diversity, we use a language model to jointly rewrite each pair while preserving the meaning and computation encoded by the reference SQL. Reference answers are extracted from database execution. Clean answers provide the requested values and units, while Noise answers return \texttt{NULL} and identify the unmet task requirement. We replay the reference queries and conduct an independent language-model audit to verify execution consistency and semantic alignment for both conditions. As an additional semantic check, two reviewers independently audit 344 challenging test pairs (688 questions), with 92.15\% judged valid after adjudication (Appendix~\ref{app:human-validation}). Identified issues are corrected and reverified, and affected tasks are reevaluated so that all reported results correspond to the finalized test set.

\textbf{Dataset Statistics.}
Table~\ref{tab:ehrcomplex-comparison} compares \method{} with existing EHR benchmarks.
\method{} contains 5,486 Clean-Noise pairs across 51 programs and six clinical intents, split into 4,478 training and 1,008 test pairs (8,956 and 2,016 questions, respectively).
The test set is balanced across two query scopes (504 pairs each) and three Noise types (336 pairs each).
The in-domain and structure-disjoint tracks each contain 504 test pairs, with six programs held out for the latter.
Each pair remains within a single split.
Patient-level tasks use disjoint patient identifiers across training and test sets, while population-level tasks use disjoint query specifications but may share patients.
Further dataset construction details are provided in Appendix~\ref{app:construction-algorithm}.

\subsection{Interactive Evaluation and Training}
\label{sec:method}

\begin{figure}[t]
\centering
\includegraphics[width=\linewidth]{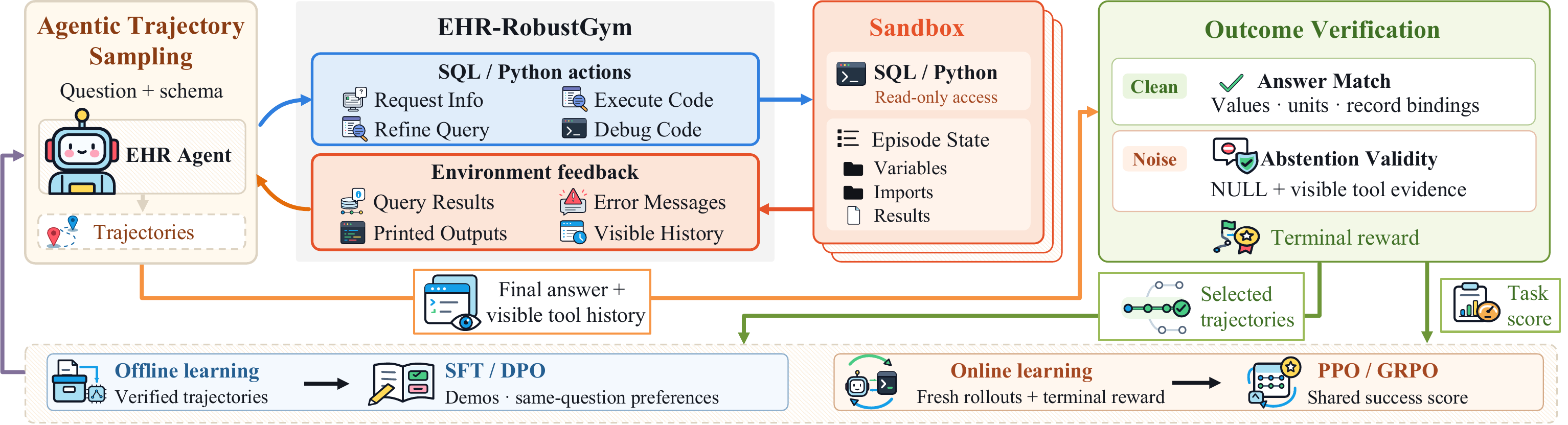}
\caption{Overview of \method{}.
It contains paired \ehrclean{} and Noise tasks with interactive SQL/Python execution and outcome verification for evaluating and training LLM agents.}
\label{fig:mitigation-overview}
\end{figure}

\textbf{Interactive Environment.}
\method{} provides an executable SQL/Python environment with interactive feedback and repeated trajectory sampling (Figure~\ref{fig:mitigation-overview}).
For each episode, the agent receives $x=(q,s)$, where $s$ contains the database schema and shared task instructions.
The agent iteratively refines its queries using execution results and error feedback, then submits a final response containing \texttt{Answer} and \texttt{Reasoning}.
Clean and Noise counterparts are presented in separate episodes under the same interface.
Condition labels, pairing information, reference SQL, and reference answers are hidden from the agent throughout each inference episode.

\textbf{Outcome Verification.}
For each trajectory, the binary success score is formulated as
\begin{equation}
R(x,\tau)=
\begin{cases}
\answermatch(\tau), & q\text{ is Clean},\\
\abstentionvalidity(\tau), & q\text{ is Noise}.
\end{cases}
\label{eq:terminal-reward}
\end{equation}
Answer Match checks the requested values, units, and associated records against the execution-verified reference answer.
Abstention Validity requires \texttt{NULL} and visible evidence from tool calls establishing an unmet task requirement within the scope specified by the question.

\begin{table}[!ht]
\centering
\caption{Task success rates of unadapted EHR agents on \method{}.}
\label{tab:main-results}
\begingroup
\fontsize{7.5}{8.3}\selectfont
\setlength{\tabcolsep}{8.6pt}
\renewcommand{\arraystretch}{1.08}
\definecolor{EHRTrainingGreen}{RGB}{0,128,90}
\definecolor{EHRDeltaPositive}{RGB}{0,115,80}
\definecolor{EHRDeltaNegative}{RGB}{180,45,55}
\newcommand{\ehrpair}[3][16pt]{%
\def\ehrvaluewidth{\dimexpr#1-2.5pt\relax}%
\def\ehrlowleft{\tabcolsep}\def\ehrlowright{0pt}%
\makebox[\dimexpr#1-2pt\relax][r]{{\fontsize{7.1}{8.6}\selectfont #2}\kern0.5pt}%
\makebox[4pt][c]{/}%
\def\ehrvaluewidth{#1}%
\def\ehrlowleft{0pt}\def\ehrlowright{\tabcolsep}%
\makebox[#1][r]{{\fontsize{8.25}{8.6}\selectfont #3}}}
\resizebox{\textwidth}{!}{%
\begin{tabular}{l*{7}{c}}
\toprule
\textbf{Model} & \textbf{Demographics} & \textbf{Vitals} & \textbf{Medications} & \textbf{Cost} & \textbf{Labs} & \textbf{Diagnoses} & \textbf{Avg.} \\
\midrule
\rowcolor{RoyalPurple!6}
\multicolumn{8}{c}{\textbf{Proprietary API models}} \\
GPT-5.4 (xhigh) & \ehrpair{76.2}{47.7} & \ehrpair{63.6}{48.5} & \ehrpair{63.2}{53.8} & \ehrpair{95.7}{65.2} & \ehrpair{91.3}{58.2} & \ehrpair{71.4}{58.2} & \ehrpair{78.3}{55.8} \\
GPT-4.1 & \ehrpair{28.5}{17.7} & \ehrpair{18.2}{18.2} & \ehrpair{38.7}{31.6} & \ehrpair{17.4}{17.4} & \ehrpair{34.1}{26.7} & \ehrpair{53.6}{37.7} & \ehrpair{37.7}{28.5} \\
Gemini-3.1-Pro & \ehrpair{89.2}{41.5} & \ehrpair{72.7}{30.3} & \ehrpair{60.4}{48.1} & \ehrpair{95.7}{60.9} & \ehrpair{93.8}{68.5} & \ehrpair{84.1}{64.5} & \ehrpair{83.4}{58.4} \\
Claude Sonnet 4.6 & \ehrpair{62.3}{28.5} & \ehrpair{60.6}{24.2} & \ehrpair{58.5}{22.6} & \ehrpair{65.2}{13.0} & \ehrpair{64.4}{19.2} & \ehrpair{63.2}{42.7} & \ehrpair{62.5}{26.3} \\
\midrule
\rowcolor{RoyalPurple!6}
\multicolumn{8}{c}{\textbf{Open-weight models $\ge$100B}} \\
Kimi-K2.5 & \ehrpair{76.9}{6.9} & \ehrpair{81.8}{9.1} & \ehrpair{63.7}{29.2} & \ehrpair{95.7}{13.0} & \ehrpair{94.4}{48.2} & \ehrpair{81.4}{49.5} & \ehrpair{82.4}{37.1} \\
DeepSeek-V3.2 & \ehrpair{66.2}{66.9} & \ehrpair{72.7}{57.6} & \ehrpair{59.4}{42.0} & \ehrpair{69.6}{60.9} & \ehrpair{87.2}{42.8} & \ehrpair{72.7}{74.1} & \ehrpair{74.6}{53.5} \\
Qwen3-235B-A22B & \ehrpair{\ehrlow{3.1}}{\ehrlow{0.0}} & \ehrpair{\ehrlow{3.0}}{\ehrlow{0.0}} & \ehrpair{23.6}{\ehrlow{4.7}} & \ehrpair{13.0}{8.7} & \ehrpair{6.7}{5.1} & \ehrpair{38.2}{10.9} & \ehrpair{16.7}{5.6} \\
\bottomrule
\end{tabular}}
\endgroup
\ehrtablenote{\ehrclean{} / Noise success rates (Pass@1, \%, $\uparrow$). \textcolor[RGB]{254,237,237}{\rule{1.2em}{0.7em}}~indicates rates below 5\%.}
\end{table}

\begin{table}[!ht]
\centering
\caption{EHR agent performance after prompting and training with \method{}.}
\label{tab:mitigation-results}
\begingroup
\fontsize{7.5}{8.3}\selectfont
\setlength{\tabcolsep}{3.4pt}
\renewcommand{\arraystretch}{1.08}
\definecolor{EHRTrainingGreen}{RGB}{0,128,90}
\definecolor{EHRDeltaPositive}{RGB}{0,115,80}
\definecolor{EHRDeltaNegative}{RGB}{180,45,55}
\newcommand{\ehrpair}[3][16pt]{%
\def\ehrvaluewidth{\dimexpr#1-2.5pt\relax}%
\def\ehrlowleft{\tabcolsep}\def\ehrlowright{0pt}%
\makebox[\dimexpr#1-2pt\relax][r]{{\fontsize{7.1}{8.6}\selectfont #2}\kern0.5pt}%
\makebox[4pt][c]{/}%
\def\ehrvaluewidth{#1}%
\def\ehrlowleft{0pt}\def\ehrlowright{\tabcolsep}%
\makebox[#1][r]{{\fontsize{8.25}{8.6}\selectfont #3}}}
\resizebox{\textwidth}{!}{%
\begin{tabular}{ll*{8}{c}}
\toprule
\textbf{Model} & \textbf{Method} & \textbf{Demographics} & \textbf{Vitals} & \textbf{Medications} & \textbf{Cost} & \textbf{Labs} & \textbf{Diagnoses} & \textbf{Avg.} & \textbf{$\Delta$ vs. Base} \\
\midrule
\rowcolor{RoyalPurple!6}
\multicolumn{10}{c}{\textbf{Prompt-only mitigation (API-based models)}} \\
GPT-5.4 (xhigh) & PE & \ehrpair{81.5}{43.8} & \ehrpair{69.7}{45.5} & \ehrpair{65.1}{52.4} & \ehrpair{82.6}{52.2} & \ehrpair{91.5}{59.5} & \ehrpair{78.2}{59.1} & \ehrpair{80.9}{55.3} & \ehrpair[21pt]{\textcolor{EHRDeltaPositive}{+2.6}}{\textcolor{EHRDeltaNegative}{-0.5}} \\
GPT-4.1 & PE & \ehrpair{30.0}{23.1} & \ehrpair{18.2}{6.1} & \ehrpair{38.2}{31.1} & \ehrpair{21.7}{21.7} & \ehrpair{35.4}{23.1} & \ehrpair{53.2}{40.5} & \ehrpair{38.3}{28.0} & \ehrpair[21pt]{\textcolor{EHRDeltaPositive}{+0.6}}{\textcolor{EHRDeltaNegative}{-0.5}} \\
Gemini-3.1-Pro & PE & \ehrpair{89.2}{37.7} & \ehrpair{75.8}{48.5} & \ehrpair{63.2}{56.1} & \ehrpair{91.3}{65.2} & \ehrpair{95.6}{70.8} & \ehrpair{85.0}{66.4} & \ehrpair{84.9}{61.6} & \ehrpair[21pt]{\textcolor{EHRDeltaPositive}{+1.5}}{\textcolor{EHRDeltaPositive}{+3.2}} \\
Claude Sonnet 4.6 & PE & \ehrpair{66.2}{33.8} & \ehrpair{78.8}{24.2} & \ehrpair{63.2}{25.0} & \ehrpair{87.0}{17.4} & \ehrpair{68.2}{29.0} & \ehrpair{70.0}{47.3} & \ehrpair{68.1}{32.3} & \ehrpair[21pt]{\textcolor{EHRDeltaPositive}{+5.6}}{\textcolor{EHRDeltaPositive}{+6.1}} \\
Kimi-K2.5 & PE & \ehrpair{77.7}{16.2} & \ehrpair{87.9}{18.2} & \ehrpair{59.9}{33.5} & \ehrpair{91.3}{8.7} & \ehrpair{93.8}{51.0} & \ehrpair{80.9}{49.5} & \ehrpair{81.5}{40.5} & \ehrpair[21pt]{\textcolor{EHRDeltaNegative}{-0.9}}{\textcolor{EHRDeltaPositive}{+3.4}} \\
DeepSeek-V3.2 & PE & \ehrpair{74.6}{66.9} & \ehrpair{81.8}{57.6} & \ehrpair{57.1}{43.4} & \ehrpair{91.3}{73.9} & \ehrpair{83.6}{44.6} & \ehrpair{79.5}{77.3} & \ehrpair{76.1}{55.5} & \ehrpair[21pt]{\textcolor{EHRDeltaPositive}{+1.5}}{\textcolor{EHRDeltaPositive}{+2.0}} \\
Qwen3-235B-A22B & PE & \ehrpair{\ehrlow{2.3}}{\ehrlow{0.0}} & \ehrpair{\ehrlow{3.0}}{\ehrlow{0.0}} & \ehrpair{23.6}{7.1} & \ehrpair{\ehrlow{4.3}}{\ehrlow{4.3}} & \ehrpair{7.9}{6.7} & \ehrpair{42.7}{15.9} & \ehrpair{17.9}{7.6} & \ehrpair[21pt]{\textcolor{EHRDeltaPositive}{+1.2}}{\textcolor{EHRDeltaPositive}{+2.1}} \\
\midrule
\rowcolor{RoyalPurple!6}
\multicolumn{10}{c}{\textbf{Prompting and training (open-weight models $<$100B)}} \\
\cellcolor{white} & Base & \ehrpair{\ehrlow{1.5}}{\ehrlow{0.0}} & \ehrpair{\ehrlow{0.0}}{\ehrlow{0.0}} & \ehrpair{\ehrlow{2.4}}{\ehrlow{0.0}} & \ehrpair{\ehrlow{0.0}}{\ehrlow{0.0}} & \ehrpair{\ehrlow{1.3}}{\ehrlow{0.0}} & \ehrpair{\ehrlow{0.0}}{\ehrlow{0.0}} & \ehrpair{\ehrlow{1.2}}{\ehrlow{0.0}} & \ehrpair[21pt]{-}{-} \\
\cellcolor{white} & PE & \ehrpair{\ehrlow{0.8}}{\ehrlow{0.0}} & \ehrpair{\ehrlow{0.0}}{\ehrlow{0.0}} & \ehrpair{\ehrlow{2.8}}{\ehrlow{0.0}} & \ehrpair{\ehrlow{0.0}}{\ehrlow{0.0}} & \ehrpair{\ehrlow{1.3}}{\ehrlow{0.0}} & \ehrpair{\ehrlow{0.0}}{\ehrlow{0.0}} & \ehrpair{\ehrlow{1.2}}{\ehrlow{0.0}} & \ehrpair[21pt]{0.0}{0.0} \\
\rowcolor{EHRTrainingGreen!7}
\cellcolor{white} & SFT & \ehrpair{17.7}{13.1} & \ehrpair{12.1}{\ehrlow{3.0}} & \ehrpair{29.2}{23.6} & \ehrpair{39.1}{13.0} & \ehrpair{28.7}{26.4} & \ehrpair{44.5}{12.3} & \ehrpair{30.6}{19.9} & \ehrpair[21pt]{\textcolor{EHRDeltaPositive}{+29.4}}{\textcolor{EHRDeltaPositive}{+19.9}} \\
\rowcolor{EHRTrainingGreen!7}
\cellcolor{white} & \hspace*{0.5em}+DPO & \ehrpair{9.2}{6.9} & \ehrpair{15.2}{\ehrlow{3.0}} & \ehrpair{29.2}{9.9} & \ehrpair{39.1}{8.7} & \ehrpair{45.1}{7.7} & \ehrpair{35.9}{9.5} & \ehrpair{34.0}{8.3} & \ehrpair[21pt]{\textcolor{EHRDeltaPositive}{+32.8}}{\textcolor{EHRDeltaPositive}{+8.3}} \\
\rowcolor{EHRTrainingGreen!7}
\cellcolor{white} & \hspace*{0.5em}+PPO & \ehrpair{16.2}{10.8} & \ehrpair{\ehrlow{3.0}}{6.1} & \ehrpair{29.2}{39.6} & \ehrpair{43.5}{30.4} & \ehrpair{52.3}{39.7} & \ehrpair{45.0}{36.8} & \ehrpair{39.4}{34.0} & \ehrpair[21pt]{\textcolor{EHRDeltaPositive}{+38.2}}{\textcolor{EHRDeltaPositive}{+34.0}} \\
\rowcolor{EHRTrainingGreen!7}
\cellcolor{white}\multirow{-6}{*}{Qwen3-4B} & \hspace*{0.5em}+GRPO & \ehrpair{33.1}{24.6} & \ehrpair{27.3}{21.2} & \ehrpair{35.8}{48.6} & \ehrpair{60.9}{56.5} & \ehrpair{60.0}{29.5} & \ehrpair{46.4}{41.4} & \ehrpair{47.4}{35.8} & \ehrpair[21pt]{\textbf{\textcolor{EHRDeltaPositive}{+46.2}}}{\textbf{\textcolor{EHRDeltaPositive}{+35.8}}} \\
\midrule
\cellcolor{white} & Base & \ehrpair{\ehrlow{0.0}}{\ehrlow{0.0}} & \ehrpair{\ehrlow{0.0}}{\ehrlow{0.0}} & \ehrpair{6.1}{\ehrlow{0.0}} & \ehrpair{\ehrlow{0.0}}{\ehrlow{0.0}} & \ehrpair{\ehrlow{0.5}}{\ehrlow{1.3}} & \ehrpair{9.1}{\ehrlow{0.0}} & \ehrpair{\ehrlow{3.5}}{\ehrlow{0.5}} & \ehrpair[21pt]{-}{-} \\
\cellcolor{white} & PE & \ehrpair{\ehrlow{0.0}}{\ehrlow{0.0}} & \ehrpair{\ehrlow{0.0}}{\ehrlow{0.0}} & \ehrpair{\ehrlow{3.3}}{\ehrlow{0.0}} & \ehrpair{\ehrlow{0.0}}{\ehrlow{4.3}} & \ehrpair{\ehrlow{0.5}}{\ehrlow{2.6}} & \ehrpair{\ehrlow{3.6}}{\ehrlow{0.5}} & \ehrpair{\ehrlow{1.7}}{\ehrlow{1.2}} & \ehrpair[21pt]{\textcolor{EHRDeltaNegative}{-1.8}}{\textcolor{EHRDeltaPositive}{+0.7}} \\
\rowcolor{EHRTrainingGreen!7}
\cellcolor{white} & SFT & \ehrpair{21.5}{18.5} & \ehrpair{15.2}{15.2} & \ehrpair{26.4}{25.9} & \ehrpair{21.7}{47.8} & \ehrpair{42.1}{20.5} & \ehrpair{37.7}{25.0} & \ehrpair{33.8}{22.8} & \ehrpair[21pt]{\textcolor{EHRDeltaPositive}{+30.4}}{\textcolor{EHRDeltaPositive}{+22.3}} \\
\rowcolor{EHRTrainingGreen!7}
\cellcolor{white} & \hspace*{0.5em}+DPO & \ehrpair{19.2}{30.8} & \ehrpair{9.1}{27.3} & \ehrpair{17.5}{25.9} & \ehrpair{39.1}{60.9} & \ehrpair{26.4}{43.8} & \ehrpair{27.3}{24.1} & \ehrpair{23.5}{33.9} & \ehrpair[21pt]{\textcolor{EHRDeltaPositive}{+20.0}}{\textcolor{EHRDeltaPositive}{+33.4}} \\
\rowcolor{EHRTrainingGreen!7}
\cellcolor{white} & \hspace*{0.5em}+PPO & \ehrpair{16.9}{20.8} & \ehrpair{6.1}{\ehrlow{3.0}} & \ehrpair{24.1}{21.2} & \ehrpair{30.4}{43.5} & \ehrpair{36.7}{30.5} & \ehrpair{34.5}{12.7} & \ehrpair{29.9}{22.8} & \ehrpair[21pt]{\textcolor{EHRDeltaPositive}{+26.4}}{\textcolor{EHRDeltaPositive}{+22.3}} \\
\rowcolor{EHRTrainingGreen!7}
\cellcolor{white}\multirow{-6}{*}{Qwen3-8B} & \hspace*{0.5em}+GRPO & \ehrpair{38.5}{47.7} & \ehrpair{33.3}{12.1} & \ehrpair{41.0}{45.8} & \ehrpair{69.6}{69.6} & \ehrpair{63.6}{40.5} & \ehrpair{50.0}{27.7} & \ehrpair{51.8}{39.5} & \ehrpair[21pt]{\textbf{\textcolor{EHRDeltaPositive}{+48.3}}}{\textbf{\textcolor{EHRDeltaPositive}{+39.0}}} \\
\midrule
\cellcolor{white} & Base & \ehrpair{\ehrlow{0.0}}{\ehrlow{0.0}} & \ehrpair{\ehrlow{0.0}}{\ehrlow{0.0}} & \ehrpair{\ehrlow{0.9}}{\ehrlow{0.0}} & \ehrpair{8.7}{\ehrlow{0.0}} & \ehrpair{14.9}{\ehrlow{4.9}} & \ehrpair{27.3}{18.2} & \ehrpair{12.1}{5.9} & \ehrpair[21pt]{-}{-} \\
\cellcolor{white} & PE & \ehrpair{\ehrlow{0.8}}{\ehrlow{0.8}} & \ehrpair{\ehrlow{3.0}}{\ehrlow{0.0}} & \ehrpair{\ehrlow{0.5}}{\ehrlow{0.0}} & \ehrpair{8.7}{17.4} & \ehrpair{14.9}{\ehrlow{3.6}} & \ehrpair{8.6}{8.2} & \ehrpair{8.1}{\ehrlow{3.7}} & \ehrpair[21pt]{\textcolor{EHRDeltaNegative}{-4.0}}{\textcolor{EHRDeltaNegative}{-2.2}} \\
\rowcolor{EHRTrainingGreen!7}
\cellcolor{white} & SFT & \ehrpair{23.8}{23.8} & \ehrpair{27.3}{18.2} & \ehrpair{41.0}{33.0} & \ehrpair{34.8}{34.8} & \ehrpair{49.2}{33.1} & \ehrpair{45.5}{48.2} & \ehrpair{42.4}{34.7} & \ehrpair[21pt]{\textcolor{EHRDeltaPositive}{+30.3}}{\textcolor{EHRDeltaPositive}{+28.9}} \\
\rowcolor{EHRTrainingGreen!7}
\cellcolor{white} & \hspace*{0.5em}+DPO & \ehrpair{28.5}{43.8} & \ehrpair{27.3}{27.3} & \ehrpair{44.8}{44.3} & \ehrpair{65.2}{65.2} & \ehrpair{50.0}{55.4} & \ehrpair{37.7}{56.8} & \ehrpair{43.1}{51.2} & \ehrpair[21pt]{\textcolor{EHRDeltaPositive}{+31.0}}{\textcolor{EHRDeltaPositive}{+45.3}} \\
\rowcolor{EHRTrainingGreen!7}
\cellcolor{white} & \hspace*{0.5em}+PPO & \ehrpair{65.4}{69.2} & \ehrpair{51.5}{39.4} & \ehrpair{60.4}{61.8} & \ehrpair{95.7}{95.7} & \ehrpair{77.2}{59.0} & \ehrpair{71.4}{65.9} & \ehrpair{70.4}{62.6} & \ehrpair[21pt]{\textbf{\textcolor{EHRDeltaPositive}{+58.3}}}{\textbf{\textcolor{EHRDeltaPositive}{+56.7}}} \\
\rowcolor{EHRTrainingGreen!7}
\cellcolor{white}\multirow{-6}{*}{Qwen3-14B} & \hspace*{0.5em}+GRPO & \ehrpair{63.1}{72.3} & \ehrpair{60.6}{51.5} & \ehrpair{47.2}{44.8} & \ehrpair{91.3}{78.3} & \ehrpair{78.2}{42.1} & \ehrpair{70.9}{74.5} & \ehrpair{67.9}{54.8} & \ehrpair[21pt]{\textcolor{EHRDeltaPositive}{+55.8}}{\textcolor{EHRDeltaPositive}{+48.9}} \\
\midrule
\cellcolor{white} & Base & \ehrpair{\ehrlow{1.5}}{\ehrlow{0.0}} & \ehrpair{9.1}{\ehrlow{0.0}} & \ehrpair{6.6}{\ehrlow{0.5}} & \ehrpair{8.7}{\ehrlow{4.3}} & \ehrpair{5.6}{\ehrlow{0.5}} & \ehrpair{7.7}{\ehrlow{0.9}} & \ehrpair{6.0}{\ehrlow{0.6}} & \ehrpair[21pt]{-}{-} \\
\cellcolor{white} & PE & \ehrpair{\ehrlow{1.5}}{\ehrlow{0.8}} & \ehrpair{9.1}{\ehrlow{0.0}} & \ehrpair{7.1}{\ehrlow{0.5}} & \ehrpair{13.0}{\ehrlow{0.0}} & \ehrpair{5.1}{\ehrlow{0.5}} & \ehrpair{\ehrlow{2.3}}{\ehrlow{0.9}} & \ehrpair{\ehrlow{4.8}}{\ehrlow{0.6}} & \ehrpair[21pt]{\textcolor{EHRDeltaNegative}{-1.2}}{0.0} \\
\rowcolor{EHRTrainingGreen!7}
\cellcolor{white} & SFT & \ehrpair{36.9}{31.5} & \ehrpair{21.2}{15.2} & \ehrpair{47.6}{45.3} & \ehrpair{30.4}{26.1} & \ehrpair{57.9}{45.1} & \ehrpair{46.8}{37.3} & \ehrpair{48.8}{40.3} & \ehrpair[21pt]{\textcolor{EHRDeltaPositive}{+42.9}}{\textcolor{EHRDeltaPositive}{+39.7}} \\
\rowcolor{EHRTrainingGreen!7}
\cellcolor{white} & \hspace*{0.5em}+DPO & \ehrpair{23.8}{32.3} & \ehrpair{24.2}{18.2} & \ehrpair{45.3}{57.5} & \ehrpair{30.4}{39.1} & \ehrpair{67.9}{61.5} & \ehrpair{41.8}{60.5} & \ehrpair{49.5}{54.8} & \ehrpair[21pt]{\textcolor{EHRDeltaPositive}{+43.6}}{\textcolor{EHRDeltaPositive}{+54.2}} \\
\rowcolor{EHRTrainingGreen!7}
\cellcolor{white} & \hspace*{0.5em}+PPO & \ehrpair{40.8}{61.5} & \ehrpair{39.4}{39.4} & \ehrpair{48.6}{57.1} & \ehrpair{95.7}{87.0} & \ehrpair{72.6}{52.6} & \ehrpair{72.3}{68.6} & \ehrpair{62.8}{58.5} & \ehrpair[21pt]{\textbf{\textcolor{EHRDeltaPositive}{+56.8}}}{\textbf{\textcolor{EHRDeltaPositive}{+57.9}}} \\
\rowcolor{EHRTrainingGreen!7}
\cellcolor{white}\multirow{-6}{*}{Qwen3-32B} & \hspace*{0.5em}+GRPO & \ehrpair{60.0}{54.6} & \ehrpair{42.4}{33.3} & \ehrpair{42.5}{25.0} & \ehrpair{78.3}{56.5} & \ehrpair{78.7}{39.0} & \ehrpair{57.3}{36.4} & \ehrpair{62.8}{37.7} & \ehrpair[21pt]{\textbf{\textcolor{EHRDeltaPositive}{+56.8}}}{\textcolor{EHRDeltaPositive}{+37.1}} \\
\midrule
\cellcolor{white} & Base & \ehrpair{\ehrlow{0.0}}{\ehrlow{0.0}} & \ehrpair{\ehrlow{0.0}}{\ehrlow{0.0}} & \ehrpair{\ehrlow{0.0}}{\ehrlow{0.9}} & \ehrpair{\ehrlow{0.0}}{\ehrlow{0.0}} & \ehrpair{\ehrlow{0.3}}{\ehrlow{0.3}} & \ehrpair{\ehrlow{0.0}}{\ehrlow{0.0}} & \ehrpair{\ehrlow{0.1}}{\ehrlow{0.3}} & \ehrpair[21pt]{-}{-} \\
\cellcolor{white} & PE & \ehrpair{\ehrlow{0.0}}{\ehrlow{0.0}} & \ehrpair{\ehrlow{0.0}}{\ehrlow{0.0}} & \ehrpair{\ehrlow{0.0}}{\ehrlow{0.0}} & \ehrpair{\ehrlow{0.0}}{\ehrlow{0.0}} & \ehrpair{\ehrlow{1.0}}{\ehrlow{0.0}} & \ehrpair{\ehrlow{0.0}}{\ehrlow{0.0}} & \ehrpair{\ehrlow{0.4}}{\ehrlow{0.0}} & \ehrpair[21pt]{\textcolor{EHRDeltaPositive}{+0.3}}{\textcolor{EHRDeltaNegative}{-0.3}} \\
\rowcolor{EHRTrainingGreen!7}
\cellcolor{white} & SFT & \ehrpair{41.5}{35.4} & \ehrpair{54.5}{30.3} & \ehrpair{48.1}{39.2} & \ehrpair{69.6}{73.9} & \ehrpair{59.2}{26.2} & \ehrpair{49.1}{28.6} & \ehrpair{52.5}{31.8} & \ehrpair[21pt]{\textbf{\textcolor{EHRDeltaPositive}{+52.4}}}{\textcolor{EHRDeltaPositive}{+31.5}} \\
\rowcolor{EHRTrainingGreen!7}
\cellcolor{white} & \hspace*{0.5em}+DPO & \ehrpair{32.3}{26.2} & \ehrpair{15.2}{30.3} & \ehrpair{37.7}{48.1} & \ehrpair{60.9}{60.9} & \ehrpair{11.5}{59.7} & \ehrpair{31.4}{28.6} & \ehrpair{25.3}{45.2} & \ehrpair[21pt]{\textcolor{EHRDeltaPositive}{+25.2}}{\textbf{\textcolor{EHRDeltaPositive}{+44.9}}} \\
\rowcolor{EHRTrainingGreen!7}
\cellcolor{white} & \hspace*{0.5em}+PPO & \ehrpair{\ehrlow{0.0}}{\ehrlow{0.8}} & \ehrpair{9.1}{21.2} & \ehrpair{29.7}{27.8} & \ehrpair{\ehrlow{4.3}}{47.8} & \ehrpair{22.1}{32.1} & \ehrpair{45.5}{13.6} & \ehrpair{25.1}{23.1} & \ehrpair[21pt]{\textcolor{EHRDeltaPositive}{+25.0}}{\textcolor{EHRDeltaPositive}{+22.8}} \\
\rowcolor{EHRTrainingGreen!7}
\cellcolor{white}\multirow{-6}{*}{Llama-3.1-8B} & \hspace*{0.5em}+GRPO & \ehrpair{49.2}{53.8} & \ehrpair{54.5}{42.4} & \ehrpair{45.3}{58.5} & \ehrpair{87.0}{65.2} & \ehrpair{53.8}{30.8} & \ehrpair{39.5}{25.9} & \ehrpair{49.1}{39.7} & \ehrpair[21pt]{\textcolor{EHRDeltaPositive}{+49.0}}{\textcolor{EHRDeltaPositive}{+39.4}} \\
\bottomrule
\end{tabular}}
\endgroup
\ehrtablenote{\ehrclean{} / Noise success rates (Pass@1, \%, $\uparrow$). \textcolor[rgb]{0.93,0.96514,0.95471}{\rule{1.2em}{0.7em}}~indicates training results. \textcolor[RGB]{254,237,237}{\rule{1.2em}{0.7em}}~indicates rates below 5\%.}
\end{table}

\begin{figure}[t]
\centering
\includegraphics[width=0.980769\textwidth]{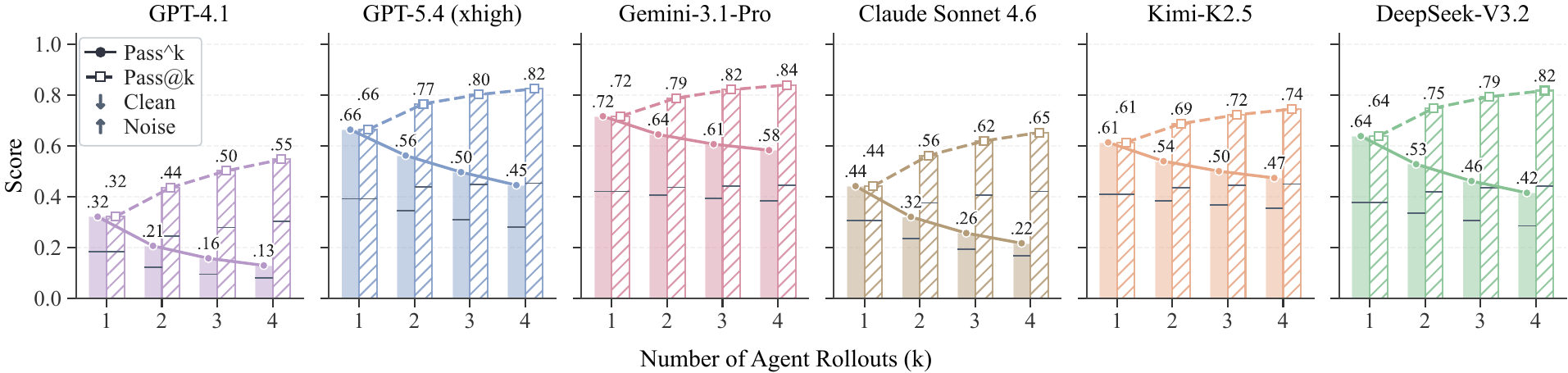}
\par\vspace{1pt}
{\small (a) Inference-Time Scaling}\par\vspace{7pt}
\begin{minipage}[t]{0.40\textwidth}
\centering
\includegraphics[width=0.955248\linewidth]{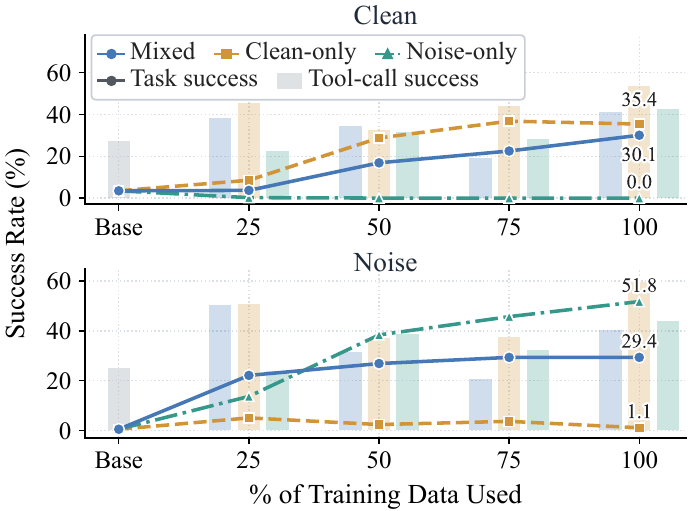}
\par\vspace{1pt}
{\small (b) Training-Data Scaling}
\end{minipage}\hfill
\begin{minipage}[t]{0.58\textwidth}
\centering
\includegraphics[width=0.979870\linewidth]{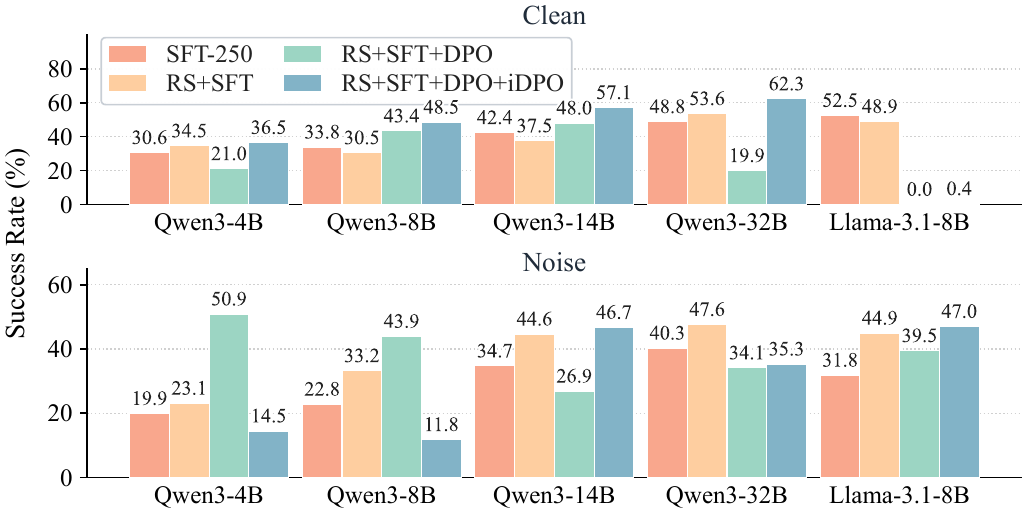}
\par\vspace{1pt}
{\small (c) Self-Improvement}
\end{minipage}
\caption{Scaling and Self-Improvement in \method{}.}
\label{fig:inference-scaling}
\end{figure}

\textbf{Agent Training.}
\method{} supports repeated sampling of successful and unsuccessful trajectories for agent training.
For SFT, we fine-tune models on verified successful teacher trajectories from both Clean and Noise tasks after removing teacher-only reference guidance.
Offline DPO~\citep{dpo} contrasts successful teacher trajectories with unsuccessful model trajectories for the same question.
We further consider online PPO~\citep{ppo} and GRPO~\citep{deepseekmath}, enabling agents to actively explore tasks and generate new training trajectories through environment interaction.
Both methods use the same success score $R(x,\tau)$ for optimization, with separate KL regularization and no intermediate or format rewards.
We explore self-improvement by refining each model on its own trajectories using RS+SFT and iterative DPO.
PE provides a prompt-only baseline by adding evidence-check instructions without parameter updates.
Training settings and prompts are provided in Section~\ref{sec:experiments} and Appendices~\ref{app:prompts} and~\ref{app:optimization-background}.

\section{Experiments}
\label{sec:experiments}
We benchmark EHR agents under Clean and Noise conditions, then evaluate prompting and trajectory-based training with \method{}. Analyses examine evidence-grounded abstention, inference-time and training-data scaling, capability retention, and tool interaction errors.

\subsection{Experimental Setup}

\textbf{Tasks, Models and Implementation Details.}
We evaluate seven models through external APIs and five locally deployed open-weight models. The API-based models include GPT-5.4 (xhigh)~\citep{gpt54}, GPT-4.1 (\texttt{gpt-4.1-2025-04-14})~\citep{gpt41}, Gemini-3.1-Pro (\texttt{gemini-3.1-pro-preview})~\citep{gemini31}, Claude Sonnet 4.6~\citep{claudesonnet46}, Kimi-K2.5~\citep{kimi25}, DeepSeek-V3.2 (\texttt{DeepSeek-V3.2-Exp})~\citep{deepseekv32exp}, and Qwen3-235B-A22B (\texttt{Qwen3-235B-A22B-Instruct-2507})~\citep{qwen3_235,qwen3}. For parameter training, we use Qwen3-4B, Qwen3-8B, Qwen3-14B, Qwen3-32B, and Llama-3.1-8B (\texttt{Llama-3.1-8B-Instruct})~\citep{qwen3,llama3}. All models are evaluated on the same 1,008 Clean-Noise test pairs across six clinical intents. The training partition contains 4,478 pairs. All models interact with the same environment described in Section~\ref{sec:method}. Prompts and implementation details appear in Appendices~\ref{app:prompts} and~\ref{app:experimental-details}.

\textbf{Evaluation Metrics.}
\label{sec:evaluation-protocol}
We report \emph{Task Success Rate} (SR) separately for Clean and Noise requests, using the criterion in Equation~\ref{eq:terminal-reward}. GPT-5.2 (high)~\citep{gpt52} performs this verification. Evaluation and online RL use the same scorer settings. Appendix~\ref{app:scoring-and-diagnosis} provides the scoring criteria.

\subsection{Evaluation Results}

\textbf{Overall Performance.}
Table~\ref{tab:main-results} reports Clean and Noise success rates for seven unadapted API-based models. All seven perform worse on Noise requests, with average success dropping from 62.2\% to 37.9\%. Gemini and Kimi achieve similar Clean success rates of 83.4\% and 82.4\%, but differ under Noise, reaching 58.4\% and 37.1\%. These results suggest that evidence-grounded reasoning remains challenging when supporting records are missing or inconsistent with the request. Previous evaluations focused on answerable tasks do not adequately capture this capability.

\textbf{Noise Types and Evidence Verification.}
Query-level Noise is the most challenging for all seven Base models. GPT-5.4 achieves 43.2\% success, compared with 58.3\% on Record-level and 65.8\% on Value-level Noise, suggesting difficulty in verifying relations among records (Appendix~\ref{app:per-type}). Abstention alone does not imply task success. GPT-4.1 returns \texttt{NULL} on 80.6\% of Noise requests, but only 35.3\% of these responses pass evidence verification, yielding 28.5\% success.

\textbf{Clinical Intent and Scope.}
On Clean requests, GPT-5.4 performs best on Cost and Labs, at 95.7\% and 91.3\%. Medications and Vitals show lower success, at 63.2\% and 63.6\%, and involve reasoning over medication states and ordered observations. Moving from patient-level to population-level requests, its Noise success drops from 65.5\% to 46.0\%. Population-level requests require agents to verify cohort eligibility and reporting support. Full results are reported in Table~\ref{tab:scope-track-results}.

\subsection{Training EHR Agents with EHR-RobustGym}

\textbf{Training Setup.}
We use trajectory-based training to jointly strengthen task completion and evidence-grounded abstention. To assess training effectiveness across model scales and families, we select four Qwen3 backbones (4B, 8B, 14B, and 32B) and Llama-3.1-8B. We randomly sample 1,000 Clean-Noise pairs from the training set, comprising 2,000 questions, and adopt a two-stage fine-tuning framework. Models first learn from successful trajectories through SFT for 250 updates (half an epoch). For each backbone, DPO, PPO, and GRPO independently continue from the same SFT checkpoint for another 250 updates using the remaining 1,000 questions. Offline DPO contrasts successful teacher trajectories with unsuccessful model trajectories for the same question, while online PPO and GRPO sample eight trajectories per question through environment interaction. We include PE as a prompt-only baseline for both the five backbones and seven API-based models. Full training configurations and additional experiments are provided in Appendix~\ref{app:training-details}.

\begin{figure}[t]
\centering
\includegraphics[width=0.978722\textwidth]{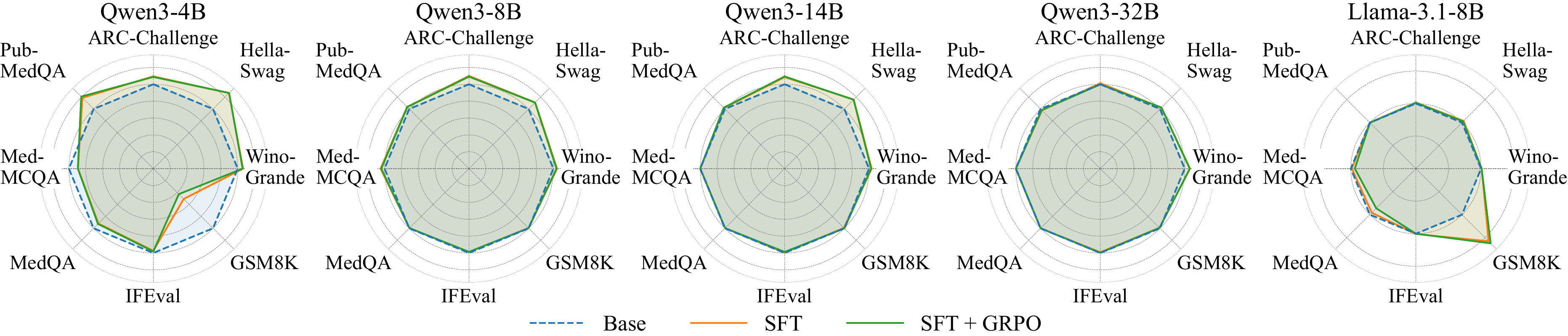}
\caption{General and Medical Capabilities after EHR Training.}
\label{fig:alignment-tax}
\end{figure}

\begin{figure}[t]
\centering
\includegraphics[width=0.977456\textwidth]{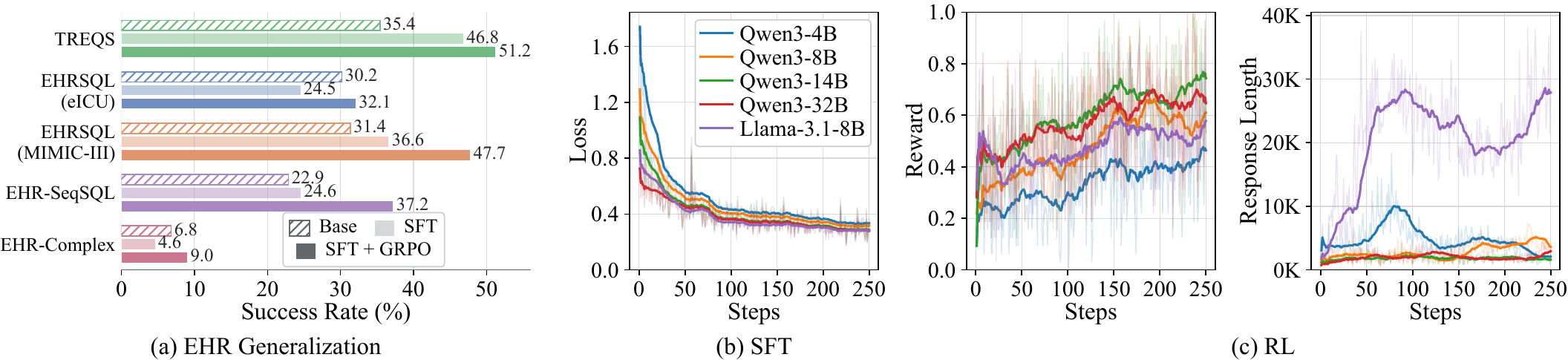}
\caption{External EHR Generalization and Training Dynamics. (a) Qwen3-14B success rates on five external EHR benchmarks. (b) SFT loss and (c) subsequent GRPO reward and response length.}
\label{fig:training-curves}
\end{figure}

\textbf{Training Results.}
Table~\ref{tab:mitigation-results} shows that SFT on successful trajectories improves Clean and Noise success across all five backbones. Qwen3-32B improves from 6.0\%/0.6\% to 48.8\%/40.3\% Clean/Noise success, while Llama-3.1-8B improves from near-zero success to 52.5\%/31.8\%. By contrast, PE yields little improvement on smaller models and mixed results on API-based models. The SFT gains demonstrate the effectiveness of trajectory supervision for task completion and evidence-grounded abstention across model scales and families. Subsequent online RL yields the strongest results on the Qwen3 backbones, with GRPO performing best on the 4B and 8B models and PPO on the 14B and 32B models. PPO raises Qwen3-14B's Clean/Noise success from 42.4\%/34.7\% to 70.4\%/62.6\% with outcome feedback from \method{}. DPO also improves both conditions for Qwen3-14B and Qwen3-32B but remains below PPO. However, subsequent optimization does not consistently improve both conditions. On Qwen3-32B, PPO and GRPO both reach 62.8\% Clean success, but their Noise scores are 58.5\% and 37.7\%, with GRPO falling below SFT. On Llama-3.1-8B, DPO raises Noise success to 45.2\% but reduces Clean success to 25.3\%, while PPO decreases both. Figure~\ref{fig:training-curves}(b,c) presents the training dynamics of SFT and subsequent GRPO.

\textbf{Inference-Time Scaling.}
We study whether models can consistently complete tasks across repeated rollouts. Pass@$k$ measures the fraction of questions solved by at least one of $k$ attempts, while pass\textasciicircum{}k requires all $k$ attempts to succeed. Figure~\ref{fig:inference-scaling}(a) shows that GPT-5.4 reaches 82.4\% Pass@4 but only 44.5\% pass\textasciicircum{}4, while Gemini achieves 83.9\% and 58.3\%, respectively. These results show that additional sampling improves task coverage, but even strong models struggle to complete the same tasks consistently. Sampling details and per-condition results are provided in Appendix~\ref{app:repeated-evaluations}.

\begin{figure}[t]
\centering
\begin{minipage}[t]{0.60\textwidth}
\centering
\includegraphics[width=0.97281\linewidth]{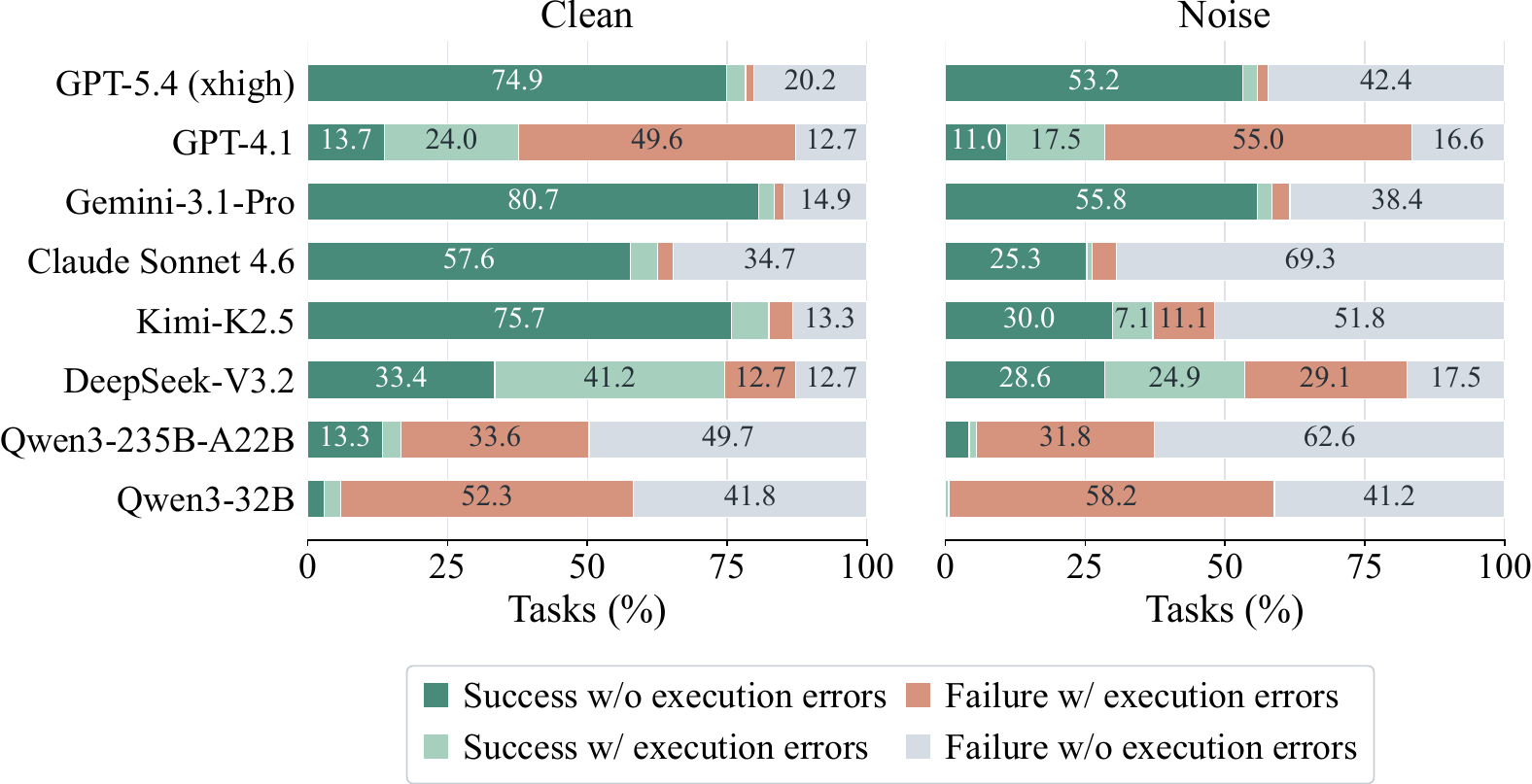}
\par
{\small (a) Tool Execution Errors and Task Outcomes}
\end{minipage}\hfill
\begin{minipage}[t]{0.38\textwidth}
\centering
\includegraphics[width=0.98068\linewidth]{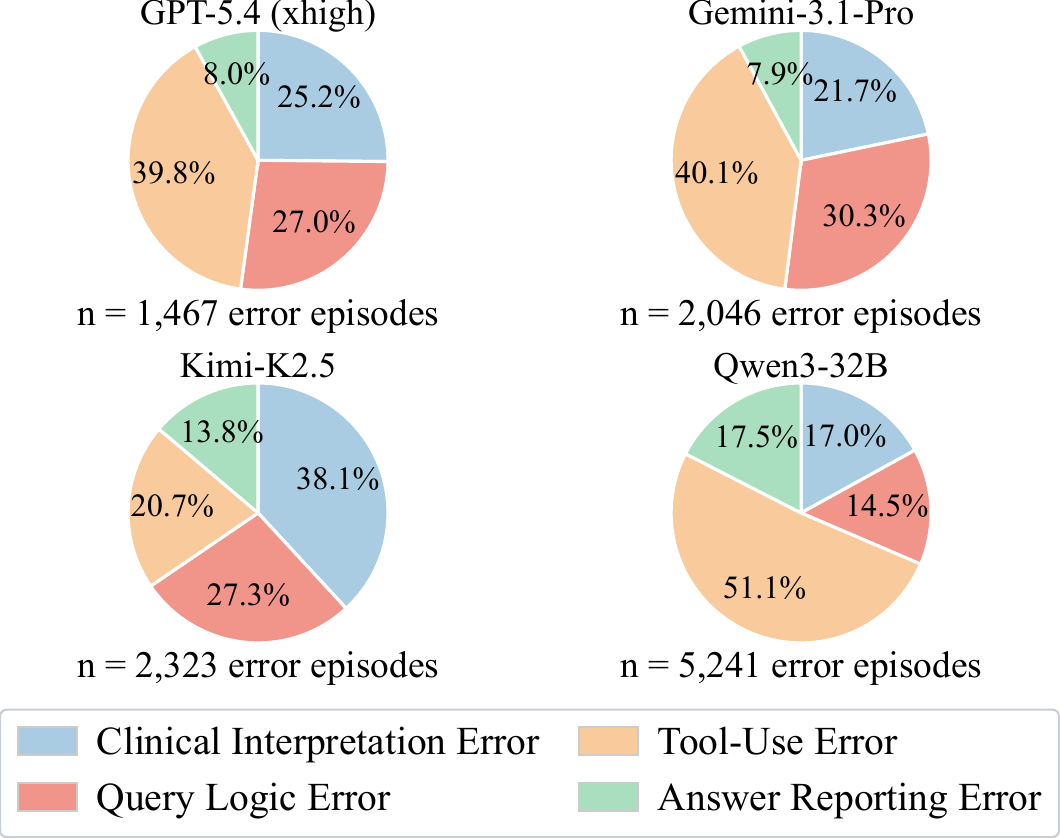}
\par
{\small (b) Distribution of Error Episodes}
\end{minipage}
\caption{Trajectory Error Analysis.}
\label{fig:tool-recovery}
\end{figure}

\textbf{Training-Data Scaling and Composition.}
Figure~\ref{fig:inference-scaling}(b) compares one-epoch SFT on Qwen3-8B with 250-1,000 trajectories, using one trajectory per pair across Clean-only, Noise-only, and Mixed training.
At 1,000 trajectories, Mixed training achieves 30.1\%/29.4\% Clean/Noise success, while single-condition training yields near-zero success on the other condition.
With both trajectories per pair, increasing Mixed training from 500 to 2,000 trajectories raises Clean/Noise success from 16.9\%/21.9\% to 39.8\%/42.8\% (Appendix~\ref{app:sft-scaling}). At 1,000 trajectories, Mixed training improves tool-call success while roughly halving calls per question. However, Clean-only training achieves 59.6\% tool-call success on Noise questions but only 1.1\% task success.
Increasing Mixed training from 500 to 750 trajectories lowers tool-call success despite higher task success on both conditions.
Successful tool execution alone does not ensure correct answers or evidence-grounded abstention.

\textbf{Self-Improvement with Model-Generated Trajectories.}
We explore self-improvement by refining each model using its own outputs. Starting from SFT-250, we perform RS+SFT followed by DPO and iDPO for one epoch each, selecting training trajectories from eight fresh rollouts per question at every stage. Figure~\ref{fig:inference-scaling}(c) shows that RS+SFT improves Noise success for all five models, while Clean success improves only for Qwen3-4B and Qwen3-32B. Subsequent DPO stages yield less consistent gains across models. After the full sequence, Qwen3-14B improves from 42.4\%/34.7\% to 57.1\%/46.7\% Clean/Noise success, while Qwen3-8B and Qwen3-32B finish below their SFT Noise scores. The final iDPO set for Llama-3.1-8B is highly imbalanced, containing only three Clean questions and 241 Noise questions. Further details are provided in Appendix~\ref{app:self-improvement}.

\textbf{General Capabilities and EHR Generalization.}
Across eight general and medical benchmarks (Figure~\ref{fig:alignment-tax}), four of five models improve their average over Base after SFT and remain above Base after GRPO. Qwen3-4B is the exception, with GSM8K success falling from 78.9\% to 33.7\% after SFT and subsequent GRPO. Full results are provided in Appendix~\ref{app:capability-comparisons}. Figure~\ref{fig:training-curves}(a) further evaluates Qwen3-14B on five external EHR benchmarks without target-specific fine-tuning. After GRPO, the model outperforms both Base and SFT on all five benchmarks. Relative to Base, success rises from 35.4\% to 51.2\% on TREQS~\citep{wang2020text}, with gains extending to the challenging EHR-Complex~\citep{ehrcomplex}, from 6.8\% to 9.0\%. Although absolute success on EHR-Complex remains low, improvements across all five benchmarks support the effectiveness of \method{} as a training environment for EHR reasoning that transfers to other benchmarks.

\textbf{Error Analysis.}
Figure~\ref{fig:tool-recovery}(a) groups Base trajectories by task outcome and the presence of tool execution errors, with timeouts and response-format errors counted separately. For GPT-5.4, failures without tool execution errors account for 42.4\% of Noise requests, compared with 20.2\% of Clean requests. This suggests that tool execution errors alone do not explain the additional difficulty under Noise. We use LLM-based annotation to analyze 12,096 Base trajectories from GPT-5.4, Gemini-3.1-Pro, Claude Sonnet 4.6, Kimi-K2.5, DeepSeek-V3.2, and Qwen3-32B, covering both successful and failed trajectories under Clean and Noise conditions. Figure~\ref{fig:tool-recovery}(b) summarizes error episode distributions across four categories for four models; Claude and DeepSeek appear in Appendix~\ref{app:error-taxonomy}. A trajectory can contain multiple error episodes, including those later corrected. For GPT-5.4, clinical interpretation and query logic errors account for 25.2\% and 27.0\% of annotated error episodes, respectively. Representative trajectories show errors in interpreting procedure counts, preserving event order, and selecting the requested hospitalization (Appendix~\ref{app:examples}). These cases show that agents can retrieve relevant records yet fail to preserve clinical constraints when using the evidence.

\section{Conclusion}
\label{sec:conclusion}
We presented \ehrrobust{}, a scalable and interactive environment for evaluating and training robust clinical agents grounded in noisy EHRs. With Clean-Noise pairs spanning Record-level, Value-level, and Query-level noise, \ehrrobust{} jointly evaluates accurate answering on supported questions and evidence-grounded abstention on unsupported requests. Our evaluation reveals substantial robustness gaps between Clean and Noise questions and unstable clinical task completion across repeated executions. Supervised fine-tuning and reinforcement learning improve performance, with gains generalizing to five external EHR benchmarks. Together, these findings position \ehrrobust{} as a testbed for evaluating and improving the evidence-grounded robustness of clinical agents, while highlighting the need to preserve accurate answering on supported questions.

\section*{AI Use Statement}

AI assistants were used to support language polishing. The use of language models in task construction, trajectory generation, outcome verification, and error analysis is described in the corresponding methodology and experimental sections. All scientific claims, experimental results, analyses, citations, and final manuscript content were reviewed and verified by the authors, who take full responsibility for the work.

\section*{Ethics Statement}

\ehrrobust{} is constructed from MIMIC-IV v3.1, a de-identified EHR database released through PhysioNet for research under credentialed access~\citep{mimiciv31}. The underlying data have undergone privacy-preserving processing, including removal of protected health information, random identifier replacement, and patient-level date shifting~\citep{mimiciv_original}. Our study uses these retrospective records for research on evidence-grounded clinical agents. Access to the underlying records remains subject to the original MIMIC-IV credentialing and data use requirements. Any released artifacts, including task definitions, SQL programs, trajectories, model checkpoints, and code, will be shared only in forms compatible with these requirements, with appropriate access controls for restricted derivatives. \ehrrobust{} is intended as a research testbed for evaluating and training clinical agents, not as a clinical decision-support system or a substitute for professional clinical judgment.

\section*{Reproducibility Statement}

Section~\ref{sec:formulation} defines the clinical tasks, the three noise types, and the construction of Clean-Noise pairs, together with the interactive environment and outcome verification used for evaluation and training. Appendix~\ref{app:statistics} details the construction programs, reference generation, question rewriting, and data splits, including the separation of patient-level tasks and the construction programs reserved for structure-disjoint evaluation. Appendix~\ref{app:human-validation} documents the selection of challenging test pairs, human review criteria, and adjudication procedure.

Section~\ref{sec:experiments} describes the evaluated models, metrics, and training comparisons. Appendix~\ref{app:experimental-details} specifies model identifiers, sampling settings, interaction limits, and outcome evaluation criteria for accurate answering and evidence-grounded abstention. It also describes the datasets and protocols used to assess general capabilities and transfer to external EHR benchmarks. Appendix~\ref{app:training-details} provides the demonstration selection procedure, training-data allocation, checkpoint initialization, and optimization settings for SFT and subsequent DPO, PPO, and GRPO, along with the trajectory selection and training schedule for self-improvement. Appendix~\ref{app:additional-results} reports additional comparisons and per-benchmark scores, while Appendix~\ref{app:repeated-evaluations} specifies the repeated-run protocol and the computation of Pass@$k$ and pass\textasciicircum{}k. The error taxonomy and trajectory annotation procedure are provided in Appendix~\ref{app:error-taxonomy}, and the shared agent instructions and response format appear in Appendix~\ref{app:prompts}.

\bibliography{references}
\bibliographystyle{iclr2027_conference}

\clearpage
\appendix
\addtocontents{toc}{\protect\setcounter{tocdepth}{2}}
\ehrappendixcontents
\newcommand{\ehrappgroup}[2]{%
  \rowcolor{EHRTablePurplePale}[0pt][0pt]
  \multicolumn{#1}{@{\hspace{4pt}}l@{\hspace{4pt}}}{\rule{0pt}{2.3ex}\textcolor{EHRTablePurple}{\textbf{#2}}}\\}
\newcommand{\ehrappbar}[2]{%
  \leavevmode\rlap{\raisebox{-0.35ex}{\textcolor{EHRTablePurpleSoft}{\rule{#1\linewidth}{1.9ex}}}}%
  \makebox[\linewidth][r]{\strut #2}}

\section{Dataset Details}
\label{app:statistics}

\subsection{Database Preprocessing}
\label{app:database-preprocessing}

We rename database tables and columns to reduce reliance on memorized MIMIC-IV schemas. Descriptive table names and consistent shared identifiers also help clarify the links between patient, admission, and ICU records. Table~\ref{tab:database-schema-mapping} lists every table and a representative column mapping. Shared identifiers include \texttt{patientid}, \texttt{admissionid}, and \texttt{icuadmissionid}.
The conversion retains all 342 source columns, their values, data types, and temporal precision. The only derived field is admission \texttt{age}, computed as \texttt{anchor\_age} plus the admission year minus \texttt{anchor\_year}. Fields absent from MIMIC-IV~3.1, such as date of birth or diagnosis timestamps, are not reconstructed. Source and converted tables are checked for matching row counts and contents. Both conditions use the same converted database; Noise construction changes question conditions rather than stored records.

\begingroup
\small
\setlength{\tabcolsep}{3pt}
\renewcommand{\arraystretch}{1.12}
\begin{longtable}{@{\hspace{4pt}}>{\raggedright\arraybackslash}p{0.20\textwidth}>{\raggedright\arraybackslash}p{0.33\textwidth}>{\raggedright\arraybackslash}p{0.42\textwidth}@{\hspace{4pt}}}
\caption{Table and column renaming for MIMIC-IV~3.1. All 31 tables are listed; the final column shows one example per table.}
\label{tab:database-schema-mapping}\\
\toprule
\textbf{Source table} & \textbf{Renamed table} & \textbf{Example column mapping} \\
\midrule
\endfirsthead
\multicolumn{3}{c}{\tablename\ \thetable\ (continued)}\\
\toprule
\textbf{Source table} & \textbf{Renamed table} & \textbf{Example column mapping} \\
\midrule
\endhead
\bottomrule
\endfoot
\ehrappgroup{3}{Hospital tables (22)}
admissions & hospitaladmissions & admittime $\rightarrow$ admitdatetime \\
d\_hcpcs & billingprocedurecodes & code $\rightarrow$ billingcode \\
diagnoses\_icd & admissiondiagnoses & icd\_version $\rightarrow$ codeversion \\
d\_icd\_diagnoses & diagnosiscodes & icd\_code $\rightarrow$ icdcode \\
d\_icd\_procedures & procedurecodes & long\_title $\rightarrow$ description \\
d\_labitems & labtesttypes & label $\rightarrow$ itemname \\
drgcodes & diagnosisrelatedgroups & drg\_type $\rightarrow$ groupingsystem \\
emar\_detail & medicationadministrationdetails & dose\_given $\rightarrow$ dosegiven \\
emar & medicationadministrations & charttime $\rightarrow$ administrationdatetime \\
hcpcsevents & billingprocedureevents & hcpcs\_cd $\rightarrow$ billingcode \\
labevents & labresults & valuenum $\rightarrow$ resultvalue \\
microbiologyevents & microbiologyresults & org\_name $\rightarrow$ organismname \\
omr & outpatientmeasurements & result\_name $\rightarrow$ measurementname \\
patients & demographics & anchor\_age $\rightarrow$ anchorage \\
pharmacy & pharmacyorderdetails & medication $\rightarrow$ medicationname \\
poe\_detail & providerorderdetails & field\_name $\rightarrow$ fieldname \\
poe & providerorders & order\_type $\rightarrow$ ordertype \\
prescriptions & medicationorders & drug $\rightarrow$ medicationname \\
procedures\_icd & admissionprocedures & chartdate $\rightarrow$ proceduredate \\
provider & providers & provider\_id $\rightarrow$ providerid \\
services & hospitalservices & curr\_service $\rightarrow$ currentservice \\
transfers & patienttransfers & eventtype $\rightarrow$ transfertype \\
\ehrappgroup{3}{ICU tables (9)}
caregiver & caregivers & caregiver\_id $\rightarrow$ caregiverid \\
chartevents & clinicalevents & value $\rightarrow$ textvalue \\
datetimeevents & clinicaldatetimeevents & value $\rightarrow$ datetimevalue \\
d\_items & clinicalitemtypes & linksto $\rightarrow$ itemtype \\
icustays & icuepisodes & first\_careunit $\rightarrow$ initialcareunit \\
ingredientevents & infusioningredients & amount $\rightarrow$ administeredingredientamount \\
inputevents & intakerecords & amount $\rightarrow$ administeredamount \\
outputevents & outputrecords & value $\rightarrow$ volume \\
procedureevents & clinicalprocedures & value $\rightarrow$ procedureduration \\
\end{longtable}
\endgroup

\subsection{Dataset Composition}

Tables~\ref{tab:evidence-programs} and~\ref{tab:medical-categories} summarize \method{} by query scope, noise type, and clinical intent. The benchmark contains 4,478 training pairs and 1,008 test pairs, with one Clean and one Noise question in each pair. Its 51 construction programs cover six primary clinical intents. Query scope and noise type define the sampling strata, whereas clinical intents describe task coverage. Program counts in the All column are distinct unions of the training and test programs rather than sums. Each pair has one primary clinical intent; background bars in Table~\ref{tab:medical-categories} show relative pair counts within that table.

\begingroup
\newcommand{\ehrdatasetbar}[2]{%
  \leavevmode\rlap{\raisebox{-0.35ex}{\textcolor{EHRTablePurpleSoft}{\rule{#1\linewidth}{1.9ex}}}}%
  \makebox[\linewidth][r]{\strut #2}%
}
\begin{table}[!htbp]
\caption{Composition of \method{} by query scope and noise type.}
\label{tab:evidence-programs}
\centering
\small
\setlength{\tabcolsep}{4pt}
\renewcommand{\arraystretch}{1.24}
\begin{tabularx}{\textwidth}{@{\hspace{4pt}}>{\raggedright\arraybackslash}X
  *{2}{>{\raggedleft\arraybackslash}p{0.066\textwidth}}
  >{\raggedleft\arraybackslash}p{0.10\textwidth}
  @{\hspace{1.5em}}
  *{3}{>{\raggedleft\arraybackslash}p{0.053\textwidth}}@{\hspace{4pt}}}
\toprule
& \multicolumn{3}{c}{\textbf{Question pairs}} & \multicolumn{3}{c}{\textbf{Programs}} \\
\cmidrule(lr){2-4}\cmidrule(l){5-7}
\textbf{Noise type} & Train & Test & \textbf{All} & Train & Test & All \\
\midrule
\rowcolor{EHRTablePurplePale}[0pt][0pt]
\multicolumn{7}{@{\hspace{4pt}}l@{\hspace{4pt}}}{\strut\textcolor{EHRTablePurple}{\textbf{Patient-level queries}}}\\
Record-level & 750 & 168 & \textbf{918} & 3 & 4 & 4 \\
Value-level & 750 & 168 & \textbf{918} & 3 & 4 & 4 \\
Query-level & 750 & 168 & \textbf{918} & 6 & 7 & 7 \\
\addlinespace[0.35em]
\rowcolor{EHRTablePurplePale}[0pt][0pt]
\multicolumn{7}{@{\hspace{4pt}}l@{\hspace{4pt}}}{\strut\textcolor{EHRTablePurple}{\textbf{Population-level queries}}}\\
Record-level & 749 & 168 & \textbf{917} & 10 & 11 & 11 \\
Value-level & 747 & 168 & \textbf{915} & 12 & 13 & 13 \\
Query-level & 732 & 168 & \textbf{900} & 11 & 12 & 12 \\
\midrule
\textbf{Total} & \textbf{4,478} & \textbf{1,008} & \textbf{5,486} & \textbf{45} & \textbf{51} & \textbf{51} \\
\bottomrule
\end{tabularx}
\end{table}

\begin{table}[!htbp]
\caption{Coverage of the six primary clinical intents.}
\label{tab:medical-categories}
\centering
\small
\setlength{\tabcolsep}{4pt}
\renewcommand{\arraystretch}{1.24}
\begin{tabularx}{\textwidth}{@{\hspace{4pt}}>{\raggedright\arraybackslash}X
  *{2}{>{\raggedleft\arraybackslash}p{0.066\textwidth}}
  >{\raggedleft\arraybackslash}p{0.10\textwidth}
  @{\hspace{1.5em}}
  *{3}{>{\raggedleft\arraybackslash}p{0.053\textwidth}}@{\hspace{4pt}}}
\toprule
& \multicolumn{3}{c}{\textbf{Question pairs}} & \multicolumn{3}{c}{\textbf{Programs}} \\
\cmidrule(lr){2-4}\cmidrule(l){5-7}
\textbf{Clinical intent} & Train & Test & \textbf{All} & Train & Test & All \\
\midrule
Demographics \& Tracking & 392 & 130 & \ehrdatasetbar{0.259057}{522} & 6 & 7 & 7 \\
Vital Signs \& Observations & 214 & 33 & \ehrdatasetbar{0.122581}{247} & 4 & 4 & 4 \\
Medications \& Orders & 1,465 & 212 & \ehrdatasetbar{0.832258}{1,677} & 10 & 11 & 11 \\
Cost \& Billing & 203 & 23 & \ehrdatasetbar{0.112159}{226} & 3 & 3 & 3 \\
Labs \& Microbiology & 1,625 & 390 & \ehrdatasetbar{1.000000}{2,015} & 18 & 20 & 20 \\
Diagnoses \& Procedures & 579 & 220 & \ehrdatasetbar{0.396526}{799} & 4 & 6 & 6 \\
\midrule
\textbf{Total} & \textbf{4,478} & \textbf{1,008} & \textbf{5,486} & \textbf{45} & \textbf{51} & \textbf{51} \\
\bottomrule
\end{tabularx}
\end{table}

The six primary clinical intents follow the naming used in EHR-Complex~\citep{ehrcomplex}. We assign each pair a primary intent based on its clinical information need and requested output. Cross-domain requests retain their additional output categories. In the current corpus, 94 laboratory/microbiology--medication pairs have two output categories, while all other pairs have one. ICU datetime events concerning devices, airways, procedures, or wound care are assigned the primary intent Diagnoses \& Procedures rather than Vital Signs \& Observations.

\endgroup

\subsection{Construction Details}
\label{app:construction-algorithm}

\textbf{Semantics and condition substitution.}
We use Kimi-K2.5~\citep{kimi25} to construct clinical requests, Clean-Noise pairs, and reference SQL. Construction programs deterministically instantiate the requests and execute their reference queries. Each pair changes one necessary condition while preserving the clinical topic, query scope, requested output, and other conditions. The noise label identifies the support condition that fails, not merely the type of locator changed. For example, changing an eMAR event ordinal yields Value-level Noise when the event exists but its state does not support an administered dose. Selecting an event before checking its status differs from selecting among events already satisfying that status. Population queries likewise compute within each eligible unit before aggregating across units, rather than pooling all measurement rows.

\textbf{Clinical semantics.}
Blood-culture tasks jointly interpret specimen source and assay identity, excluding non-blood fluids in culture bottles, serological assays, and stem-cell product cultures. ICU intake tasks combine codes for the same named therapy before computing within-stay summaries. A recorded-unit eligibility restriction is stated explicitly and is not implied by a requested reporting unit. Medication doses require a compatible administration state. Dose references normalize unit aliases and aggregate original product-component rows within the same administration event and unit before clinical joins. Distinct product rows with equal doses both contribute. Non-numeric dose expressions require explicit interpretation rather than being silently omitted.

\textbf{Reference generation.}
A deterministic renderer converts execution results into Clean reference values, units, and record context. Noise references identify an unsupported necessary condition and distinguish it from related evidence. Population references distinguish zero eligible units from one to four. Both require \texttt{NULL}, and the latter withhold exact counts and aggregates. Lack of support is relative to the requested records and reporting rules, not absolute nonexistence in the real world. Evaluation also accepts another sufficient reason established by the agent's visible evidence.

\textbf{Question rewriting and review.}
Kimi-K2.5 rewrites both canonical questions together using the locked evidence contract, selected substitution, required locators, and permitted language exemplars. Rewriting preserves record identities, linked conditions, data sources, units, temporal direction, and the order of filtering and ordinal selection. Exemplars supply phrasing, not clinical facts. Automatic checks reject information leakage, locator or linkage errors, altered operation order, excessive length, copied exemplar wording, and duplicate questions. Kimi-K2.5 separately reviews and repairs wording against the same contract, after which checks are repeated.

\subsection{Question Vocabulary Visualization}

\begin{figure}[!htbp]
\centering
\includegraphics[width=0.983151\textwidth]{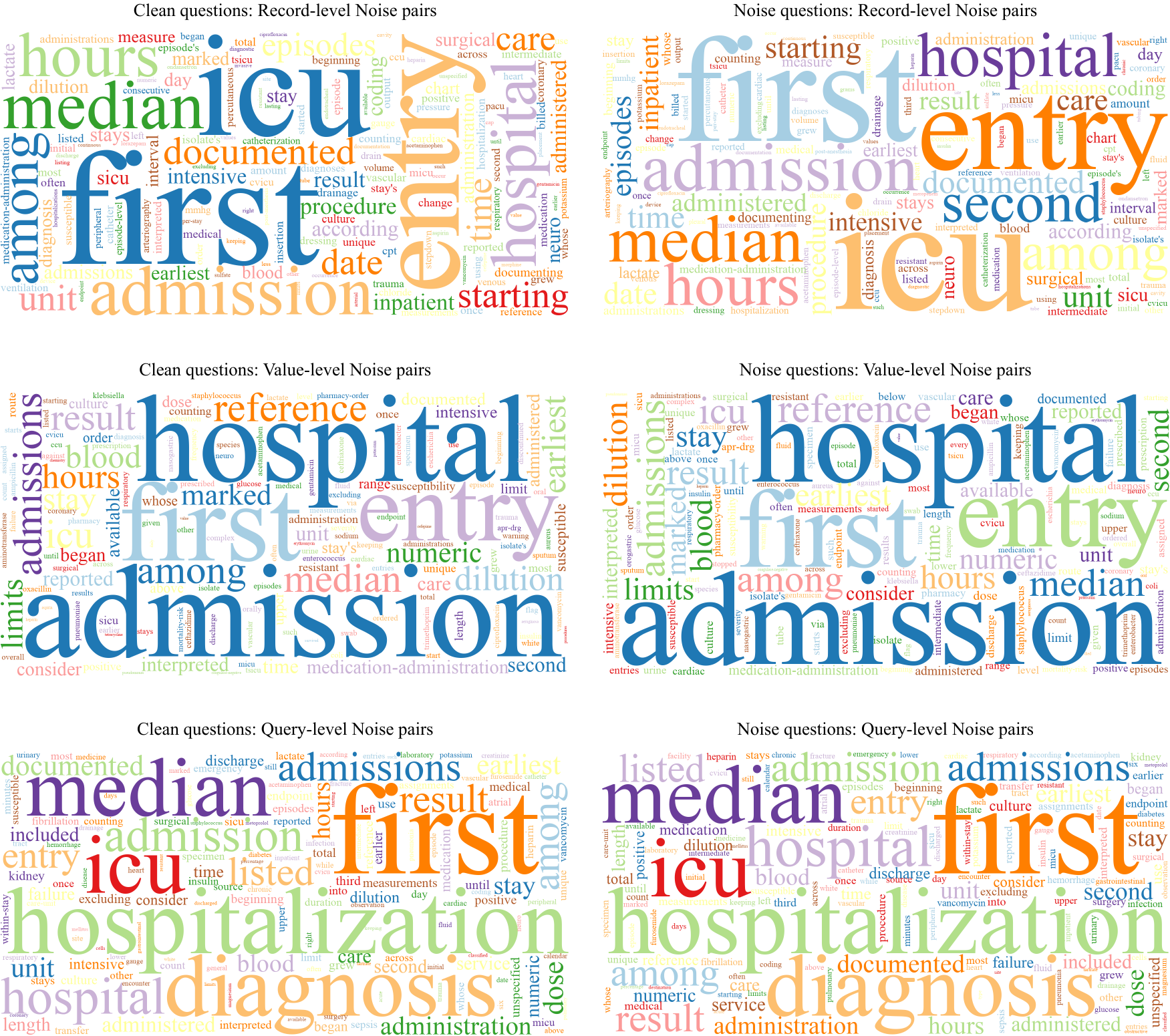}
\caption{Question vocabulary by noise type. Rows show Record-level, Value-level, and Query-level pairs. Columns show \ehrclean{} and Noise questions from the final training and test sets. Word size indicates frequency within each panel.}
\label{fig:question-wordcloud}
\end{figure}

Figure~\ref{fig:question-wordcloud} summarizes the final 4,478 training and 1,008 test pairs, pooling the two splits but keeping conditions and noise types separate. We lowercase the question text, retain alphabetic tokens (including internal apostrophes and hyphens), and remove tokens of at most two letters and a fixed stopword list. Numeric identifiers, numerical values, and numeric date components therefore do not contribute. We use only questions, not SQL, reference answers, or teacher instructions. Each panel displays up to 160 words with its own frequency scaling and random seed 42. The visualization describes vocabulary rather than demonstrating absence of leakage or equal difficulty.

\subsection{Data Splits, Provenance, and Validation}

\textbf{Split design.}
The test set contains 504 in-domain and 504 structure-disjoint pairs. Each track has 252 patient-level and 252 population-level pairs, with 168 pairs per noise type. The in-domain track uses training construction programs, whereas the structure-disjoint track reserves six programs, one per query-scope--noise-type stratum, with 84 pairs each (Table~\ref{tab:held-out-programs}). The latter also separates program-level information-need IDs, surface-form families, SQL structures without literal values, and semantic contracts from training. These IDs are not the six clinical intents. Patient-level questions are patient-disjoint. Population questions use disjoint query contracts, but contributing patients may overlap. We check exact reference SQL and final questions across partitions and keep pairs intact.

\begin{table}[!htbp]
\caption{Construction programs reserved for the structure-disjoint test track. Construction tasks appear below their clinical intents.}
\label{tab:held-out-programs}
\centering
\small
\setlength{\tabcolsep}{5pt}
\renewcommand{\arraystretch}{1.18}
\begin{tabularx}{\textwidth}{@{\hspace{4pt}}p{0.16\textwidth}>{\raggedright\arraybackslash}X r@{\hspace{4pt}}}
\toprule
\textbf{Noise type} & \textbf{Clinical intent / construction task} & \textbf{Pairs}\\
\midrule
\rowcolor{EHRTablePurplePale}[0pt][0pt]
\multicolumn{3}{@{\hspace{4pt}}l@{\hspace{4pt}}}{\strut\textcolor{EHRTablePurple}{\textbf{Patient-level queries}}}\\
Record & Diagnoses \& Procedures & 84\\
\multicolumn{3}{@{\hspace{4pt}}l@{\hspace{4pt}}}{\footnotesize\textcolor{black!65}{Hospital admission order}}\\
\addlinespace[0.35em]
Value & Labs \& Microbiology & 84\\
\multicolumn{3}{@{\hspace{4pt}}l@{\hspace{4pt}}}{\footnotesize\textcolor{black!65}{Initial ICU care unit}}\\
\addlinespace[0.35em]
Query & Labs \& Microbiology & 84\\
\multicolumn{3}{@{\hspace{4pt}}l@{\hspace{4pt}}}{\footnotesize\textcolor{black!65}{ICU unit and lactate across admissions}}\\
\addlinespace[0.35em]
\rowcolor{EHRTablePurplePale}[0pt][0pt]
\multicolumn{3}{@{\hspace{4pt}}l@{\hspace{4pt}}}{\strut\textcolor{EHRTablePurple}{\textbf{Population-level queries}}}\\
Record & Diagnoses \& Procedures & 84\\
\multicolumn{3}{@{\hspace{4pt}}l@{\hspace{4pt}}}{\footnotesize\textcolor{black!65}{ICU procedure records}}\\
\addlinespace[0.35em]
Value & Medications \& Orders & 84\\
\multicolumn{3}{@{\hspace{4pt}}l@{\hspace{4pt}}}{\footnotesize\textcolor{black!65}{Medication event status}}\\
\addlinespace[0.35em]
Query & Demographics \& Tracking & 84\\
\multicolumn{3}{@{\hspace{4pt}}l@{\hspace{4pt}}}{\footnotesize\textcolor{black!65}{Diagnosis and admission-source conditions}}\\
\bottomrule
\end{tabularx}
\end{table}

\textbf{Sources and validation.}
The 51 construction programs draw on 94 distinct DiSCQ chart-review question sources and 35 EHRSQL information-need templates. These mappings motivate clinical information needs rather than independently validate every generated question. Compound queries are executable MIMIC-IV operationalizations, not requests claimed to appear verbatim in either source. Reference queries and results are checked against MIMIC-IV, alongside partition and provenance checks. Split restrictions also govern rewriting exemplars, and revisions to clinical scope or reference semantics require re-execution.

\section{Human Validation of Task Quality}
\label{app:human-validation}

Task construction in \ehrrobust{} is grounded in database evidence and execution-verified reference outcomes.
Each Clean question is instantiated from observed records and paired with a reference SQL query and its execution result.
Its Noise counterpart is constructed by modifying one necessary question condition against the same database, followed by verification of the unmet requirement and the corresponding abstention outcome.
After joint question rewriting, query replay and an independent language-model audit further check execution consistency and semantic alignment.
Building on these checks, we conduct an additional human audit of challenging test pairs, focusing on clinical plausibility and semantic alignment rather than SQL syntax alone.

\paragraph{Selection of challenging pairs.}
We identify candidates using the initial single-run Base evaluations of GPT-5.4 (xhigh) and Gemini-3.1-Pro under the configurations used for the main comparison.
GPT-5.4 uses its default sampling configuration, while Gemini-3.1-Pro uses temperature 0.0.
A pair is included if it contains at least one question that both models fail to solve.
This criterion identifies 109 jointly failed Clean questions and 281 jointly failed Noise questions, with 46 pairs appearing in both groups.
The candidate set contains $109 + 281 - 46 = 344$ distinct pairs.
We review all 344 pairs, covering 688 questions, including counterparts that do not themselves meet the joint-failure criterion.
This screening identifies challenging cases without certifying task validity.

\paragraph{Annotation procedure.}
Each selected pair is independently reviewed by two annotators with medical AI backgrounds and experience in EHR data analysis and SQL-based database querying.
Annotators are shown both natural-language questions, the reference SQL queries, execution results, extracted answers, relevant schema information, and the database evidence used during construction.
They assess whether each question faithfully expresses the intended clinical information need and whether the reference outcome is justified by the available records.
For Clean questions, the review focuses on the evidence supporting the requested answer.
For Noise questions, it focuses on whether the modified condition leads to the stated unmet requirement that justifies abstention.
A pair is marked valid only if both questions and their pairing satisfy all four criteria below:

\begin{itemize}[leftmargin=*]
\item \textbf{Question clarity and semantic alignment.}
Both questions are clear, and reference queries implement the intended clinical concepts, temporal constraints, query scopes, and computations.

\item \textbf{Clean answer support.}
The reference execution and supporting records justify the Clean answer, including the relevant values, units, and aggregation rules where applicable.

\item \textbf{Noise abstention support.}
The available records support the stated unmet requirement and justify abstention under the question's intended semantics.
An empty or \texttt{NULL} query result alone is insufficient without evidence establishing why the requested answer cannot be supported.

\item \textbf{Pair consistency.}
The two questions preserve the intended clinical goal and differ in the designated condition.
Question rewriting does not introduce unintended changes to other constraints, and the difference between answering and abstaining follows from the intended modification.
\end{itemize}

Annotators assign a binary pair-level validity label: 1 indicates that both questions and their pairing satisfy all criteria, while 0 indicates a substantive concern requiring further inspection.
Minor wording issues that do not affect task interpretation, reference logic, or the final outcome are recorded separately and are not counted as invalid.
Disagreements are resolved through joint adjudication after inspecting the questions, reference queries, execution results, and supporting evidence.

\paragraph{Results.}
Table~\ref{tab:human-validation-summary} summarizes the human validation results.
After adjudication and before correction, 317 of the 344 audited pairs are judged valid, corresponding to a pair-level validity rate of 92.15\%.
The remaining 27 pairs (7.85\%) contain substantive issues requiring correction.
This rate describes the quality of the original audited pairs, rather than their validity after issues have been corrected.
Because the audit targets pairs containing jointly failed questions, the result provides a targeted assessment of challenging cases rather than a random-sample estimate for the benchmark.

\paragraph{Inter-annotator agreement.}
Before adjudication, the two annotators agree on 338 of the 344 binary pair-level labels, yielding 98.26\% raw agreement.
Disagreements on the remaining 6 pairs (1.74\%) are resolved through joint adjudication.
Both the validity rate and inter-annotator agreement use pairs as the unit of analysis, with a denominator of 344, although all 688 questions are inspected.

\begin{table}[t]
\centering
\small
\caption{Human validation summary for challenging EHR-RobustGym test pairs.
Validity is assessed after adjudication but before correction; agreement is measured before adjudication.}
\label{tab:human-validation-summary}
\renewcommand{\arraystretch}{1.1}
\begin{tabular}{lr}
\toprule
\textbf{Item} & \textbf{Count (rate)} \\
\midrule
Jointly failed Clean questions & 109 \\
Jointly failed Noise questions & 281 \\
Pairs with joint failure on both sides & 46 \\
\midrule
Audited pairs / questions & 344 / 688 \\
\midrule
Valid pairs before correction & 317 / 344 (92.15\%) \\
Pairs requiring substantive correction & 27 / 344 \\
Pairs with initial reviewer agreement & 338 / 344 (98.26\%) \\
Pairs with initial reviewer disagreement & 6 / 344 \\
\bottomrule
\end{tabular}
\end{table}

\paragraph{Correction and final evaluation.}
All 27 pairs with confirmed substantive issues are corrected rather than removed.
We revise affected question wording, reference queries, or reference outcomes, then replay the queries and recheck both questions against the annotation criteria.
This process verifies that the Clean answer remains supported, the Noise abstention is justified, and the intended relationship between the two questions is preserved.
We rerun the affected tasks for all evaluated agents under the same experimental settings, including repeated-run evaluations where applicable.
Updated scores and trajectories replace their pre-correction versions, and affected aggregate results are recomputed.
All reported EHR-RobustGym results correspond to the finalized test set.

\section{Experimental Details}
\label{app:experimental-details}

\subsection{Environment and Interaction Settings}
\label{app:environment-settings}

Each episode uses a task-scoped, read-only DuckDB connection named \texttt{conn}. Python variables, imports, and intermediate results persist across actions within the episode. A new episode starts with fresh execution state against the same database. The agent sees printed outputs and execution feedback, with omitted portions of long outputs explicitly marked. Table~\ref{tab:environment-settings} summarizes the limits.

\begin{table}[!htbp]
\caption{Environment and interaction settings.}
\label{tab:environment-settings}
\centering
\small
\setlength{\tabcolsep}{5pt}
\renewcommand{\arraystretch}{1.24}
\begin{tabularx}{\textwidth}{@{\hspace{4pt}}>{\raggedright\arraybackslash\bfseries}p{\dimexpr0.25\textwidth-8pt\relax}>{\raggedright\arraybackslash}X@{\hspace{4pt}}}
\toprule
\textbf{Setting} & \textbf{Value or behavior} \\
\midrule
Input limit & 65,536 tokens in the accumulated model input \\
Per-response limit & 4,096 output tokens under the model service's completion-token accounting \\
Interaction limit & 20 accepted model responses, including invalid responses and final submission \\
Tool timeout & 120 seconds of execution, excluding queue waiting \\
\bottomrule
\end{tabularx}
\end{table}

The question triggers the first turn. Each accepted model completion counts as one turn, including invalid or empty responses and final submission. Tool observations and transport retries without a completion do not add turns. Input limits use the tokenizer locally and reported API input usage for external models. Exhausting the input or turn limit without a valid final answer scores zero.

Model-generated SQL/Python errors and ordinary tool timeouts remain in the trajectory as feedback and do not alone determine task success. Infrastructure or judge failures must be resolved before evaluation is reported and are not converted to zero scores or removed from the denominator. During online training, unresolved failures stop the affected update rather than supplying a false reward.

\subsection{Models and Sampling Settings}
\label{app:sampling-settings}

Table~\ref{tab:model-endpoints} lists model identifiers and reasoning settings. API-based denotes access, not whether weights are public. The five local open-weight models use vLLM~\citep{vllm} for EHR evaluation.

\begin{table}[!htbp]
\caption{Model identifiers and reasoning settings.}
\label{tab:model-endpoints}
\centering
\small
\setlength{\tabcolsep}{5pt}
\renewcommand{\arraystretch}{1.24}
\begin{tabularx}{\textwidth}{@{\hspace{4pt}}>{\raggedright\arraybackslash}p{0.265\textwidth}>{\raggedright\arraybackslash}X>{\raggedright\arraybackslash}p{0.19\textwidth}@{\hspace{4pt}}}
\toprule
\textbf{Display name} & \textbf{Model identifier} & \textbf{Reasoning setting} \\
\midrule
\ehrappgroup{3}{API-based models}
\textbf{GPT-5.4 (xhigh)} & \texttt{gpt-5.4} & xhigh \\
\textbf{GPT-4.1} & \texttt{gpt-4.1-2025-04-14} & No override \\
\textbf{Gemini-3.1-Pro} & \texttt{gemini-3.1-pro-preview} & No override \\
\textbf{Claude Sonnet 4.6} & \texttt{claude-sonnet-4-6} & Thinking disabled \\
\textbf{Kimi-K2.5} & \texttt{Kimi-K2.5} & Thinking disabled \\
\textbf{DeepSeek-V3.2} & \texttt{DeepSeek-V3.2-Exp} & Thinking disabled \\
\textbf{Qwen3-235B-A22B} & \texttt{Qwen3-235B-A22B-}\allowbreak\texttt{Instruct-2507} & Thinking disabled \\
\addlinespace[0.4em]
\ehrappgroup{3}{Locally deployed open-weight models}
\textbf{Qwen3-4B/8B/14B/32B} & Corresponding \texttt{Qwen3-} instruction checkpoint & Thinking disabled \\
\textbf{Llama-3.1-8B} & \texttt{Llama-3.1-8B-Instruct} & Native instruction \\
\bottomrule
\end{tabularx}
\end{table}

The seven API-based models and Qwen3-32B each have four Base runs on the same 1,008 Clean-Noise test pairs: two at temperature 0 and two at 0.6, except GPT-5.4, which uses four service-default xhigh runs.
The main tables use the first Base run, and PE, other local Base models, and trained checkpoints use one temperature-zero run per setting (service-default xhigh for GPT-5.4).

\subsection{Outcome Evaluation}
\label{app:scoring-and-diagnosis}
\label{app:outcome-evaluation}

Evaluation and online training use the same outcome scorer. Clean questions use string matching to compare the submitted Answer with the execution-verified reference answer. Noise questions require an Answer of \texttt{NULL}, with an LLM judge checking whether visible tool evidence supports abstention. The final Reasoning is not a separate task-score component and cannot replace an incorrect Answer or missing tool evidence. Trajectory error analysis separately examines interaction and reporting errors.

\textbf{Answer extraction and matching.}
The scorer extracts the complete Answer. Ambiguous or competing Answer fields fail. Clean answer matching preserves the requested values, units, and associated records. For Noise questions, after whitespace, markup, and case normalization, the whole Answer must be \texttt{NULL} or \texttt{NULL.}. A mixed partial answer is not accepted as a NULL submission.

Unrequested auxiliary counts may be omitted. Supplied counts for the requested result must agree in both population and counting unit. The size of an eligible cohort is different from the number of admissions reporting a modal value. Clearly identified supplementary results for another population cannot replace the requested answer and are assessed against their own evidence and reporting requirements in error analysis. Unresolved alternatives for the same requested field do not constitute a complete answer.

\textbf{Evidence for abstention.}
After the NULL gate, the scorer checks the ordered tool calls and feedback shown to the agent. A trajectory without any executed tool call fails directly. Otherwise, one valid GPT-5.2 verdict determines Abstention Validity and cites the supporting tool turns.
The question and evidence contract define the required scope. Reference SQL provides an accepted implementation but does not supply evidence acquired by the agent. Codebook definitions and precomputed record-scope equivalences clarify names and linkage only within their stated domains. Evidence verification uses executed tool calls and observations actually shown during the interaction. Agent-generated tables, unprinted variable contents, and reference replays cannot replace missing observations. Equivalent queries and stepwise computations are accepted when their visible outputs establish a sufficient reason for abstention. A correctly scoped empty result or a complete event list lacking the requested ordinal can suffice. A failed query, wrong-scope empty result, or truncated preview alone cannot. Earlier sufficient evidence remains usable after a later failed query.

\textbf{Population reporting and numerical zero.}
Population summaries with zero or one to four eligible units require withholding the aggregate. The latter also withhold the exact count. A visible count for a broader population can establish that the requested subgroup is below the threshold only when the subgroup relation is valid. Such a bound does not establish an exact subgroup count or distinguish zero from one to four. A broader population above the threshold does not establish that the requested subgroup is reportable. First/last selection is not monotone, so enlarging the source before selecting an event does not automatically yield a valid bound on the requested population.

A defined, reportable zero is a supported value, not the NULL marker. A zero count showing missing records cannot replace an unavailable measurement or undefined aggregate. Tasks concern attributes, ordered events, and summaries, rather than standalone count or existence questions. Population disclosure restrictions appear in the agent prompt and reporting-error taxonomy.

\textbf{Judge configuration and failures.}
For Noise evidence verification, GPT-5.2 uses high reasoning effort, service-default sampling, and an 8,192-token completion limit. One format-repair response is permitted after an invalid judgment, but a valid verdict is not resampled. Transient failures may be retried. Unresolved infrastructure or judge failures raise an error rather than assigning zero task success to the agent.

\subsection{Reporting and Interaction Statistics}
\label{app:result-reporting}
\label{app:interaction-statistics}

Tables~\ref{tab:main-results} and~\ref{tab:mitigation-results} use 1,008 questions per condition. In column order, the six intents contain 130, 33, 212, 23, 390, and 220 questions. Avg.\ weights by question count. Scores use the first complete temperature-zero run, or first service-default xhigh run for GPT-5.4. Base has no task-specific prompting or training. For mitigation, SFT uses 1,000 demonstrations over 250 updates, and each continuation adds 250 updates independently from SFT-250. Size groups use total parameter counts.

Success rates are computed from integer successes over the fixed question set. The Noise NULL Rate uses all Noise questions as its denominator. Conditional Abstention Validity uses only submitted NULL answers and is undefined when there are none. Thus Noise success is the product of NULL Rate and conditional validity before rounding. Aggregated validity is weighted by NULL counts, not by averaging subgroup percentages. Displayed scores and $\Delta$ values are independently rounded from exact quantities. Light-red cells indicate displayed success below 5\% for readability, not significance. Bold $\Delta$ includes exact ties. Training comparisons use one training seed, and smoothed update curves do not represent uncertainty across independent training seeds.

\begin{table}[!htbp]
\caption{Interaction statistics and their counting units.}
\label{tab:interaction-definitions}
\centering
\small
\setlength{\tabcolsep}{5pt}
\renewcommand{\arraystretch}{1.24}
\begin{tabularx}{\textwidth}{@{\hspace{4pt}}>{\raggedright\arraybackslash\bfseries}p{0.245\textwidth}>{\raggedright\arraybackslash}X>{\raggedright\arraybackslash}p{0.32\textwidth}@{\hspace{4pt}}}
\toprule
\textbf{Statistic} & \textbf{Counted event} & \textbf{Rate denominator or exclusion} \\
\midrule
\ehrappgroup{3}{Model responses and tool interaction}
Interaction Turns & Accepted model completions & Includes invalid and final responses \\
Tool Calls & Dispatched code actions & Excludes final submissions \\
Response Format Errors & Responses rejected by the action parser & Accepted model responses \\
Tool Call Format Errors & Malformed responses attempting a tool action & Dispatched calls plus malformed tool requests \\
Tool Execution Errors & Failed dispatched calls & Dispatched calls, excluding timeouts from the numerator \\
Tool Timeouts & Calls interrupted at the execution limit & Dispatched calls \\
\bottomrule
\end{tabularx}
\end{table}

A successful tool call may contain a caught query error, and an empty result is not automatically an error. Figure~\ref{fig:tool-recovery}(a) groups first-run Base trajectories from eight models. Within each condition, all four outcome/error shares use 1,008 questions as the denominator. The groups without execution errors can include trajectories that make no tool calls or encounter timeouts or response-format errors. Interaction statistics are descriptive and do not contribute extra reward terms.

\subsection{Capability Evaluation}
\label{app:capability-evaluation}

We evaluate eight general and medical benchmarks with 10,072 questions per checkpoint. The reported mean weights these eight benchmarks equally, not by question count. Table~\ref{tab:capability-settings} lists subsets and metrics, including five external EHR benchmarks. Figure~\ref{fig:training-curves}(a) reports Qwen3-14B's task success on these external benchmarks before and after SFT and subsequent GRPO. These external benchmarks use their respective final-answer string-matching criteria.

\begin{table}[!htbp]
\caption{Evaluation subsets and metrics for general, medical, and external EHR benchmarks.}
\label{tab:capability-settings}
\centering
\small
\setlength{\tabcolsep}{4.5pt}
\renewcommand{\arraystretch}{1.24}
\begin{tabularx}{\textwidth}{@{\hspace{4pt}}>{\raggedright\arraybackslash}p{0.375\textwidth}>{\raggedright\arraybackslash}p{0.175\textwidth}r>{\raggedright\arraybackslash}X@{\hspace{4pt}}}
\toprule
\textbf{Benchmark} & \textbf{Evaluation subset} & \textbf{Questions} & \textbf{Metric} \\
\midrule
\ehrappgroup{4}{General capabilities}
ARC-Challenge~\citep{arc} & Test & 1,172 & Length-normalized accuracy \\
HellaSwag~\citep{hellaswag} & Validation & 2,000 & Length-normalized accuracy \\
WinoGrande XL~\citep{winogrande} & Validation & 1,267 & Accuracy \\
GSM8K~\citep{gsm8k} & Test & 1,319 & Exact match, strict extraction \\
IFEval~\citep{ifeval} & Official prompts (named train) & 541 & Prompt-level strict accuracy \\
\addlinespace[0.4em]
\ehrappgroup{4}{Medical question answering}
MedQA (English)~\citep{jin2021disease} & Four-option test & 1,273 & Accuracy \\
MedMCQA~\citep{pal2022medmcqa} & Validation & 2,000 & Accuracy \\
PubMedQA PQA-L~\citep{jin-etal-2019-pubmedqa} & Original test & 500 & Accuracy \\
\addlinespace[0.4em]
\ehrappgroup{4}{External EHR benchmarks}
TREQS~\citep{wang2020text} & Test subset & 613 & Task success rate \\
EHRSQL (eICU)~\citep{ehrsql} & Test subset & 371 & Task success rate \\
EHRSQL (MIMIC-III)~\citep{ehrsql} & Test subset & 421 & Task success rate \\
EHR-SeqSQL~\citep{ryu2024ehr} & Test subset & 293 & Task success rate \\
EHR-Complex~\citep{ehrcomplex} & Fixed subset & 500 & Task success rate \\
\bottomrule
\end{tabularx}
\end{table}

For the eight general and medical benchmarks, sampling is without replacement using seed 42, retaining complete evaluation splits smaller than 2,000 questions. MedMCQA preserves subject proportions, and PubMedQA preserves answer-label proportions. GSM8K uses five fixed training examples. Compared checkpoints of each model use that model's Base tokenizer, and all comparisons share the same sample identities. The evaluator is lm-evaluation-harness 0.4.13~\citep{evalharness}, using native chat templates without an extra system instruction. Qwen3 thinking is disabled, and few-shot examples are included within one user prompt rather than as separate turns.

Multiple-choice tasks compare conditional continuation log-likelihoods. ARC-Challenge and HellaSwag use the harness's character-length normalization. GSM8K uses greedy generation with a 1,024-token limit and strict extraction after \texttt{\#\#\#\#}. IFEval uses greedy generation with a 1,280-token limit and requires all instructions in a prompt to pass for prompt-level strict accuracy. MedQA and PubMedQA require field-name conversion to the native templates, without changing questions, options, or labels. Answer-bearing auxiliary fields do not enter model inputs.

The capability evaluations use a documented IFEval implementation difference: the letter-frequency check counts the specified literal character, including punctuation, instead of replacing a nonalphabetic target. Only this check is changed; the native result processor is otherwise used. Two of the 541 prompts are affected by this character-handling rule. Saved Base, SFT, and SFT-initialized continuation evaluations use this rule. Original generated responses are preserved when metrics are recomputed. Raw scores for the main capability comparisons appear in Appendix~\ref{app:capability-comparisons}.

\section{Training Details}
\label{app:training-details}

\subsection{Training Data and Checkpoints}
\label{app:demonstration-selection}

The selected training trajectories (demonstrations) come from GPT-5.4, GPT-5.5~\citep{gpt55}, GPT-5.6-sol~\citep{gpt56sol}, Gemini-3.1-Pro, and Kimi K3~\citep{kimi3}. The GPT teachers use xhigh reasoning with service-default sampling, Gemini uses temperature 0.6, and Kimi uses service-default sampling. Of the 2,000 demonstrations, 1,150 are unguided and 850 use teacher-only reference explanations and queries. Candidates must pass task evaluation and include successful tool-based evidence acquisition. Recovered tool errors are allowed, but three consecutive equivalent failed code executions are excluded. Guided candidates additionally pass naturalness and self-containment checks by Gemini-3.1-Pro at temperature zero. Noise curation by Gemini-3.1-Pro requires a quality score of at least four out of five, cited supporting tool feedback, and no unresolved issues. Student inputs preserve the original dialogue but exclude teacher-only guidance, separately returned reasoning fields, and evaluator metadata.

We randomly sample 1,000 Clean-Noise pairs from the training set, comprising 2,000 questions, with one verified demonstration per question. SFT-250 processes the first 1,000 demonstrations in a fixed shuffled order over 250 updates, half an epoch of the full dataset. DPO, PPO, and GRPO independently continue from this checkpoint for another 250 updates on the remaining 1,000 questions, comprising 495 Clean and 505 Noise requests. The stages share no question IDs, although counterparts from the same pair may occur in different stages. All five students use the same fixed preference set for offline DPO, pairing verified teacher demonstrations with scored unsuccessful trajectories from the five local models for the same question. Online PPO and GRPO collect fresh rollouts through environment interaction. Each branch starts a fresh optimizer and is evaluated at its specified endpoint without test-score-based checkpoint selection. See Table~\ref{tab:training-config} for settings. We also retain an additional SFT-500 checkpoint after one epoch over all 2,000 demonstrations. Its capability scores and Qwen3-8B training-data scaling appear in Appendices~\ref{app:capability-comparisons} and~\ref{app:sft-scaling}.

\subsection{Training Configuration}
\label{app:optimization-background}

\begin{table}[!htbp]
\caption{Training settings shared across methods and specific to each stage.}
\label{tab:training-config}
\centering
\small
\begin{tabularx}{\textwidth}{@{\hspace{4pt}}>{\raggedright\arraybackslash\bfseries}p{\dimexpr0.20\textwidth-8pt\relax}>{\raggedright\arraybackslash}X@{\hspace{4pt}}}
\toprule
Scope & Settings \\
\midrule
All methods & Full-parameter training; 250 updates; policy learning rate $10^{-6}$; AdamW~\citep{adamw} ($\beta_1=0.9$, $\beta_2=0.999$, weight decay 0.01); constant schedule without warm-up; BF16; seed 42. \\
SFT & Instruction initialization; batch of four dialogues; ms-swift 4.5.3~\citep{ms_swift}; 69,632 encoded tokens without truncation or packing. \\
DPO & SFT-250 policy and frozen reference; batch of four preferences; $\beta=0.1$; the same framework and token limit as SFT. \\
PPO and GRPO & SFT-250 policy and frozen reference; verl~\citep{verl}; four questions per update, eight rollouts per question; temperature 0.6 and top-$p$ 1.0; initial-prompt limit 8,192 tokens and engine context 69,632 tokens. \\
Online policy loss & KL coefficient 0.001; actor clip 0.2; dual clip 3~\citep{dualclip}; entropy coefficient 0; low-variance KL estimator; log-ratio bounds $[-20,20]$ and KL cap 10. \\
PPO critic & Learning rate $10^{-5}$; $\gamma=\lambda=1$; no warm-up; value clip 0.5. \\
GRPO normalization & Sample standard deviation with $\epsilon=10^{-6}$. \\
\bottomrule
\end{tabularx}
\ehrtablenote{DPO, PPO, and GRPO independently continue from SFT-250. Interaction limits are listed in Table~\ref{tab:environment-settings}.}
\end{table}

Losses apply only to assistant tokens, with tool observations retained as context and inserted template-only tokens masked. DPO uses summed assistant-token log-probabilities without length normalization or an auxiliary SFT loss. PPO and GRPO use token-mean reduction and KL regularization in the policy loss, separately from the terminal task reward. PPO initializes the critic backbone from SFT-250 with a new scalar value head. PPO policy advantages are whitened, while value targets use unwhitened advantages and skip tool-observation positions. SFT and DPO use eight GPUs with ZeRO-3~\citep{zero3} and sequence parallelism 4. GRPO uses 16 GPUs with FSDP~\citep{fsdp}. PPO uses FSDP on eight GPUs for Qwen3-4B/8B, 16 for Qwen3-14B and Llama, and 32 for Qwen3-32B. Training-data scaling settings appear in Appendix~\ref{app:sft-scaling}.

\subsection{Self-Improvement}
\label{app:self-improvement}

Starting from SFT-250, each model undergoes RS+SFT, DPO, and one additional DPO iteration. Before each stage, the current policy samples eight trajectories per question from the remaining 1,000-question pool at temperature 0.6. RS+SFT retains one successful trajectory per selected question, while DPO and iDPO retain one successful and one unsuccessful trajectory for the same question. Each stage trains for one epoch using the shared outcome scorer for selection, a fresh optimizer, batch size four, and seed 42. Other settings follow SFT or DPO in Table~\ref{tab:training-config}, with the DPO reference reset to the stage-start checkpoint. Table~\ref{tab:self-improvement-selection} reports the selected question counts.

\begin{table}[!htbp]
\caption{Number of questions selected at each self-improvement stage.}
\label{tab:self-improvement-selection}
\centering
\small
\setlength{\tabcolsep}{5pt}
\renewcommand{\arraystretch}{1.24}
\begin{tabularx}{\textwidth}{@{\hspace{4pt}}>{\raggedright\arraybackslash}X*{3}{>{\raggedleft\arraybackslash}p{0.20\textwidth}}@{\hspace{4pt}}}
\toprule
\textbf{Model} & \textbf{RS+SFT} & \textbf{DPO} & \textbf{iDPO} \\
\midrule
Qwen3-4B & \ehrappbar{0.683}{683} & \ehrappbar{0.614}{614} & \ehrappbar{0.519}{519} \\
Qwen3-8B & \ehrappbar{0.736}{736} & \ehrappbar{0.670}{670} & \ehrappbar{0.647}{647} \\
Qwen3-14B & \ehrappbar{0.832}{832} & \ehrappbar{0.675}{675} & \ehrappbar{0.702}{702} \\
Qwen3-32B & \ehrappbar{0.864}{864} & \ehrappbar{0.685}{685} & \ehrappbar{0.631}{631} \\
Llama-3.1-8B & \ehrappbar{0.882}{882} & \ehrappbar{0.586}{586} & \ehrappbar{0.244}{244} \\
\bottomrule
\end{tabularx}
\ehrtablenote{Each stage trains for one epoch on its selected questions. Bars share a scale of 1,000 candidate questions.}
\end{table}

\section{Additional Results}
\label{app:additional-results}

Tables~\ref{tab:noise-type-results} and~\ref{tab:scope-track-results} use the same evaluations as Tables~\ref{tab:main-results} and~\ref{tab:mitigation-results}. Here, SFT denotes SFT-250; +DPO, +PPO, and +GRPO are independent 250-update continuations. Reporting follows Appendix~\ref{app:result-reporting}.

\subsection{Performance by Noise Type}
\label{app:per-type}

Table~\ref{tab:noise-type-results} reports noise types and their Clean counterparts, with 336 questions per condition and type. Noise type describes the unsupported condition, not the agent error. Noise success requires NULL and sufficient evidence; high conditional abstention validity does not imply high success.

\begingroup
\fontsize{8}{10}\selectfont
\setlength{\tabcolsep}{2pt}
\setlength{\fboxsep}{0pt}
\renewcommand{\arraystretch}{1.10}
\definecolor{EHRTrainingGreen}{RGB}{0,128,90}
\setlength{\LTcapwidth}{\textwidth}
\begin{longtable}{@{\hspace{4pt}}>{\raggedright\arraybackslash}p{\dimexpr.205\textwidth-8pt\relax}>{\raggedright\arraybackslash}p{.085\textwidth}*{6}{>{\centering\arraybackslash}p{.105\textwidth}}@{\hspace{4pt}}}
\caption{Performance by noise type and on paired Clean questions.}\label{tab:noise-type-results}\label{tab:entity-results}\label{tab:attribute-results}\label{tab:context-results}\\
\toprule
& & \multicolumn{2}{c}{\textbf{Record-level}} & \multicolumn{2}{c}{\textbf{Value-level}} & \multicolumn{2}{c@{\hspace{4pt}}}{\textbf{Query-level}} \\
\cmidrule(lr){3-4}\cmidrule(lr){5-6}\cmidrule(l){7-8}
\textbf{Model} & \textbf{Method} & Success & NR / AV & Success & NR / AV & Success & NR / AV \\
\midrule
\endfirsthead
\multicolumn{8}{@{\hspace{4pt}}l}{\tablename~\thetable{} (continued)}\\
\toprule
& & \multicolumn{2}{c}{\textbf{Record-level}} & \multicolumn{2}{c}{\textbf{Value-level}} & \multicolumn{2}{c@{\hspace{4pt}}}{\textbf{Query-level}} \\
\cmidrule(lr){3-4}\cmidrule(lr){5-6}\cmidrule(l){7-8}
\textbf{Model} & \textbf{Method} & Success & NR / AV & Success & NR / AV & Success & NR / AV \\
\midrule
\endhead
\midrule
\multicolumn{8}{r@{\hspace{4pt}}}{Continued on next page}\\
\endfoot
\bottomrule
\endlastfoot
\textbf{GPT-5.4 (xhigh)} & Base & 73.5/58.3 & 66.7/87.5 & 78.3/65.8 & 74.4/88.4 & 83.0/43.2 & 52.1/82.9\\
 & PE & 78.0/58.6 & 67.3/87.2 & 79.5/65.5 & 72.9/89.8 & 85.1/41.7 & 50.0/83.3\\
\midrule
\textbf{GPT-4.1} & Base & 51.2/40.5 & 79.8/50.7 & 29.2/23.8 & 84.5/28.2 & 32.7/21.1 & 77.4/27.3\\
 & PE & 51.2/39.3 & 82.4/47.7 & 27.7/22.0 & 85.1/25.9 & 36.0/22.6 & 78.3/28.9\\
\midrule
\textbf{Gemini-3.1-Pro} & Base & 82.1/61.9 & 66.1/93.7 & 78.6/62.2 & 66.1/94.1 & 89.6/51.2 & 53.9/95.0\\
 & PE & 85.4/66.4 & 70.8/93.7 & 79.2/67.9 & 72.9/93.1 & 90.2/50.6 & 51.8/97.7\\
\midrule
\textbf{Claude Sonnet 4.6} & Base & 64.6/39.9 & 42.3/94.4 & 64.3/26.2 & 29.5/88.9 & 58.6/12.8 & 15.2/84.3\\
 & PE & 73.8/46.1 & 50.0/92.3 & 68.8/35.7 & 41.1/87.0 & 61.6/15.2 & 15.8/96.2\\
\midrule
\textbf{Kimi-K2.5} & Base & 83.0/41.7 & 42.6/97.9 & 78.9/48.5 & 48.5/100.0 & 85.4/21.1 & 21.1/100.0\\
 & PE & 83.6/46.1 & 46.7/98.7 & 75.0/49.7 & 50.0/99.4 & 86.0/25.6 & 25.9/98.9\\
\midrule
\textbf{DeepSeek-V3.2} & Base & 75.3/66.1 & 72.9/90.6 & 70.2/52.1 & 61.0/85.4 & 78.3/42.3 & 47.0/89.9\\
 & PE & 79.5/69.0 & 75.9/91.0 & 68.8/54.2 & 60.4/89.7 & 80.1/43.2 & 48.2/89.5\\
\midrule
\textbf{Qwen3-235B-A22B} & Base & 33.0/9.2 & 16.7/55.4 & 14.0/6.5 & 14.9/44.0 & \colorbox{EHRLowSuccess}{3.0}/\colorbox{EHRLowSuccess}{0.9} & 4.5/20.0\\
 & PE & 37.2/13.7 & 24.4/56.1 & 15.2/7.1 & 14.9/48.0 & \colorbox{EHRLowSuccess}{1.2}/\colorbox{EHRLowSuccess}{2.1} & 8.3/25.0\\
\midrule
\textbf{Qwen3-4B} & Base & \colorbox{EHRLowSuccess}{0.0}/\colorbox{EHRLowSuccess}{0.0} & 52.7/0.0 & \colorbox{EHRLowSuccess}{0.3}/\colorbox{EHRLowSuccess}{0.0} & 27.7/0.0 & \colorbox{EHRLowSuccess}{3.3}/\colorbox{EHRLowSuccess}{0.0} & 18.8/0.0\\
 & PE & \colorbox{EHRLowSuccess}{0.3}/\colorbox{EHRLowSuccess}{0.0} & 58.3/0.0 & \colorbox{EHRLowSuccess}{0.6}/\colorbox{EHRLowSuccess}{0.0} & 39.3/0.0 & \colorbox{EHRLowSuccess}{2.7}/\colorbox{EHRLowSuccess}{0.0} & 29.2/0.0\\
 & \cellcolor{EHRTrainingGreen!9}SFT & 42.3/18.5 & 33.3/55.4 & 31.3/24.1 & 37.8/63.8 & 18.2/17.3 & 42.3/40.8\\
 & \cellcolor{EHRTrainingGreen!9}\hspace*{0.5em}+DPO & 36.9/7.4 & 8.0/92.6 & 37.2/10.1 & 12.2/82.9 & 28.0/7.4 & 9.5/78.1\\
 & \cellcolor{EHRTrainingGreen!9}\hspace*{0.5em}+PPO & 45.2/38.4 & 55.1/69.7 & 41.4/36.6 & 49.4/74.1 & 31.5/27.1 & 47.0/57.6\\
 & \cellcolor{EHRTrainingGreen!9}\hspace*{0.5em}+GRPO & 50.3/45.5 & 74.7/61.0 & 48.5/36.3 & 61.3/59.2 & 43.5/25.6 & 48.5/52.8\\
\midrule
\textbf{Qwen3-8B} & Base & 9.2/\colorbox{EHRLowSuccess}{0.9} & 13.7/6.5 & \colorbox{EHRLowSuccess}{0.6}/\colorbox{EHRLowSuccess}{0.6} & 16.1/3.7 & \colorbox{EHRLowSuccess}{0.6}/\colorbox{EHRLowSuccess}{0.0} & 14.9/0.0\\
 & PE & \colorbox{EHRLowSuccess}{4.5}/\colorbox{EHRLowSuccess}{1.8} & 17.6/10.2 & \colorbox{EHRLowSuccess}{0.3}/\colorbox{EHRLowSuccess}{1.8} & 22.6/7.9 & \colorbox{EHRLowSuccess}{0.3}/\colorbox{EHRLowSuccess}{0.0} & 34.8/0.0\\
 & \cellcolor{EHRTrainingGreen!9}SFT & 39.6/25.0 & 52.1/48.0 & 34.2/25.6 & 54.8/46.7 & 27.7/17.9 & 56.3/31.7\\
 & \cellcolor{EHRTrainingGreen!9}\hspace*{0.5em}+DPO & 26.8/31.0 & 40.5/76.5 & 22.9/33.3 & 41.1/81.2 & 20.8/37.5 & 48.5/77.3\\
 & \cellcolor{EHRTrainingGreen!9}\hspace*{0.5em}+PPO & 33.6/22.9 & 24.4/93.9 & 33.3/22.0 & 27.1/81.3 & 22.6/23.5 & 29.2/80.6\\
 & \cellcolor{EHRTrainingGreen!9}\hspace*{0.5em}+GRPO & 52.7/36.6 & 52.7/69.5 & 51.5/47.0 & 54.8/85.9 & 51.2/34.8 & 50.3/69.2\\
\midrule
\textbf{Qwen3-14B} & Base & 20.5/16.7 & 67.6/24.7 & 12.8/\colorbox{EHRLowSuccess}{0.9} & 67.0/1.3 & \colorbox{EHRLowSuccess}{3.0}/\colorbox{EHRLowSuccess}{0.0} & 72.0/0.0\\
 & PE & 9.2/8.6 & 91.1/9.5 & 13.7/\colorbox{EHRLowSuccess}{1.8} & 82.4/2.2 & \colorbox{EHRLowSuccess}{1.5}/\colorbox{EHRLowSuccess}{0.6} & 90.5/0.7\\
 & \cellcolor{EHRTrainingGreen!9}SFT & 48.8/50.9 & 76.8/66.3 & 45.2/28.3 & 58.0/48.7 & 33.0/25.0 & 66.7/37.5\\
 & \cellcolor{EHRTrainingGreen!9}\hspace*{0.5em}+DPO & 43.2/56.8 & 74.7/76.1 & 44.0/51.2 & 75.3/68.0 & 42.0/45.5 & 66.7/68.3\\
 & \cellcolor{EHRTrainingGreen!9}\hspace*{0.5em}+PPO & 74.4/67.6 & 78.6/86.0 & 70.8/65.8 & 75.3/87.4 & 66.1/54.5 & 64.6/84.3\\
 & \cellcolor{EHRTrainingGreen!9}\hspace*{0.5em}+GRPO & 71.4/73.2 & 79.5/92.1 & 62.5/42.6 & 48.5/87.7 & 69.6/48.5 & 55.7/87.2\\
\midrule
\textbf{Qwen3-32B} & Base & 6.8/\colorbox{EHRLowSuccess}{1.2} & 20.5/5.8 & 6.3/\colorbox{EHRLowSuccess}{0.6} & 17.9/3.3 & \colorbox{EHRLowSuccess}{4.8}/\colorbox{EHRLowSuccess}{0.0} & 22.6/0.0\\
 & PE & \colorbox{EHRLowSuccess}{3.0}/\colorbox{EHRLowSuccess}{0.9} & 48.8/1.8 & 5.7/\colorbox{EHRLowSuccess}{0.3} & 32.4/0.9 & 5.7/\colorbox{EHRLowSuccess}{0.6} & 31.8/1.9\\
 & \cellcolor{EHRTrainingGreen!9}SFT & 52.1/43.8 & 71.4/61.3 & 49.1/44.9 & 70.8/63.4 & 45.2/32.1 & 64.3/50.0\\
 & \cellcolor{EHRTrainingGreen!9}\hspace*{0.5em}+DPO & 49.7/61.0 & 90.5/67.4 & 51.8/57.7 & 85.7/67.4 & 47.0/45.5 & 83.0/54.8\\
 & \cellcolor{EHRTrainingGreen!9}\hspace*{0.5em}+PPO & 73.5/69.0 & 81.5/84.7 & 61.6/55.7 & 72.3/77.0 & 53.3/50.9 & 67.3/75.7\\
 & \cellcolor{EHRTrainingGreen!9}\hspace*{0.5em}+GRPO & 63.4/37.8 & 47.3/79.9 & 61.9/39.9 & 45.2/88.2 & 63.1/35.4 & 42.9/82.6\\
\midrule
\textbf{Llama-3.1-8B} & Base & \colorbox{EHRLowSuccess}{0.0}/\colorbox{EHRLowSuccess}{0.0} & 9.2/0.0 & \colorbox{EHRLowSuccess}{0.0}/\colorbox{EHRLowSuccess}{0.9} & 32.7/2.7 & \colorbox{EHRLowSuccess}{0.3}/\colorbox{EHRLowSuccess}{0.0} & 21.1/0.0\\
 & PE & \colorbox{EHRLowSuccess}{0.0}/\colorbox{EHRLowSuccess}{0.0} & 11.9/0.0 & \colorbox{EHRLowSuccess}{0.9}/\colorbox{EHRLowSuccess}{0.0} & 16.4/0.0 & \colorbox{EHRLowSuccess}{0.3}/\colorbox{EHRLowSuccess}{0.0} & 17.3/0.0\\
 & \cellcolor{EHRTrainingGreen!9}SFT & 53.3/33.6 & 52.4/64.2 & 53.9/39.3 & 47.9/82.0 & 50.3/22.6 & 43.8/51.7\\
 & \cellcolor{EHRTrainingGreen!9}\hspace*{0.5em}+DPO & 35.1/39.6 & 60.7/65.2 & 19.9/54.8 & 78.6/69.7 & 20.8/41.4 & 69.3/59.7\\
 & \cellcolor{EHRTrainingGreen!9}\hspace*{0.5em}+PPO & 41.4/24.4 & 66.4/36.8 & 28.9/25.3 & 65.2/38.8 & 5.1/19.6 & 70.2/28.0\\
 & \cellcolor{EHRTrainingGreen!9}\hspace*{0.5em}+GRPO & 48.2/32.7 & 54.8/59.8 & 51.8/50.0 & 59.2/84.4 & 47.3/36.3 & 50.0/72.6\\
\end{longtable}
\ehrtablenote{All values are percentages. Success is Clean / Noise, with 336 questions per condition and noise type. NR is the NULL rate among Noise questions; AV is abstention validity conditional on NULL. \colorbox{EHRLowSuccess}{\strut\phantom{00}} marks success below 5\%; \colorbox{EHRTrainingGreen!9}{\strut\phantom{00}} marks parameter training.}
\endgroup

\subsection{Performance by Scope, Test Track, and Population Size}

Table~\ref{tab:scope-track-results} reports patient- and population-level questions and the in-domain and structure-disjoint test tracks. Each partition contains 504 questions per condition. The two tracks differ in task composition, so their score difference does not isolate the causal effect of structural generalization. Table~\ref{tab:population-support-results} distinguishes zero eligible units from one to four eligible units. The latter accounts for 260 of 504 population Noise questions. Both require evidence-grounded abstention, but only the former establishes that no units qualify. A small eligible population does not imply missing records.

\begingroup
\fontsize{8}{10}\selectfont
\setlength{\tabcolsep}{2pt}
\setlength{\fboxsep}{0pt}
\renewcommand{\arraystretch}{1.10}
\definecolor{EHRTrainingGreen}{RGB}{0,128,90}
\setlength{\LTcapwidth}{\textwidth}
\begin{longtable}{@{\hspace{4pt}}>{\raggedright\arraybackslash}p{\dimexpr.20\textwidth-8pt\relax}>{\raggedright\arraybackslash}p{.078\textwidth}*{5}{>{\centering\arraybackslash}p{.132\textwidth}}@{\hspace{4pt}}}
\caption{Task success by query scope, test track, and eligible population size.}\label{tab:scope-track-results}\label{tab:population-support-results}\\
\toprule
& & \multicolumn{2}{c}{\textbf{Query scope}} & \multicolumn{2}{c}{\textbf{Test track}} & \textbf{Population} \\
\cmidrule(lr){3-4}\cmidrule(lr){5-6}\cmidrule(l){7-7}
\textbf{Model} & \textbf{Method} & Patient & Population & In-domain & Disjoint & Zero / 1--4 \\
\midrule
\endfirsthead
\multicolumn{7}{@{\hspace{4pt}}l}{\tablename~\thetable{} (continued)}\\
\toprule
& & \multicolumn{2}{c}{\textbf{Query scope}} & \multicolumn{2}{c}{\textbf{Test track}} & \textbf{Population} \\
\cmidrule(lr){3-4}\cmidrule(lr){5-6}\cmidrule(l){7-7}
\textbf{Model} & \textbf{Method} & Patient & Population & In-domain & Disjoint & Zero / 1--4 \\
\midrule
\endhead
\midrule
\multicolumn{7}{r@{\hspace{4pt}}}{Continued on next page}\\
\endfoot
\bottomrule
\endlastfoot
\textbf{GPT-5.4 (xhigh)} & Base & 88.5/65.5 & 68.1/46.0 & 79.4/48.8 & 77.2/62.7 & 72.1/21.5\\
 & PE & 91.1/66.9 & 70.6/43.7 & 80.2/47.2 & 81.5/63.3 & 73.8/15.4\\
\midrule
\textbf{GPT-4.1} & Base & 56.2/40.7 & 19.2/16.3 & 39.1/25.0 & 36.3/31.9 & 25.4/7.7\\
 & PE & 56.3/41.1 & 20.2/14.9 & 39.9/23.8 & 36.7/32.1 & 25.0/5.4\\
\midrule
\textbf{Gemini-3.1-Pro} & Base & 94.8/78.6 & 72.0/38.3 & 83.1/49.2 & 83.7/67.7 & 38.9/37.7\\
 & PE & 96.4/79.8 & 73.4/43.5 & 85.1/53.6 & 84.7/69.6 & 47.1/40.0\\
\midrule
\textbf{Claude Sonnet 4.6} & Base & 76.2/35.7 & 48.8/16.9 & 63.3/22.0 & 61.7/30.6 & 29.9/\colorbox{EHRLowSuccess}{4.6}\\
 & PE & 81.3/43.7 & 54.8/21.0 & 71.2/25.2 & 64.9/39.5 & 38.9/\colorbox{EHRLowSuccess}{4.2}\\
\midrule
\textbf{Kimi-K2.5} & Base & 97.2/59.5 & 67.7/14.7 & 82.9/28.4 & 81.9/45.8 & 27.9/\colorbox{EHRLowSuccess}{2.3}\\
 & PE & 98.8/63.5 & 64.3/17.5 & 83.5/31.9 & 79.6/49.0 & 33.6/\colorbox{EHRLowSuccess}{2.3}\\
\midrule
\textbf{DeepSeek-V3.2} & Base & 93.7/56.3 & 55.6/50.6 & 75.0/49.6 & 74.2/57.3 & 49.6/51.5\\
 & PE & 93.8/58.5 & 58.3/52.4 & 75.2/49.8 & 77.0/61.1 & 54.5/50.4\\
\midrule
\textbf{Qwen3-235B-A22B} & Base & 31.3/10.5 & \colorbox{EHRLowSuccess}{2.0}/\colorbox{EHRLowSuccess}{0.6} & 20.6/6.9 & 12.7/\colorbox{EHRLowSuccess}{4.2} & \colorbox{EHRLowSuccess}{0.8}/\colorbox{EHRLowSuccess}{0.4}\\
 & PE & 34.7/15.1 & \colorbox{EHRLowSuccess}{1.0}/\colorbox{EHRLowSuccess}{0.2} & 21.6/9.9 & 14.1/5.4 & \colorbox{EHRLowSuccess}{0.4}/\colorbox{EHRLowSuccess}{0.0}\\
\midrule
\textbf{Qwen3-4B} & Base & \colorbox{EHRLowSuccess}{1.8}/\colorbox{EHRLowSuccess}{0.0} & \colorbox{EHRLowSuccess}{0.6}/\colorbox{EHRLowSuccess}{0.0} & \colorbox{EHRLowSuccess}{1.8}/\colorbox{EHRLowSuccess}{0.0} & \colorbox{EHRLowSuccess}{0.6}/\colorbox{EHRLowSuccess}{0.0} & \colorbox{EHRLowSuccess}{0.0}/\colorbox{EHRLowSuccess}{0.0}\\
 & PE & \colorbox{EHRLowSuccess}{1.8}/\colorbox{EHRLowSuccess}{0.0} & \colorbox{EHRLowSuccess}{0.6}/\colorbox{EHRLowSuccess}{0.0} & \colorbox{EHRLowSuccess}{2.4}/\colorbox{EHRLowSuccess}{0.0} & \colorbox{EHRLowSuccess}{0.0}/\colorbox{EHRLowSuccess}{0.0} & \colorbox{EHRLowSuccess}{0.0}/\colorbox{EHRLowSuccess}{0.0}\\
 & \cellcolor{EHRTrainingGreen!9}SFT & 50.4/31.0 & 10.7/8.9 & 32.3/19.6 & 28.8/20.2 & 15.6/\colorbox{EHRLowSuccess}{2.7}\\
 & \cellcolor{EHRTrainingGreen!9}\hspace*{0.5em}+DPO & 59.1/12.3 & 8.9/\colorbox{EHRLowSuccess}{4.4} & 33.1/10.5 & 34.9/6.2 & \colorbox{EHRLowSuccess}{4.5}/\colorbox{EHRLowSuccess}{4.2}\\
 & \cellcolor{EHRTrainingGreen!9}\hspace*{0.5em}+PPO & 69.4/56.3 & 9.3/11.7 & 34.7/39.7 & 44.0/28.4 & 20.5/\colorbox{EHRLowSuccess}{3.5}\\
 & \cellcolor{EHRTrainingGreen!9}\hspace*{0.5em}+GRPO & 77.2/49.8 & 17.7/21.8 & 45.8/34.9 & 49.0/36.7 & 32.4/11.9\\
\midrule
\textbf{Qwen3-8B} & Base & 6.9/\colorbox{EHRLowSuccess}{1.0} & \colorbox{EHRLowSuccess}{0.0}/\colorbox{EHRLowSuccess}{0.0} & \colorbox{EHRLowSuccess}{2.8}/\colorbox{EHRLowSuccess}{0.8} & \colorbox{EHRLowSuccess}{4.2}/\colorbox{EHRLowSuccess}{0.2} & \colorbox{EHRLowSuccess}{0.0}/\colorbox{EHRLowSuccess}{0.0}\\
 & PE & \colorbox{EHRLowSuccess}{3.4}/\colorbox{EHRLowSuccess}{2.2} & \colorbox{EHRLowSuccess}{0.0}/\colorbox{EHRLowSuccess}{0.2} & \colorbox{EHRLowSuccess}{1.8}/\colorbox{EHRLowSuccess}{1.2} & \colorbox{EHRLowSuccess}{1.6}/\colorbox{EHRLowSuccess}{1.2} & \colorbox{EHRLowSuccess}{0.4}/\colorbox{EHRLowSuccess}{0.0}\\
 & \cellcolor{EHRTrainingGreen!9}SFT & 56.3/29.8 & 11.3/15.9 & 36.1/24.8 & 31.5/20.8 & 21.7/10.4\\
 & \cellcolor{EHRTrainingGreen!9}\hspace*{0.5em}+DPO & 36.5/50.0 & 10.5/17.9 & 18.1/33.5 & 29.0/34.3 & 15.6/20.0\\
 & \cellcolor{EHRTrainingGreen!9}\hspace*{0.5em}+PPO & 52.4/36.3 & 7.3/9.3 & 28.2/28.8 & 31.5/16.9 & 8.6/10.0\\
 & \cellcolor{EHRTrainingGreen!9}\hspace*{0.5em}+GRPO & 80.0/52.4 & 23.6/26.6 & 52.8/45.6 & 50.8/33.3 & 29.1/24.2\\
\midrule
\textbf{Qwen3-14B} & Base & 23.2/11.7 & \colorbox{EHRLowSuccess}{1.0}/\colorbox{EHRLowSuccess}{0.0} & \colorbox{EHRLowSuccess}{4.0}/\colorbox{EHRLowSuccess}{3.4} & 20.2/8.3 & \colorbox{EHRLowSuccess}{0.0}/\colorbox{EHRLowSuccess}{0.0}\\
 & PE & 14.5/6.3 & \colorbox{EHRLowSuccess}{1.8}/\colorbox{EHRLowSuccess}{1.0} & 5.2/\colorbox{EHRLowSuccess}{3.4} & 11.1/\colorbox{EHRLowSuccess}{4.0} & \colorbox{EHRLowSuccess}{0.8}/\colorbox{EHRLowSuccess}{1.2}\\
 & \cellcolor{EHRTrainingGreen!9}SFT & 66.3/52.0 & 18.5/17.5 & 46.0/33.1 & 38.7/36.3 & 22.5/12.7\\
 & \cellcolor{EHRTrainingGreen!9}\hspace*{0.5em}+DPO & 57.7/74.0 & 28.4/28.4 & 48.8/50.4 & 37.3/52.0 & 25.4/31.2\\
 & \cellcolor{EHRTrainingGreen!9}\hspace*{0.5em}+PPO & 87.3/73.2 & 53.6/52.0 & 68.3/61.5 & 72.6/63.7 & 63.5/41.2\\
 & \cellcolor{EHRTrainingGreen!9}\hspace*{0.5em}+GRPO & 89.5/65.5 & 46.2/44.0 & 67.7/57.1 & 68.1/52.4 & 36.9/50.8\\
\midrule
\textbf{Qwen3-32B} & Base & 9.3/\colorbox{EHRLowSuccess}{0.8} & \colorbox{EHRLowSuccess}{2.6}/\colorbox{EHRLowSuccess}{0.4} & 6.5/\colorbox{EHRLowSuccess}{0.6} & 5.4/\colorbox{EHRLowSuccess}{0.6} & \colorbox{EHRLowSuccess}{0.8}/\colorbox{EHRLowSuccess}{0.0}\\
 & PE & 6.9/\colorbox{EHRLowSuccess}{0.8} & \colorbox{EHRLowSuccess}{2.6}/\colorbox{EHRLowSuccess}{0.4} & 7.1/\colorbox{EHRLowSuccess}{0.6} & \colorbox{EHRLowSuccess}{2.4}/\colorbox{EHRLowSuccess}{0.6} & \colorbox{EHRLowSuccess}{0.8}/\colorbox{EHRLowSuccess}{0.0}\\
 & \cellcolor{EHRTrainingGreen!9}SFT & 67.7/55.8 & 30.0/24.8 & 49.2/40.5 & 48.4/40.1 & 31.6/18.5\\
 & \cellcolor{EHRTrainingGreen!9}\hspace*{0.5em}+DPO & 76.0/78.6 & 23.0/31.0 & 52.0/52.2 & 47.0/57.3 & 37.3/25.0\\
 & \cellcolor{EHRTrainingGreen!9}\hspace*{0.5em}+PPO & 85.3/72.4 & 40.3/44.6 & 63.1/60.5 & 62.5/56.5 & 44.3/45.0\\
 & \cellcolor{EHRTrainingGreen!9}\hspace*{0.5em}+GRPO & 81.7/46.2 & 43.8/29.2 & 62.7/41.5 & 62.9/33.9 & 23.0/35.0\\
\midrule
\textbf{Llama-3.1-8B} & Base & \colorbox{EHRLowSuccess}{0.0}/\colorbox{EHRLowSuccess}{0.0} & \colorbox{EHRLowSuccess}{0.2}/\colorbox{EHRLowSuccess}{0.6} & \colorbox{EHRLowSuccess}{0.2}/\colorbox{EHRLowSuccess}{0.2} & \colorbox{EHRLowSuccess}{0.0}/\colorbox{EHRLowSuccess}{0.4} & \colorbox{EHRLowSuccess}{1.2}/\colorbox{EHRLowSuccess}{0.0}\\
 & PE & \colorbox{EHRLowSuccess}{0.6}/\colorbox{EHRLowSuccess}{0.0} & \colorbox{EHRLowSuccess}{0.2}/\colorbox{EHRLowSuccess}{0.0} & \colorbox{EHRLowSuccess}{0.4}/\colorbox{EHRLowSuccess}{0.0} & \colorbox{EHRLowSuccess}{0.4}/\colorbox{EHRLowSuccess}{0.0} & \colorbox{EHRLowSuccess}{0.0}/\colorbox{EHRLowSuccess}{0.0}\\
 & \cellcolor{EHRTrainingGreen!9}SFT & 77.0/37.9 & 28.0/25.8 & 55.2/36.9 & 49.8/26.8 & 28.3/23.5\\
 & \cellcolor{EHRTrainingGreen!9}\hspace*{0.5em}+DPO & 34.1/71.4 & 16.5/19.0 & 29.8/46.6 & 20.8/43.8 & 14.8/23.1\\
 & \cellcolor{EHRTrainingGreen!9}\hspace*{0.5em}+PPO & 48.8/40.3 & \colorbox{EHRLowSuccess}{1.4}/6.0 & 23.0/30.6 & 27.2/15.7 & 7.4/\colorbox{EHRLowSuccess}{4.6}\\
 & \cellcolor{EHRTrainingGreen!9}\hspace*{0.5em}+GRPO & 65.5/43.8 & 32.7/35.5 & 52.0/44.4 & 46.2/34.9 & 40.6/30.8\\
\end{longtable}
\ehrtablenote{All values are percentages. Scope and track columns report Clean / Noise success, with 504 questions per condition in each partition. The last column reports Noise success for zero / one to four eligible units, with 244 / 260 questions. Colors follow Table~\ref{tab:noise-type-results}.}
\endgroup

\subsection{Training-Data Scaling and Self-Improvement Results}
\label{app:training-additional-results}
\label{app:sft-scaling}

\begin{figure}[htbp]
\centering
\includegraphics[width=0.683360\textwidth]{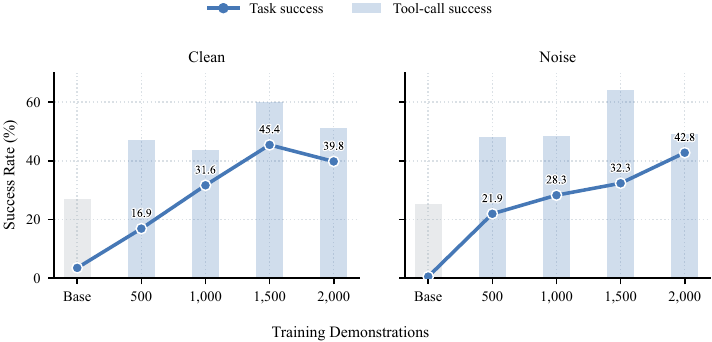}
\caption{Paired Mixed SFT scaling on Qwen3-8B. Lines show task success; bars show successful executions divided by all tool calls, including errors and timeouts. Each selected pair contributes one Clean and one Noise demonstration; each checkpoint is trained for one epoch. Base uses no training demonstrations.}
\label{fig:paired-training-scaling}
\end{figure}

\textbf{Training-data scaling protocol.}
Qwen3-8B uses nested subsets of 250, 500, 750, and 1,000 training pairs, selected with seed 42. Each subset balances query scopes and covers all 45 training programs. Paired Mixed retains both demonstrations per pair, while Clean-only and Noise-only retain the corresponding question only. Every point starts from the instruction checkpoint and trains for one epoch with the SFT settings in Table~\ref{tab:training-config}. Final partial batches are retained, giving 63 and 188 updates for 250 and 750 single-condition demonstrations. Paired Mixed 100\% reuses SFT-500 after all 2,000 demonstrations, not an independent run. At equal pair fractions, Mixed uses twice as many demonstrations and approximately twice as many updates, so equal-example comparisons instead match Mixed 25\% with single-condition 50\%, or Mixed 50\% with single-condition 100\%.

The Mixed setting in Figure~\ref{fig:inference-scaling}(b) uses one demonstration from each of the same 1,000 pairs as the full Clean-only and Noise-only datasets. We assign 500 pairs to Clean and the remaining 500 to Noise, stratifying by query scope, noise type, and training program with seed 42. Each condition contains 250 patient-level and 250 population-level demonstrations. The demonstrations and pair order are unchanged before the same sampler shuffle. All three runs therefore use 1,000 demonstrations, the same pair coverage, and 250 updates; their token counts need not be identical. This run starts from the original instruction checkpoint and is distinct from both the 500-complete-pair Mixed subset and SFT-250 after 1,000 shuffled questions from the 2,000-example dataset. The 250-, 500-, and 750-demonstration Mixed subsets retain the fixed arm assignment and the same pairs as the single-condition subsets, with Clean/Noise counts of 127/123, 251/249, and 377/373. Each subset has equal numbers of patient-level and population-level requests and starts from the original instruction checkpoint for one epoch, giving 63, 125, and 188 updates. Figure~\ref{fig:inference-scaling}(b) includes only these one-demonstration-per-pair settings. Figure~\ref{fig:paired-training-scaling} separately shows the paired Mixed scaling results.

Figure~\ref{fig:inference-scaling}(b,c) and Figure~\ref{fig:paired-training-scaling} report results on the same 1,008 Clean and 1,008 Noise questions. All training results use a single seed. Table~\ref{tab:self-improvement-full-results} reports training updates for sequential self-improvement. At an equal budget of 1,000 demonstrations drawn from the same 1,000 pairs, Mixed reaches 30.1\% Clean and 29.4\% Noise success, whereas Clean-only reaches 35.4\% and 1.1\%, and Noise-only reaches 0.0\% and 51.8\%. Mixed retains success on both conditions. The 500-complete-pair Mixed subset achieves 31.6\%/28.3\% with the same number of demonstrations but covers fewer pairs.

\begin{table}[htbp]
\caption{Training updates during sequential self-improvement.}
\label{tab:self-improvement-full-results}
\centering
\small
\begin{tabularx}{\textwidth}{@{\hspace{4pt}}l*{4}{>{\centering\arraybackslash}X}@{\hspace{4pt}}}
\toprule
\textbf{Model} & \textbf{SFT-250} & \textbf{RS+SFT} & \textbf{DPO} & \textbf{iDPO} \\
\midrule
\ehrappgroup{5}{Updates: stage / cumulative}
Qwen3-4B & 250/250 & 171/421 & 154/575 & 130/705\\
Qwen3-8B & 250/250 & 184/434 & 168/602 & 162/764\\
Qwen3-14B & 250/250 & 208/458 & 169/627 & 176/803\\
Qwen3-32B & 250/250 & 216/466 & 172/638 & 158/796\\
Llama-3.1-8B & 250/250 & 221/471 & 147/618 & 61/679\\
\bottomrule
\end{tabularx}
\ehrtablenote{Stages proceed sequentially from SFT-250. Each stage uses newly collected trajectories and one epoch over its selected set. Cumulative updates include the initial 250 SFT updates.}
\end{table}

\subsection{Capability Retention}
\label{app:capability-comparisons}
\begin{figure}[!ht]
\centering
\includegraphics[width=0.978606\textwidth]{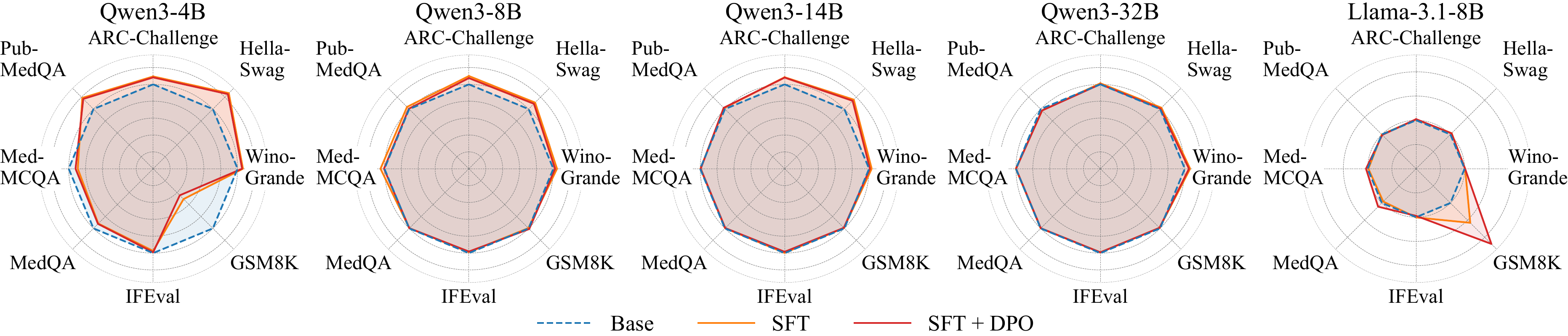}
\includegraphics[width=0.978606\textwidth]{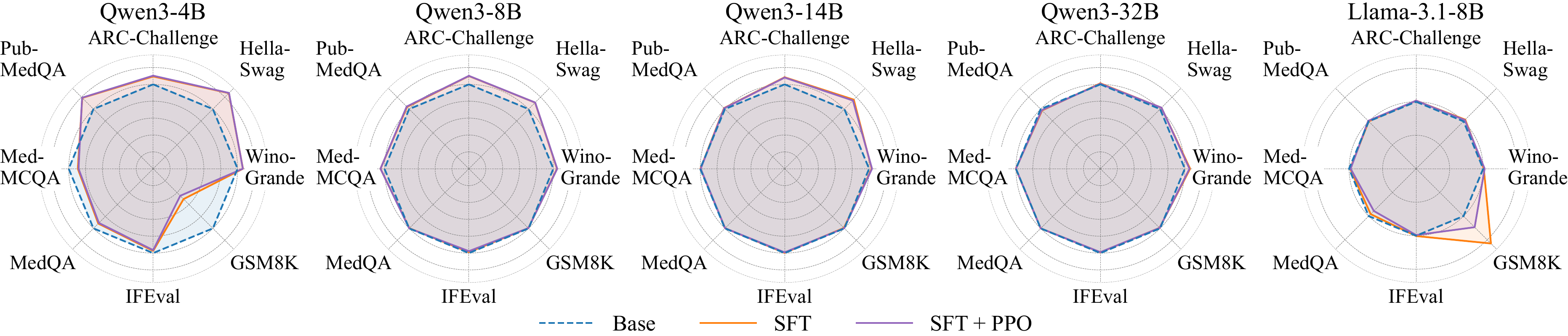}
\caption{Capability retention after SFT-initialized DPO and PPO. Top: DPO. Bottom: PPO. Both continue from SFT-250 for 250 updates. Radii show scores relative to each model's Base score. Raw scores appear in Table~\ref{tab:capability-raw}; radial limits vary across models.}
\label{fig:capability-preferences}
\end{figure}

Figure~\ref{fig:capability-preferences} extends the main capability comparison to DPO and PPO. Each method starts from SFT-250 and performs 250 further updates. Base and SFT-250 are shared with Figure~\ref{fig:alignment-tax}, and all panels use the same eight fixed capability subsets, totaling 10,072 questions per completed checkpoint. Both comparisons include all five models at their final checkpoints. Table~\ref{tab:capability-raw} reports the eight raw benchmark scores and their equal-weight mean. The ratio used for radar radii is separate from these percentage scores, and polygon areas are not directly comparable across model-specific radial limits.

\begingroup
\small
\setlength{\tabcolsep}{2.8pt}
\renewcommand{\arraystretch}{1.20}
\setlength{\LTcapwidth}{\textwidth}
\begin{longtable}{@{\hspace{4pt}}>{\raggedright\arraybackslash}p{\dimexpr.17\textwidth-8pt\relax}*{9}{>{\raggedleft\arraybackslash}p{.078\textwidth}}@{\hspace{4pt}}}
\caption{Raw scores on the eight general and medical benchmarks.}\label{tab:capability-raw}\\
\toprule
& \multicolumn{5}{c}{\textbf{General capabilities}} & \multicolumn{3}{c}{\textbf{Medical QA}} & \\
\cmidrule(lr){2-6}\cmidrule(lr){7-9}
\textbf{Checkpoint} & \makecell[r]{ARC-C} & \makecell[r]{Hella-\\Swag} & \makecell[r]{Wino-\\Grande} & GSM8K & IFEval & \makecell[r]{Med-\\QA} & \makecell[r]{Med-\\MCQA} & \makecell[r]{Pub-\\MedQA} & \textbf{Mean} \\
\midrule
\endfirsthead
\toprule
& \multicolumn{5}{c}{\textbf{General capabilities}} & \multicolumn{3}{c}{\textbf{Medical QA}} & \\
\cmidrule(lr){2-6}\cmidrule(lr){7-9}
\textbf{Checkpoint} & \makecell[r]{ARC-C} & \makecell[r]{Hella-\\Swag} & \makecell[r]{Wino-\\Grande} & GSM8K & IFEval & \makecell[r]{Med-\\QA} & \makecell[r]{Med-\\MCQA} & \makecell[r]{Pub-\\MedQA} & \textbf{Mean} \\
\midrule
\endhead
\midrule
\endfoot
\bottomrule
\endlastfoot
\rowcolor{EHRTablePurplePale}[0pt][0pt]
\multicolumn{10}{@{\hspace{4pt}}l@{\hspace{4pt}}}{\textcolor{EHRTablePurple}{\textbf{Qwen3-4B}}}\\
Base & 37.4 & 46.3 & 55.9 & 78.9 & 79.1 & 31.2 & 39.1 & 58.8 & \textbf{53.3} \\
SFT-250 & 40.9 & 58.6 & 59.4 & 40.2 & 77.1 & 28.7 & 34.9 & 69.8 & \textbf{51.2} \\
SFT-500 & 41.0 & 58.6 & 60.1 & 42.4 & 75.6 & 28.6 & 35.1 & 70.8 & \textbf{51.5} \\
\hspace*{0.5em}+DPO & 40.4 & 57.9 & 58.9 & 35.2 & 78.4 & 28.8 & 36.0 & 68.8 & \textbf{50.5} \\
\hspace*{0.5em}+PPO & 41.2 & 58.8 & 59.4 & 35.8 & 76.3 & 28.4 & 34.6 & 70.0 & \textbf{50.5} \\
\hspace*{0.5em}+GRPO & 40.6 & 58.7 & 59.1 & 33.7 & 77.6 & 28.9 & 35.0 & 71.0 & \textbf{50.6} \\
\addlinespace[0.4em]
\rowcolor{EHRTablePurplePale}[0pt][0pt]
\multicolumn{10}{@{\hspace{4pt}}l@{\hspace{4pt}}}{\textcolor{EHRTablePurple}{\textbf{Qwen3-8B}}}\\
Base & 42.3 & 59.8 & 61.9 & 87.4 & 81.3 & 28.0 & 35.3 & 66.4 & \textbf{57.8} \\
SFT-250 & 46.5 & 66.3 & 64.6 & 87.0 & 80.4 & 27.9 & 36.9 & 68.8 & \textbf{59.8} \\
SFT-500 & 46.7 & 66.9 & 64.7 & 86.9 & 79.1 & 27.9 & 36.8 & 69.8 & \textbf{59.8} \\
\hspace*{0.5em}+DPO & 45.5 & 65.1 & 63.1 & 88.3 & 79.9 & 27.9 & 35.5 & 66.8 & \textbf{59.0} \\
\hspace*{0.5em}+PPO & 46.6 & 66.3 & 64.7 & 87.0 & 79.5 & 28.0 & 36.9 & 69.2 & \textbf{59.8} \\
\hspace*{0.5em}+GRPO & 46.2 & 66.2 & 64.2 & 87.3 & 80.2 & 27.9 & 36.6 & 68.6 & \textbf{59.6} \\
\addlinespace[0.4em]
\rowcolor{EHRTablePurplePale}[0pt][0pt]
\multicolumn{10}{@{\hspace{4pt}}l@{\hspace{4pt}}}{\textcolor{EHRTablePurple}{\textbf{Qwen3-14B}}}\\
Base & 42.5 & 58.9 & 61.3 & 91.4 & 84.8 & 27.8 & 32.0 & 76.2 & \textbf{59.4} \\
SFT-250 & 46.1 & 68.2 & 63.2 & 90.7 & 84.7 & 27.7 & 31.9 & 77.4 & \textbf{61.2} \\
SFT-500 & 47.1 & 69.0 & 64.6 & 91.4 & 83.5 & 27.7 & 31.9 & 77.6 & \textbf{61.6} \\
\hspace*{0.5em}+DPO & 45.9 & 67.0 & 62.0 & 90.5 & 83.5 & 27.7 & 31.8 & 77.8 & \textbf{60.8} \\
\hspace*{0.5em}+PPO & 45.9 & 67.3 & 63.5 & 91.1 & 84.1 & 27.7 & 31.8 & 77.2 & \textbf{61.1} \\
\hspace*{0.5em}+GRPO & 46.4 & 68.0 & 62.8 & 91.5 & 83.7 & 27.7 & 32.0 & 77.8 & \textbf{61.3} \\
\addlinespace[0.4em]
\rowcolor{EHRTablePurplePale}[0pt][0pt]
\multicolumn{10}{@{\hspace{4pt}}l@{\hspace{4pt}}}{\textcolor{EHRTablePurple}{\textbf{Qwen3-32B}}}\\
Base & 49.0 & 71.0 & 62.8 & 94.5 & 84.3 & 28.0 & 32.0 & 76.0 & \textbf{62.2} \\
SFT-250 & 49.6 & 72.6 & 66.8 & 93.7 & 83.5 & 28.0 & 32.1 & 74.2 & \textbf{62.5} \\
SFT-500 & 49.4 & 73.1 & 66.3 & 93.6 & 82.4 & 28.0 & 32.0 & 74.4 & \textbf{62.4} \\
\hspace*{0.5em}+DPO & 49.1 & 71.5 & 65.9 & 93.6 & 83.5 & 27.9 & 32.0 & 74.2 & \textbf{62.2} \\
\hspace*{0.5em}+PPO & 49.3 & 72.6 & 66.3 & 93.8 & 83.4 & 28.0 & 32.0 & 74.6 & \textbf{62.5} \\
\hspace*{0.5em}+GRPO & 48.9 & 72.7 & 66.6 & 93.9 & 84.1 & 28.0 & 32.0 & 74.6 & \textbf{62.6} \\
\addlinespace[0.4em]
\rowcolor{EHRTablePurplePale}[0pt][0pt]
\multicolumn{10}{@{\hspace{4pt}}l@{\hspace{4pt}}}{\textcolor{EHRTablePurple}{\textbf{Llama-3.1-8B}}}\\
Base & 53.6 & 73.7 & 67.8 & 34.0 & 74.3 & 34.2 & 48.6 & 74.6 & \textbf{57.6} \\
SFT-250 & 54.4 & 76.3 & 68.4 & 53.6 & 74.7 & 32.8 & 47.8 & 74.6 & \textbf{60.3} \\
SFT-500 & 52.8 & 75.7 & 69.1 & 59.7 & 74.5 & 31.0 & 46.4 & 74.2 & \textbf{60.4} \\
\hspace*{0.5em}+DPO & 54.8 & 76.0 & 68.4 & 74.5 & 72.3 & 37.9 & 50.2 & 73.6 & \textbf{63.5} \\
\hspace*{0.5em}+PPO & 54.4 & 75.7 & 69.3 & 41.9 & 73.6 & 30.5 & 47.4 & 75.4 & \textbf{58.5} \\
\hspace*{0.5em}+GRPO & 54.3 & 75.5 & 68.4 & 55.1 & 74.3 & 29.5 & 45.6 & 74.4 & \textbf{59.6} \\
\end{longtable}

\ehrtablenote{Scores are percentages under the metrics in Table~\ref{tab:capability-settings}. Mean weights the eight benchmarks equally. +DPO, +PPO, and +GRPO each add 250 updates from SFT-250. SFT-500 is included as an additional checkpoint, not as their initialization.}
\endgroup

\subsection{Repeated Evaluations}
\label{app:repeated-evaluations}
\begin{figure}[!htbp]
\centering
\includegraphics[width=0.680102\textwidth]{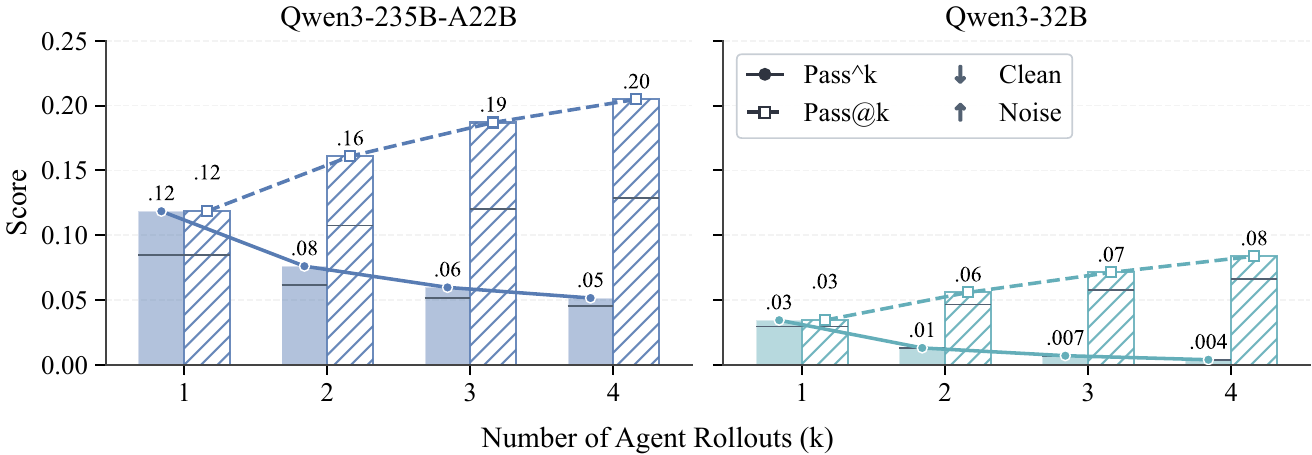}
\caption{Repeated Base evaluations for two additional Qwen3 models. Each model uses two runs at temperature 0 and two at 0.6. Bars stack equally weighted \ehrclean{} and Noise contributions over 2,016 questions. Metrics match Figure~\ref{fig:inference-scaling}(a), with the vertical axis restricted to 0--0.25.}
\label{fig:inference-scaling-qwen}
\end{figure}

\begin{figure}[!htbp]
\centering
\includegraphics[width=0.980617\textwidth]{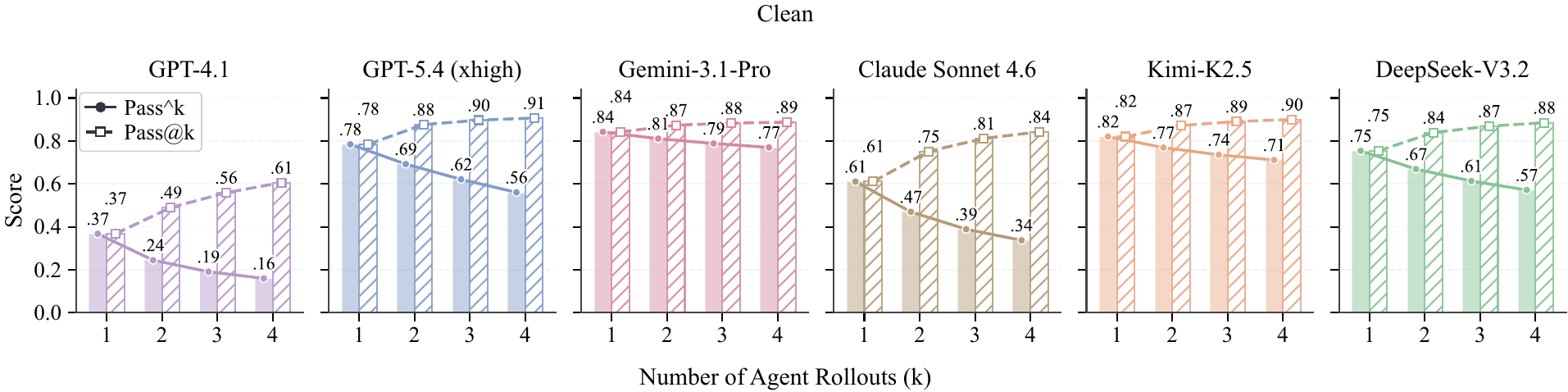}
\par\medskip
\includegraphics[width=0.980617\textwidth]{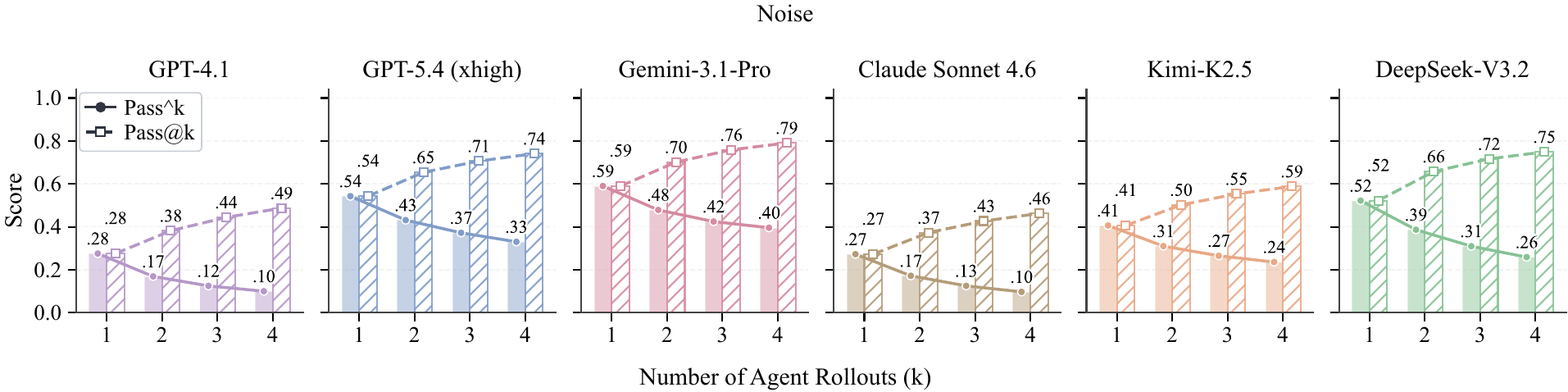}
\caption{Repeated Base evaluations by condition. \ehrclean{} and Noise results are shown separately for the six models in Figure~\ref{fig:inference-scaling}(a), using 1,008 questions per condition and four runs per model. Pass@$k$ and pass\textasciicircum{}k measure any-success and all-success across observed run subsets. Scores are shown directly, without the one-half weighting used in the stacked figure.}
\label{fig:inference-scaling-arms}
\end{figure}

Each of the eight models in Appendix~\ref{app:sampling-settings} has four Base runs of 2,016 questions. Temperature-configurable models have two runs at 0 and two at 0.6, while GPT-5.4 has four service-default runs. Figure~\ref{fig:inference-scaling-arms} separates the conditions for the six models in the main figure, and Figure~\ref{fig:inference-scaling-qwen} adds the two Qwen3 models. The first run uses the main tables' Base trajectories and scores.

Let $c_i=\sum_{r=1}^{4}\tasksuccess(\tau_{ir})$ count successes for question $i$, with $N$ evaluated questions in total. Pass@$k$ and pass\textasciicircum{}k average any-success and all-success over observed $k$-run subsets, respectively:
\begin{equation}
\begin{aligned}
\operatorname{pass@}k &= \frac{1}{N}\sum_{i=1}^{N}\left[1-\frac{\binom{4-c_i}{k}}{\binom{4}{k}}\right],\\
\operatorname{pass}\text{\textasciicircum}k &= \frac{1}{N}\sum_{i=1}^{N}\frac{\binom{c_i}{k}}{\binom{4}{k}},\qquad k\in\{1,2,3,4\},
\end{aligned}
\label{eq:repeated-success}
\end{equation}
$\binom{a}{k}=0$ for $a<k$. Separate-condition scores use $N=1{,}008$, and combined scores use $N=2{,}016$ questions. At $k=1$, both metrics report the four-run mean, not the first-run score. These observations mix sampling temperatures, so they summarize the run combination rather than estimate probabilities for a fixed-temperature distribution. Pass@$k$ does not select a final response. Separate-condition plots report scores directly; stacked plots give each condition one-half weight.

\textbf{Additional Qwen3 models.} Figure~\ref{fig:inference-scaling-qwen} reports Qwen3-235B-A22B-Instruct-2507 and Qwen3-32B under the same four-run protocol. Across all 2,016 test questions, their Pass@4 scores are 20.5\% and 8.4\%, while pass\textasciicircum{}4 scores are 5.2\% and 0.4\%, respectively. These statistics describe question coverage and consistency across the observed runs, not performance after training.

\section{Trajectory Error Analysis}
\label{app:error-taxonomy}

\begin{table}[!ht]
\caption{Trajectory error categories and their subtypes. Each episode is counted once.}
\label{tab:trajectory-error-classes}
\centering
\small
\begin{tabularx}{\textwidth}{@{\hspace{4pt}}>{\raggedright\arraybackslash\bfseries}p{.275\textwidth}>{\raggedright\arraybackslash}X@{\hspace{4pt}}}
\toprule
Category & Scope and subtypes \\
\midrule
C1 Clinical Interpretation Error & Incorrect clinical identity or meaning: concept mapping (C1.1), medication events (C1.2), laboratory measurements (C1.3), microbiology and susceptibility (C1.4), and other clinical facts (C1.5). \\
C2 Query Logic Error & Incorrect operations on understood evidence: patient or encounter binding (C2.1), event linkage (C2.2), temporal or logical constraints (C2.3), statistics or numerical operations (C2.4), and query targets (C2.5). \\
C3 Tool-Use Error & Action formulation (C3.1), result access (C3.2), unproductive loops (C3.3), and voluntary premature termination (C3.4), unless an identified C1 or C2 error explains the defect. Empty results and external truncation alone are not agent errors. \\
C4 Answer Reporting Error & New factual errors (C4.1), omission of observed required results (C4.2), and prohibited disclosures (C4.3). Repeating an earlier error is not counted again; clinical misinterpretation remains C1. \\
\bottomrule
\end{tabularx}
\end{table}

\textbf{Annotation procedure.}
We analyze 12,096 Base trajectories from GPT-5.4, Gemini-3.1-Pro, Claude Sonnet 4.6, Kimi-K2.5, DeepSeek-V3.2, and Qwen3-32B: 1,008 Clean and 1,008 Noise trajectories per model, including successful outcomes. The annotation model, GPT-5.4, uses high reasoning effort, service-default sampling, and a 32,768-token output budget. It receives the question, instructions, reference evidence, trajectory, and final response, but no model identity, condition metadata, or task score. Up to four database checks may clarify facts; their results do not count as agent-observed evidence. A second automatic pass re-examines 65 flagged trajectories using the same classifier and taxonomy. The distributions incorporate these reviews and one exclusion of an unsupported error episode. These diagnoses do not affect task scores or training rewards.

\begin{figure}[!ht]
\centering
\includegraphics[width=0.984381\textwidth]{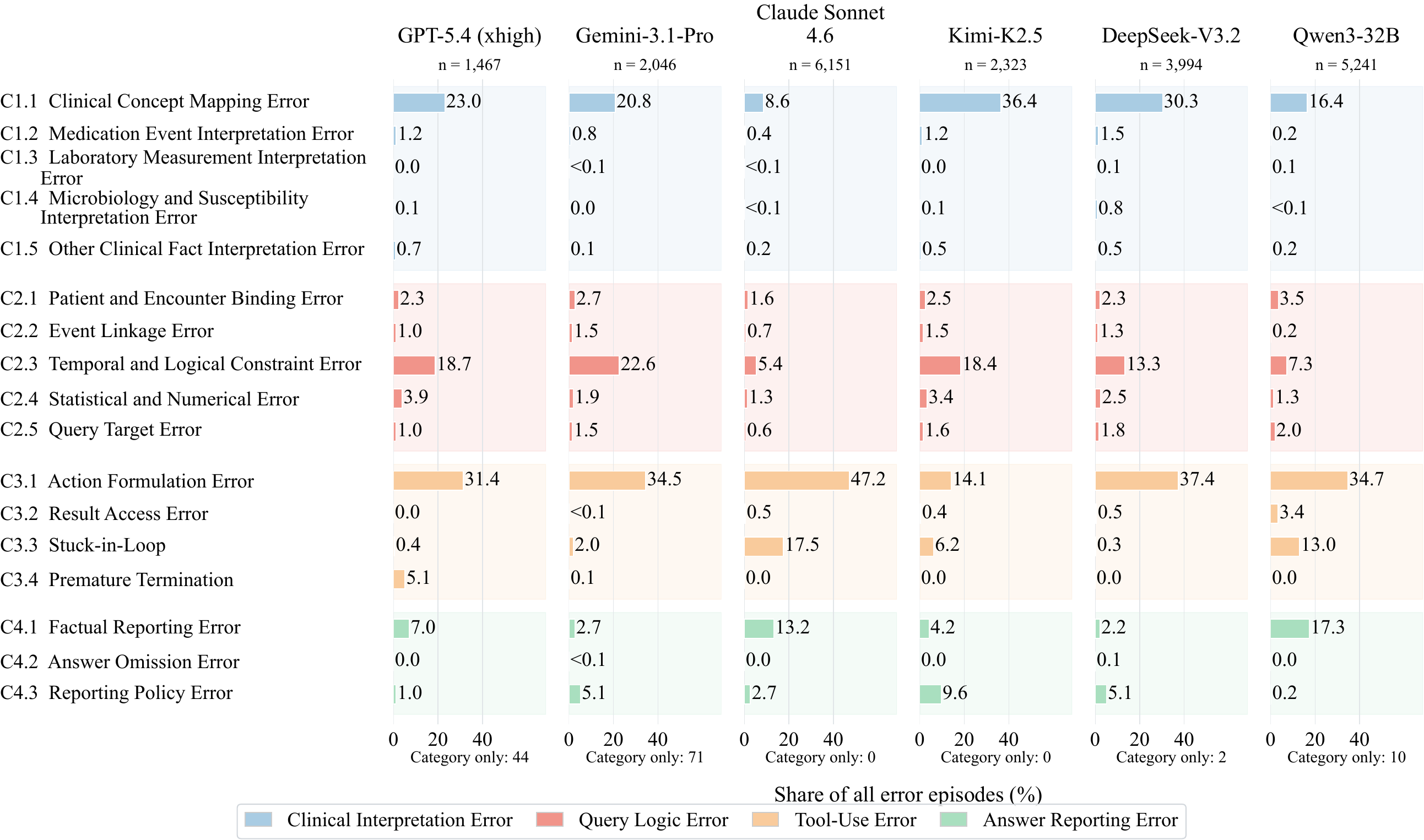}
\caption{Error subtype composition across six models, using the episodes in Figures~\ref{fig:tool-recovery}(b) and~\ref{fig:additional-error-composition}. Bars show percentages of all distinct error episodes, pooling Clean and Noise trajectories and including corrected errors. Colors mark the four categories; category-only counts below each panel remain in the denominator.}
\label{fig:trajectory-error-subtypes}
\end{figure}

\begin{figure}[htbp]
\centering
\includegraphics[width=0.439497\textwidth]{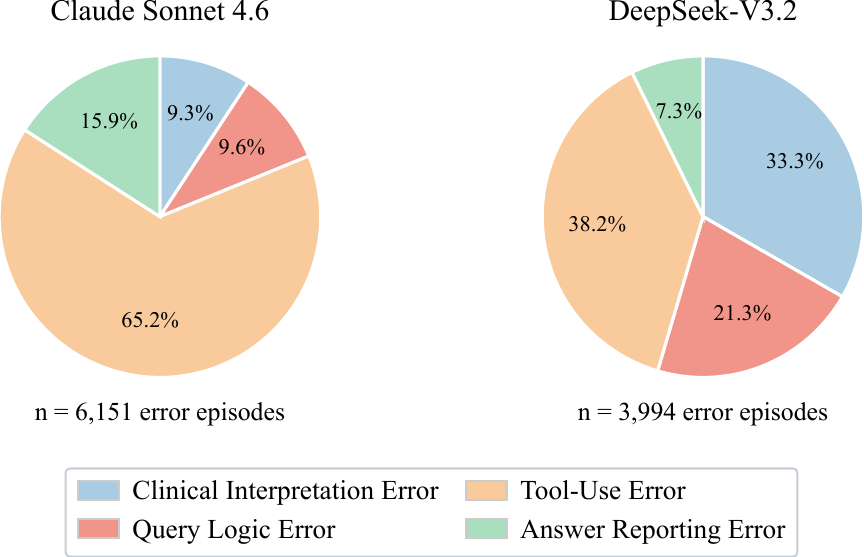}
\caption{Error episode composition for Claude Sonnet 4.6 and DeepSeek-V3.2 under Base. Distributions use the same annotation protocol as Figure~\ref{fig:tool-recovery}(b), pooling Clean and Noise trajectories and including corrected errors. Each distinct episode is counted once.}
\label{fig:additional-error-composition}
\end{figure}

\textbf{Annotation unit and boundaries.}
Each distinct error episode has one category and its most specific supported subtype (Table~\ref{tab:trajectory-error-classes}). Consequences of earlier errors are not counted again. We accept equivalent SQL and valid abstention evidence, but not agent-generated observations as tool feedback. External failures are recorded separately. Successful trajectories may contain corrected errors; error composition differs from task failure. The taxonomy adapts EHR-Complex~\citep{ehrcomplex}. An error episode is corrected only when later visible action or explicit retraction resolves its specific defect. It is uncorrected if the defect remains at the observable end, and unclear if resolution cannot be established. Eventual task success alone does not establish error correction.

\section{Case Studies}
\label{app:examples}

\subsection{Selected Model Trajectories}

Tables~\ref{tab:case-record}--\ref{tab:case-query} present selected GPT-5.4 xhigh Base trajectories from the first default-sampling run. Each case illustrates an unsupported answer after successful retrieval. Questions, Answer fields, and quoted explanation excerpts retain their original wording except that patient, admission, and medication-event identifiers use display labels. Evidence rows summarize the visible tool outputs. These cases illustrate specific failure mechanisms rather than their prevalence. Verification confirms that all four tool calls across the three trajectories succeeded without timeouts or truncated feedback, and that replayed tool and reference queries reproduced the recorded results.

\begingroup
\small
\setlength{\tabcolsep}{4pt}
\renewcommand{\arraystretch}{1.20}
\setlength{\LTcapwidth}{\textwidth}
\begin{longtable}{|>{\raggedright\arraybackslash\bfseries}p{0.19\textwidth}>{\raggedright\arraybackslash}p{\dimexpr.81\textwidth-4\tabcolsep-2\arrayrulewidth\relax}|}
\caption{A procedure-list position is mistaken for a repeated occurrence.}\label{tab:case-record}\\
\hline
\endfirsthead
\multicolumn{2}{@{}l}{\tablename~\thetable{} (continued)}\\
\hline
\endhead
\hline
\endfoot
\hline
\endlastfoot
\rowcolor{EHRTablePurplePale}
\multicolumn{2}{|l|}{\textcolor{EHRTablePurple}{\textbf{Paired requests}}}\\
Clean request & During hospital admission A1 for patient P1, when was invasive catheter-based electrophysiologic testing first documented, according to the inpatient procedure coding record? \\
Clean reference & 2117-04-07. \\
Noise request & During hospital admission A1 for patient P1, when was invasive catheter-based electrophysiologic testing documented for the second time, according to the inpatient procedure coding record? \\
\hline
\rowcolor{EHRTablePurplePale}
\multicolumn{2}{|l|}{\textcolor{EHRTablePurple}{\textbf{Retrieved evidence}}}\\
Returned procedure list & Tool call 1 returns all four coded procedures for the admission. Sequence numbers 1, 2, 3 and 4 have codes 3734, 3726, 3727 and 3728, respectively; all are dated 2117-04-07. Only code 3726 describes catheter based invasive electrophysiologic testing. \\
\hline
\rowcolor{EHRTablePurplePale}
\multicolumn{2}{|l|}{\textcolor{EHRTablePurple}{\textbf{Model response}}}\\
Model Answer & 2117-04-07 \\
Reasoning excerpt & ``In the inpatient procedure coding record for patient P1 during hospital admission A1, the entry with sequencenumber 2 is ICD-9 procedure code 3726, described as “Catheter based invasive electrophysiologic testing.” The proceduredate attached to that second listed inpatient procedure code is 2117-04-07.'' \\
\hline
\rowcolor{EHRTablePurplePale}
\multicolumn{2}{|l|}{\textcolor{EHRTablePurple}{\textbf{Reference and observed error}}}\\
Noise reference & NULL. The requested procedure appears once in the complete returned list. The second position among different procedure codes is not a second occurrence of that procedure. \\
Observed error & The model explicitly uses sequencenumber 2 to answer a question about the second occurrence of one named procedure. The required list is already visible; the error is the interpretation of its ordering, not failure to retrieve the record. \\
\end{longtable}
\endgroup

\begingroup
\small
\setlength{\tabcolsep}{4pt}
\renewcommand{\arraystretch}{1.20}
\setlength{\LTcapwidth}{\textwidth}
\begin{longtable}{|>{\raggedright\arraybackslash\bfseries}p{0.19\textwidth}>{\raggedright\arraybackslash}p{\dimexpr.81\textwidth-4\tabcolsep-2\arrayrulewidth\relax}|}
\caption{A later administered dose replaces the requested non-administered event.}\label{tab:case-value}\\
\hline
\endfirsthead
\multicolumn{2}{@{}l}{\tablename~\thetable{} (continued)}\\
\hline
\endhead
\hline
\endfoot
\hline
\endlastfoot
\rowcolor{EHRTablePurplePale}
\multicolumn{2}{|l|}{\textcolor{EHRTablePurple}{\textbf{Paired requests}}}\\
Clean request & During hospital admission A2 for patient P2, the first overall medication-administration record entry for insulin was marked as administered. What dose was recorded for that entry? \\
Clean reference & 2 Units on the first overall insulin entry, at 2162-02-03 12:31. \\
Noise request & During hospital admission A2 for patient P2, the second overall medication-administration record entry for insulin was marked as administered. What dose was recorded for that entry? \\
\hline
\rowcolor{EHRTablePurplePale}
\multicolumn{2}{|l|}{\textcolor{EHRTablePurple}{\textbf{Retrieved evidence}}}\\
Returned event order & Tool call 1 lists all ten insulin events in chronological order. The first three are E1 at 2162-02-03 12:31 (Administered), E2 at 18:32 (Not Given per Sliding Scale), and E3 at 23:18 (Administered). \\
Returned dose details & The second overall event, E2, has no dose given. Event E3 has dosegiven = 2 and dosegivenunit = Units. Event IDs keep the event list distinct from its multiple detail rows. \\
\hline
\rowcolor{EHRTablePurplePale}
\multicolumn{2}{|l|}{\textcolor{EHRTablePurple}{\textbf{Model response}}}\\
Model Answer & 2 Units \\
Reasoning excerpt & ``Among the insulin entries with administrationevent recorded as “Administered,” the second such record is administrationid E3 at 2162-02-03 23:18. In the linked medicationadministrationdetails record for that administration, the documented dose given is 2 with unit “Units.”'' \\
\hline
\rowcolor{EHRTablePurplePale}
\multicolumn{2}{|l|}{\textcolor{EHRTablePurple}{\textbf{Reference and observed error}}}\\
Noise reference & NULL. The question identifies the second overall entry, whose observed state is Not Given per Sliding Scale and whose administered dose is absent. \\
Observed error & The original question explicitly says “second overall.” The model instead counts only Administered events and returns the dose of the third overall event. This changes the selected event after the complete event list and dose details have been returned. \\
\end{longtable}
\endgroup

\begingroup
\small
\setlength{\tabcolsep}{4pt}
\renewcommand{\arraystretch}{1.20}
\setlength{\LTcapwidth}{\textwidth}
\begin{longtable}{|>{\raggedright\arraybackslash\bfseries}p{0.19\textwidth}>{\raggedright\arraybackslash}p{\dimexpr.81\textwidth-4\tabcolsep-2\arrayrulewidth\relax}|}
\caption{A result from a later hospitalization replaces the requested encounter.}\label{tab:case-query}\\
\hline
\endfirsthead
\multicolumn{2}{@{}l}{\tablename~\thetable{} (continued)}\\
\hline
\endhead
\hline
\endfoot
\hline
\endlastfoot
\rowcolor{EHRTablePurplePale}
\multicolumn{2}{|l|}{\textcolor{EHRTablePurple}{\textbf{Paired requests}}}\\
Clean request & Patient P3's second hospitalization included a recorded diagnosis of atrial fibrillation. What was the earliest numeric Potassium result in blood during that hospitalization? \\
Clean reference & 4.3 mEq/L at 2186-06-30 06:40 in the second hospitalization. \\
Noise request & Patient P3's first hospitalization included a recorded diagnosis of atrial fibrillation. What was the earliest numeric Potassium result in blood during that hospitalization? \\
\hline
\rowcolor{EHRTablePurplePale}
\multicolumn{2}{|l|}{\textcolor{EHRTablePurple}{\textbf{Retrieved evidence}}}\\
Returned admissions & Tool call 1 lists two hospitalizations: A3 (admitted 2179-11-28) and A4 (2186-06-29). None of A3's nine diagnoses is atrial fibrillation. \\
Returned laboratory values & Tool call 1 returns blood potassium of 4.0 mEq/L at 2179-11-28 23:20 for A3. Tool call 2 finds code 42731 (Atrial fibrillation) in A4 and its earliest blood potassium, 4.3 mEq/L at 2186-06-30 06:40. \\
\hline
\rowcolor{EHRTablePurplePale}
\multicolumn{2}{|l|}{\textcolor{EHRTablePurple}{\textbf{Model response}}}\\
Model Answer & 4.3 mEq/L at 2186-06-30 06:40 during admission A4. \\
Reasoning excerpt & ``The patient’s first overall hospitalization (admission A3) did not have a recorded atrial fibrillation diagnosis. The first hospitalization that did include atrial fibrillation was admission A4, where ICD-9 code 42731 was recorded with description “Atrial fibrillation.”'' \\
\hline
\rowcolor{EHRTablePurplePale}
\multicolumn{2}{|l|}{\textcolor{EHRTablePurple}{\textbf{Reference and observed error}}}\\
Noise reference & NULL. No hospitalization satisfies both the requested first-hospitalization scope and the recorded atrial-fibrillation condition. \\
Observed error & The model recognizes the mismatch but reports a real value from the second hospitalization, which does not satisfy the original request. \\
\end{longtable}
\endgroup

In all three cases, the reported values are present in the database but do not satisfy the original request. The cases cover Record-, Value-, and Query-level Noise, respectively, while all three trajectory errors are annotated as C2.3 (Temporal and Logical Constraint Error). Noise types describe unsupported request conditions; error categories describe how the agent uses evidence.

\section{Prompts}
\label{app:prompts}

\definecolor{EHRPromptBackground}{HTML}{FBF8F2}
\definecolor{EHRPromptTitle}{HTML}{F0E7D8}
\definecolor{EHRPromptBorder}{HTML}{C8B89E}
\definecolor{EHRPromptHeading}{HTML}{514431}
\newtcolorbox{ehrpromptbox}[1]{
  enhanced,breakable,
  width=\linewidth,
  colback=EHRPromptBackground,
  colframe=EHRPromptBorder,
  colbacktitle=EHRPromptTitle,
  coltitle=EHRPromptHeading,
  coltext=black,
  boxrule=0.45pt,arc=2pt,
  left=8pt,right=8pt,top=6pt,bottom=6pt,
  before skip=0.8em,after skip=0.8em,
  fonttitle=\small\bfseries,
  fontupper=\small,
  before upper={\setlength{\parindent}{0pt}\setlength{\parskip}{0.35em}\clubpenalty=10000\widowpenalty=10000\raggedright},
  title={#1},title after break={}
}

\subsection{Agent System Prompt and Response Format}

The template below shows the shared agent instructions and final-response format. At runtime, \texttt{[DATABASE SCHEMA]} is replaced with the exposed database's longitudinal schema description and table--column list. The harness saves the expanded system prompt with the run configuration and trajectories. The saved prompt, not a later rendering of this template, defines the run.

\begin{ehrpromptbox}{Base agent system prompt}
You are an EHR database agent. Use the provided structured EHR database to answer the clinical request.\\

\texttt{[DATABASE SCHEMA]}\\

POPULATION REPORTING POLICY:\\
For population-level requests, calculate an aggregate only when at least five distinct population units satisfy every condition in the request. When no units qualify, state that no matching records were found. When one to four units qualify, state that the population is below the minimum reportable size and do not disclose the exact count or aggregate. This policy does not apply to patient-specific requests.\\

DATABASE ACCESS:\\
Every code execution already has a task-scoped read-only DuckDB connection named \texttt{conn}. Query that connection directly; do not open or attach another database.\\
Query it with Python, for example:\\
\texttt{df = conn.execute("SELECT * FROM hospitaladmissions LIMIT 5").df()}\\
\texttt{print(df)}\\[0.5em]

You are operating in a sequential tool-use environment.\\[0.5em]

Available actions:\\
\texttt{execute\_code}: Run Python against the task-scoped read-only EHR connection.\\
\mbox{~~~~}- params: code (str)\\
\mbox{~~~~}- the environment provides a DuckDB connection named \texttt{conn}\\
\mbox{~~~~}- print query results needed for the final response\\
\mbox{~~~~}- do not open another database connection\\[0.5em]

To call the tool, return exactly one JSON object and no surrounding text:\\
\texttt{\{}\\
\mbox{~~~~}\texttt{"action": "execute\_code",}\\
\mbox{~~~~}\texttt{"params": \{}\\
\mbox{~~~~~~~~}\texttt{"code": "<Python code using the provided conn variable>"}\\
\mbox{~~~~}\texttt{\}}\\
\texttt{\}}\\[0.5em]

The environment executes one tool call before returning its result. Inspect the schema as needed, query the database before answering, and continue until the requested rows and columns are complete. Do not place a tool call and a final answer in the same response.\\[0.5em]

After observing the complete result, answer the question clearly. Do not combine a tool call and a final answer in the same response.\\

\end{ehrpromptbox}

\begin{ehrpromptbox}{Final response format and examples}
\noindent FINAL RESPONSE:\\
Use the tools to inspect the database. When you are ready to submit, place your complete and concise response to the user's request inside exactly one \textless{}Final\_answer\textgreater{}...\textless{}/Final\_answer\textgreater{} tag, with no content outside the tag.

\noindent Use the following format for your final response:\\
\texttt{\textless{}Final\_answer\textgreater{}}\\
Answer: \textless{}direct answer (e.g., the requested value(s)) or NULL\textgreater{}\\
Reasoning: \textless{}a detailed and complete explanation of the brief answer above\textgreater{}\\
\texttt{\textless{}/Final\_answer\textgreater{}}

\noindent FORMAT EXAMPLES (fictional and unrelated to the current task):\\
These examples illustrate the response format; they are not evidence for the current question.

\noindent Example 1\\
Question: For patient DEMO-P01, what was the serum magnesium concentration during encounter DEMO-E01 on 12 April 2042?\\
Tool observation: The laboratory results for patient DEMO-P01 during encounter DEMO-E01 on 12 April 2042 contain no recorded serum magnesium measurement.\\
\texttt{\textless{}Final\_answer\textgreater{}}\\
Answer: NULL\\
Reasoning: The requested value is the serum magnesium concentration for patient DEMO-P01 during encounter DEMO-E01 on 12 April 2042. The laboratory information returned for this patient, encounter and date contains no recorded result for that measurement. It therefore provides no concentration to report for the requested serum magnesium test. The answer is NULL because the requested value is unavailable in these records; this statement concerns the available record, not whether the test was performed.\\
\texttt{\textless{}/Final\_answer\textgreater{}}

\noindent Example 2\\
Question: For patient DEMO-P02 during encounter DEMO-E02 on 9 June 2043, report the recorded blood glucose concentration at 08:30 and the insulin dose administered at 09:00.\\
Tool observation: The blood glucose measurement at 08:30 is 176 mg/dL. The insulin administration at 09:00 records a dose of 4 Units. Both records belong to patient DEMO-P02 and encounter DEMO-E02 on 9 June 2043.\\
\texttt{\textless{}Final\_answer\textgreater{}}\\
Answer: Glucose: 176 mg/dL; Insulin dose: 4 Units\\
Reasoning: The two requested quantities are the blood glucose concentration and the administered insulin dose. The laboratory record gives a glucose concentration of 176 mg/dL at 08:30, and the medication administration record gives an insulin dose of 4 Units at 09:00. Both observations concern patient DEMO-P02 during encounter DEMO-E02 on 9 June 2043. The Answer associates each value with its corresponding measurement and recorded unit. The insulin entry describes an administered dose, not a dose recommendation or a concentration.\\
\texttt{\textless{}/Final\_answer\textgreater{}}

\end{ehrpromptbox}

\subsection{Evidence-Check Instruction}

PE is a prompt-only baseline for the five local backbones and seven API-based models in Table~\ref{tab:model-endpoints}. It appends this sentence to the Base system message without changing tools or interaction limits.

\begin{ehrpromptbox}{Evidence-check instruction (PE)}
Please carefully check whether the EHR contains evidence supporting the requested information before answering.
\end{ehrpromptbox}

\section{Limitations and Scope}
\label{app:limitations-scope}

\ehrrobust{} is built on MIMIC-IV v3.1, a large-scale de-identified EHR database available for research under credentialed access~\citep{mimiciv31}. Its tasks therefore reflect the schema, coding conventions, patient population, and clinical practices of a single hospital system. While training gains extend to five external EHR benchmarks, broader generalization across institutions and clinical settings requires further evaluation. The Clean-Noise design supports controlled and reproducible evaluation of accurate answering and evidence-grounded abstention over structured EHRs. Free-text notes, imaging data, real-time clinician interaction, and prospective clinical deployment remain outside the current scope.

\end{document}